\documentclass[twoside,11pt]{article}
\usepackage[utf8]{inputenc}

\usepackage[nohyperref]{jmlr2e}
\usepackage{preamble} % additional packages compatible w jmlr2e.sty -- hyperref loaded here for compatibility with cleverref
\usepackage{macros} % personal macros
\usepackage{glossary} % personal glossary settings

\colorlet{darklilac}{blue!40!white}

\usepackage{lastpage}
\jmlrheading{27}{2026}{1-\pageref{LastPage}}{5/24; Revised 9/25}{5/26}{24-0808}{Laura Iacovissi, Nan Lu, Robert C. Williamson}

\ShortHeadings{Corruptions of Supervised Learning Problems: Typology and Mitigations}{Iacovissi, Lu, Williamson}
\firstpageno{1}

\begin{document}

\title{Corruptions of Supervised Learning Problems: \\ Typology and Mitigations}

\author{%
    {\name Laura Iacovissi} 
    \email laura.iacovissi@uni-tuebingen.de \\
    \addr T\"{u}bingen AI Center, University of T\"{u}bingen 
    \AND
    {\name Nan Lu} 
    \email nan.lu@uni-tuebingen.de \\
    %\email lunan.bit@gmail.com \\
    \addr T\"{u}bingen AI Center, University of T\"{u}bingen 
    %\addr University of Bristol
    \AND
    {\name Robert C. Williamson}  
    \email bob.williamson@uni-tuebingen.de \\
    \addr T\"{u}bingen AI Center, University of T\"{u}bingen 
}

% \author{%
%   \name Laura Iacovissi \email laura.iacovissi@uni-tuebingen.de \\
%   \name Nan Lu \email nan.lu@uni-tuebingen.de \\
%   \name Robert C. Williamson \email bob.williamson@uni-tuebingen.de 
%   \AND
%   {\addr University of T\"{u}bingen and T\"{u}bingen AI Center}
% }

\editor{Ambuj Tewari}

\maketitle

\begin{abstract}%   <- trailing '%' for backward compatibility of .sty file
    Corruption is notoriously widespread in data collection. Despite extensive research, the existing literature predominantly focuses on specific settings and learning scenarios, lacking a unified view of corruption modelization and mitigation. 
    In this work, we develop a general theory of corruption, which incorporates all modifications to a supervised learning problem, including changes in model class and loss.
    % We then focus on changes in probability distributions induced by Markov kernels; this perspective leads to three novel opportunities. 
    Focusing on changes to the underlying probability distributions via Markov kernels, our approach leads to three novel opportunities.
    First, it enables the construction of a novel, provably \emph{exhaustive} corruption framework, distinguishing among different corruption types.
    This serves to unify existing models and establish a consistent nomenclature.
    Second, it facilitates a systematic analysis of corruption consequences on learning tasks, by considering Bayes risks in the clean and corrupted scenarios.
    Notably, while label corruptions affect only the loss function, attribute corruptions additionally influence the hypothesis class. 
    Third, building upon these results, we investigate mitigations for various corruption types.
    We expand existing loss-correction methods for label corruption to handle dependent corruption types. 
    Our findings highlight the necessity to generalize this classical corruption-corrected learning framework to a new paradigm with weaker requirements to encompass more corruption types. 
    We provide such a paradigm as well as loss correction formulas in the attribute and joint corruption cases.
\end{abstract}

\begin{keywords}
  learning theory, Markov kernels, Markovian corruption, noisy data, loss correction
\end{keywords}

% Core

\section{Introduction}
Machine learning starts with data. 
The most widespread conception of data defines it as atomic facts, perfectly describing some reality of interest \citep{poovey1998history}. 
In learning theories, this is reflected by the often-used assumption that training and test data are drawn identically and independently from some fixed probability distribution. 
The goal of learning is then understood as identifying and synthesizing patterns embedded in these data. 
In the field of machine learning, considerable attention has been devoted by engineers and researchers to the task of designing suitable loss functions or model architectures; however, less effort has been put into data, given that they are often not responsible for collecting them but rather for processing them \citep{sambasivan2021everyone}. 
In practice, corruption regularly occurs in data collection, creating a mismatch between training and test distributions and forcing models to learn from imperfect facts.

We should thus doubt the view of data as static facts, and consider them as a dynamic element of a learning task \citep{Williamson2020Process}.
In addition to the traditional emphasis on prediction models and loss functions in machine learning, one may focus on the data dynamic itself, so as to understand how different processes may have led us to the observation of certain data, and furthermore, how they subsequently impact the learning process. 
While the necessity of investigating this topic is recognized both at a practical \citep{world2018prevent,malinin2021shifts, koh2021wilds} and a theoretical \citep{meng2021enhancing, rostamzadeh2021thinking} level, no standardized way to model and analyze the dynamic generative process of data has been so far created.

In the field of machine learning, changes in such dynamic process are often referred to as \emph{distribution shift} or \emph{noisy data}. 
Here, we adopt a more inclusive term \emph{corruption}, drawing from the computer science literature.
%\footnote{Notably, the terms ``distribution shift'' and ``data corruption'' implicitly presuppose the existence of unknown unbiased/clean data or a true generative process of the data. Whether this assumption is legit in every case is not discussed here.}  
Our conceptualization of corruption goes beyond traditional notions: it encompasses all modifications to a learning problem, including changes to the loss function, hypothesis class, or probability distribution from which data are drawn. 
We interpret corruption not as inherently pejorative, but as a general \emph{modification process}. 
Whether the corruption is positive, negative, or neutral, depends on the specific context in which it is applied. 

A similar stance has been taken in the recent work from \citet{memoli2024geometry}: they define corruption in an analogous general fashion, while additionally allowing for changes in the attribute and label sets.
Their goal is to quantify the distance between learning problems, which they achieve by defining a suitable pseudo-metric.
On the other hand, in the present paper we specifically focus on modifications of the probability distribution, aiming to address the lack of understanding of \emph{data as a process} and provide a complete description of this class of corruptions.
%However, we emphasize that this is only one of the possible ways a learning problem can change.

Recognizing corruption as a dynamic element of learning has led to considerable research into specific data corruption models \citep{angluin1988learning, zhang2013domain, natarajan2013learning, patrini2017making, shimodaira2000improving, quinonero2008dataset, zhang2020one}.
However, these approaches cannot, even in principle, answer questions regarding the comparison of different types of corruption.
%The lack of a unified nomenclature further complicates progress, underscoring the need for a comprehensive, systematic framework.
Inconsistent naming of the same corruption model across different works further slows down progress, and highlights the need for a comprehensive framework.

Whilst there have been attempts to build such a framework, certain limitations persist in terms of homogeneity and exhaustiveness. 
A famous early endeavor is \citet{quinonero2008dataset}, which groups together works about the multi-faceted topic of dataset shift, yet not in a unifying or comprehensive manner.
Later on, several studies sought to offer a more homogeneous view of corruption \citep{moreno2012unifying,kull2014patterns,saez2022taxonomy,subbaswamy2022unifying}; nonetheless, these frameworks typically rely on the assumption of a corruption-invariant marginal or conditional probability.
The extent of their exhaustiveness in representing all potential corruption models within their framework is merely conjectured, or left unexplored.

Therefore, the primary objective of this work is to improve the existing understanding of corruption by introducing a novel perspective, while making use of the classical probabilistic approaches. 
Probability distributions are the only representation of data that will be used in this work, so the terms ``data'' and ``data distribution'' are used interchangeably. 
%\footnote{It is worth noting that one could abstract further, and formalize data and its corruption as a non-probabilistic \citep{meng2022comments,boyd2023we} or imprecise process \Laura{ref datamodels}; this aspect will be briefly discussed, but is beyond the scope of the present paper.}  
This approach allows us to systematically study and compare the possible types of corruption in supervised learning problems, and provides a general framework for analyzing their mitigation. %as an initial step toward unraveling these fundamental questions. 
%However, this does not imply that this ``probabilistic stance'' which effectively equates data with a distribution is universally valid or justified.

\subsection{Motivations, Approach, and Contributions}

\begin{table*}[t]
    \centering
    {\footnotesize
    \caption{Examples of models proposed in the literature that capture data corruptions with probabilistic descriptions. Here, $\rv X$ represents the attribute, and $\rv Y$ represents the label. Details and references are included in \cref{sec:related}.}
    \label{tab:corr-model-intro}
    \begin{tabular}{ll} 
        \toprule
        \textbf{Models}  & \textbf{Descriptions} \\
        \midrule
        Attribute noise & \makecell[l]{$P(\rv X)$ is corrupted due to, e.g., additive attribute noise or missingness,\\ while the labels remain untouched} \\
        \midrule
        \makecell[l]{Random classification\\ noise}
        & \makecell[l]{Considering $P(\rv Y \mid \rv X)P(\rv X)$, $P(\rv Y \mid \rv X)$ is corrupted by flipping each label\\ independently with a constant probability, while $P(\rv X)$ remains invariant} \\
        \midrule
        Class-conditional noise 
        & \makecell[l]{Considering $P(\rv Y \mid \rv X)P(\rv X)$, $P(\rv Y \mid \rv X)$ is corrupted by flipping labels with \\ a probability dependent on the label, while $P(\rv X)$ remains invariant} \\
        \midrule
        Instance-dependent noise 
        & \makecell[l]{Considering $P(\rv Y \mid \rv X)P(\rv X)$, $P(\rv Y \mid \rv X)$ is corrupted by flipping labels with \\ a probability dependent on the instance, while $P(\rv X)$ remains invariant} \\
        \midrule
        \makecell[l]{Instance- \& label-\\ dependent noise}
        & \makecell[l]{Considering $P(\rv Y \mid \rv X)P(\rv X)$, $P(\rv Y \mid \rv X)$ is corrupted by flipping labels with \\ an instance- \& label-dependent probability, while $P(\rv X)$ remains invariant} \\
        \midrule
        \makecell[l]{Mutually contaminated \\ distributions}
        & \makecell[l]{Considering $P(\rv X \mid \rv Y)P(\rv Y)$, $P(\rv X \mid \rv Y)$ is corrupted by a mixture model,\\and $P(\rv Y)$ can also be corrupted} \\
        \midrule
        Combined simple noise 
        & \makecell[l]{Considering $P(\rv Y \mid \rv X)P(\rv X)$, $P(\rv X)$ is corrupted by additive noise, and \\ $P(\rv Y \mid \rv X)$ is corrupted by flipping labels with a probability dependent \\ on the label} \\
        \midrule
        Target shift 
        & \makecell[l]{Considering $P(\rv X \mid \rv Y)P(\rv Y)$, $P(\rv Y)$ is corrupted while $P(\rv X \mid \rv Y)$ remains \\ invariant } \\
        \midrule
        Covariate shift 
        & \makecell[l]{Considering $P(\rv Y \mid \rv X)P(\rv X)$, $P(\rv X)$ is corrupted while $P(\rv Y \mid \rv X)$ remains \\ invariant} \\
        \midrule
        Generalized target shift 
        & \makecell[l]{Considering $P(\rv X \mid \rv Y)P(\rv Y)$, $P(\rv Y)$ and $P(\rv X \mid \rv Y)$ are corrupted, subject to \\ specific  invariance  assumptions on conditional distributions in the  latent \\ space} \\
        \midrule
        Style transfer 
        & \makecell[l]{To model it probabilistically, we express it as $P(\rv X \mid \rv Y)$ being changed given \\ the designated style} \\
        \midrule
        Adversarial noise & \makecell[l]{To model it probabilistically, we express it as $P(\rv X)$ being intentionally \\ corrupted by an adversary to alter the correct prediction for each instance} \\
        \midrule
        Concept drift 
        & $P(\rv X, \rv Y)$ changes over time \\
        \midrule
        Concept shift 
        & Considering $P(\rv Y \mid \rv X)P(\rv X)$, $P(\rv Y \mid \rv X)$ changes over time \\
        \midrule
        Sampling shift 
        & \makecell[l]{Considering $P(\rv Y \mid \rv X)P(\rv X)$, $P(\rv X)$ changes over time, while $P(\rv Y \mid \rv X)$ is \\ invariant} \\
        \midrule
        Selection bias 
        & $P(\rv X, \rv Y)$ is corrupted to $\t P(\rv X, \rv Y)$ s.t. $\t P \ll P, \exists!~\alpha = \frac{d\t P}{dP} ~\&~ || \alpha ||_{\infty} < \infty$ \\
        \bottomrule
    \end{tabular}
    }
\end{table*}

We observed in the previous section a continued surge of research papers dedicated to specific models of corruption. %predominantly relying on explanations of changes and invariance in specific probabilities. 
In \cref{tab:corr-model-intro}, we provide a non-comprehensive list of such models that can be conceptualized as corruption in our sense. 
Rather than adding to this already diverse landscape, we propose a taxonomy to systematically organize them. 
This taxonomy serves as a comprehensive map of probabilistic corruption models, currently absent in the field. Our approach distinguishes itself from the majority of papers in this line, which often propose new corruption models and tailored mitigation algorithms. 
We, on the other hand, explore their mitigation without the aim of introducing a new algorithm, but for gaining a deeper understanding of theory behind the existing ones. 
Details about our contributions are summarized in the following.

\textbf{C1 Understanding Corruptions and Their Types.} 
A common definition of corruption found in the literature is the one of distributional shift. We shape our notion of Markovian corruption inspired by such a concept and making use of Markov kernels.
However, attributing failures in learning solely to changes in probability distributions is restrictive.
For this reason we broaden the concept of corruption to a general one that includes changes in model class and loss function {(\cref{df:gen-corruption})}. 
Focusing on Markovian corruption {(\cref{df:corruption})}, we establish a taxonomy grounded in its dependence on the input and output spaces {(\cref{fig:corruptions,fig:combinations})}. 
This allows us to uncover commonalities among different models of corruptions (refer to \cref{tab:corr-model-sec3} for the correspondence of \cref{tab:corr-model-intro} in our taxonomy), thus transcending the diverse terminologies used by different authors. 
Our resulting framework is proven to be exhaustive for all possible one-step probabilistic corruptions. 
More generally, we prove that every change in probability distribution can be represented via a one-step Markovian corruption or a non-factorized one (\cref{prop:exhaustiveness}). The statement has interesting consequences in terms of how we think of alternative corruption models, \ie, not within the taxonomy. For instance, arguments may be made for non-probabilistic corruptions that change the probability associated with events in a manner not adhering to probability principles (\eg, \cite{boyd2023we}). In this context, \cref{subsec:relation} analyses two popular corruption models---selection bias and mutually contaminated distributions---and demonstrates that they are not one-step Markovian in their original definition; however, they have a one-step Markovian representation. For both of them we gain new insights by relating them to our framework.

\textbf{C2 Consequences of Corruption on Learning Problems.}
Recently, \citet{williamson2022information} deepened the understanding of the relationship between information and  Bayes risk of a statistical decision problem, and yielded Information Processing \emph{Equalities} for a certain class of simple corruptions. 
We connect to this work and relate the Bayes risk of clean and corrupted supervised learning problems through equality results for all corruptions in our taxonomy. Such equalities, illustrated in \cref{sec:brchange} (Theorems \ref{thm:br-s-s} to \ref{thm:br-2-2}), effectively prove the equivalence between two learning problems: the former corrupted in a \emph{Markovian fashion}, the latter via a \emph{general corruption} changing only model class and loss function through a Markov kernel.
A characteristic of this analysis is its neat avoidance of dependence on specific algorithms, which provides an agnostic means of comparison for corruption types. Such comparison is, in our results, only qualitative, and lays the foundation for future quantitative studies. 
One of our main findings is the understanding that label corruptions only affects the loss function, while the model class remains untouched by the corruption kernel; however, for more intricate cases also involving attribute corruptions, both the loss function and the model class are modified.  

\textbf{C3 A Systematic Analysis of Kernel-Based Mitigations.}
Applying the Bayes risk equalities, we derive in Section \cref{sec:losscorrect} corruption-corrected loss functions for each of the different corruption instances within our framework. We first identify the need of generalizing the concept of classical corruption-corrected learning since it becomes outmoded when considering forms of corruption beyond label corruption.
Within the proposed generalized corruption-corrected learning framework, we find a hierarchy-induced set of results on how the optimization problem changes under various corruptions, and how to abstractly compute their loss corrections in \cref{thm:losscorr-label,thm:losscorr-gen}. 
We conclude that more complex corruptions are more detrimental, and require more sophisticated designs than mitigation via classical loss correction. 

\section{Technical Background}
\label[section]{sec:background}
\subsection{Markov Kernels}

We now introduce the mathematical machinery used for modeling corruption in learning problems; that is, Markov kernels and some of their relevant properties. The material reported here is drawn from %\citep{chang1997conditioning, cinlar2011ProbabilityAS, van2017theory, kallenberg2017random, klenke2007probability, statistical2023johnston}
\citep{klenke2007probability, cinlar2011ProbabilityAS, van2017theory, kallenberg2017random, statistical2023johnston}; the reader can refer to them for a comprehensive understanding of kernels in probability and learning theory.

\begin{definition}[\citet{klenke2007probability}] \label[definition]{def:markov-kernel}
    Let $(X_1,\mc{X}_1)$ and $(X_2,\mc{X}_2)$ be Polish spaces equipped with Borel $\sigma$-algebras, \ie, standard Borel measurable spaces. Let $\ka$ be a mapping from $X_1 \tm \mc{X}_2$ into $[0, +\infty]$. Then, $\ka$ is called a \textbf{transition kernel} from $(X_1,\mc{X}_1)$ to $(X_2,\mc{X}_2)$ if
    \vspace{1ex}
    \begin{enumerate}
        \item the mapping $x_1 \in X_1 \to \ka(x_1,B)$ is $\mc{X}_1$-measurable for every set $B \in \mc{X}_2$, and
        \item the mapping $B \in \mc{X}_2 \to \ka(x_1,B)$ is a measure on $(X_2,\mc{X}_2)$ for every $x_1 \in X_1$.
    \end{enumerate}
    \vspace{1ex}
    A kernel is said to be a \textbf{Markov kernel} if $\ka(x_1,X_2) = 1 \; \forall x_1 \in X_1$, \ie, it maps to a probability measure; this is denoted by the compact notation $\ka\colon X_1 \karrow X_2$.\footnote{This notation is borrowed from category theory, see \citep{markov2020parzygnat} for a primer.} 
    The set $X_1$ is said to be the domain of $\ka$, and $X_2$ its image, \ie, 
    $$ D(\ka) = X_1\,, \quad I(\ka)=X_2 \,. $$
    We refer to the set of kernels as $\mathcal{T}(X_1, X_2)$ and its subset of Markov kernels as $\mathcal{M}(X_1, X_2)$.
\end{definition}

To better grasp the concept of Markov kernel, we can think of it as a parameterized family $\ka(x_1, \cdot) \,, x_1 \in X_1$ of probability measures on the space $(X_2,\mc{X}_2)$. 
It can be interpreted as an \emph{observation channel}, a concept rooted in information theory and properly formalized in \citep{csiszar1972class}. 
In this context, a Markov kernel serves as a detailed probabilistic description of the generative process leading from a ``hidden value'' $X_1$ to \emph{observed} distribution on $X_2$.%
\footnote{In fact, under our assumptions they are the \emph{regular conditional probabilities} associated to the coupling of the spaces $(X_1,\mc{X}_1)$ and $(X_2,\mc{X}_2)$, see \citet{cinlar2011ProbabilityAS}.}
As such, for finite spaces they can be represented as stochastic matrices.

\begin{example}
    Consider a set $Z=\{0,1\}$, a kernel $\ka \in \mc M(Z,Z)$, and random variables $\rv Z, \t{\rv{Z}}$ on $(Z, \mc Z)$. We can conveniently write the kernel as the matrix of the conditional probabilities:
    \begin{align*}
        \ka \coloneqq 
        \begin{bmatrix}
            P(\t{\rv{Z}}=1 \mid \rv{Z}=1) & P(\t{\rv{Z}}=1 \mid \rv{Z}=0) \\ 
            P(\t{\rv{Z}}=0 \mid \rv{Z}=1) & P(\t{\rv{Z}}=0 \mid \rv{Z}=0)          
        \end{bmatrix} =
        \begin{bmatrix}
             0.7 & 0.5 \\ 
             0.3 & 0.5          
        \end{bmatrix}  
        \,.
    \end{align*}
\end{example}

Special types of kernels, which will be extensively used in our analysis, are:
\begin{itemize}
    \item A \defemph{trivial} Markov kernel, defined on the trivial domain space $\{*\}$ and taking values in the set $\{\nu\}$, $\nu$ being a probability distribution. This kernel will therefore be equivalent to the probability distribution itself. We can formally write 
    \be \label{def:trivial-kernel}
    \ka_{\nu} \colon \{*\} \karrow X_1 \,, \quad \ka_{\nu} \equiv \nu \,, \quad \nu \in \mc P(X_1) \,,
    \ee
    with $(\{*\}, \{ \{*\} , \emptyset \} )$ a measurable space with only one element.%
    \footnote{This set should be regarded as a placeholder. The value of $*$ does not influence the output of the kernel, which will in any case be $\nu$.}
    In the text, we will simplify the notation by directly using $\nu$ instead of $\ka_{\nu}$;
    \item A Dirac delta kernel, \ie, an \defemph{identity} kernel, defined as $\delta_{X_1}\colon X_1 \karrow X_1$, such that for all $A\in \mc{X}_1$, we have $\delta_{X_1}(x,A) = 1$ if $x \in X_1$, $\delta_{X_1}(x,A) = 0$ otherwise.
\end{itemize}

\subsubsection{Kernel Actions} \label{sec:actions}
A Markov kernel naturally induces two useful functionals, one on distributions and one on functions. 
They are defined as:
\begin{align*}
    \cdot \, \ka \colon \mc{P}(X_1) \to \mc{P}(X_2) \quad \qquad \mu \ka (B)  &\coloneqq \int_{X_1} \mu(d x_1)\, \ka(x_1, B) \ \qquad \forall B \in \mc X_2 \,, \\
    \ka \, \cdot \colon L^0(X_2, \bb R) \to L^0(X_1, \bb R) \qquad \ka f(x_1) &\coloneqq \int_{X_2}  \ka(x_1, d x_2)\, f(x_2)  \qquad \forall x_1 \in X_1 \,,
\end{align*}
provided the integral exists and assuming that $\mc{P}(X)$ refers to the set of probabilities on a set $X$. We refer to these operators as the \emph{actions} of kernels on distributions and functions, respectively.

Equipped with their action, Markov kernels can now be seen as a point-wise probabilistic description of the distortion process applied to a probability distribution $\mu$ on $X_1$, transforming it into another \emph{observed} distribution on $X_2$; equivalently, we can make a similar comment for functions $f$ of $X_2$. Again, Markov kernels are nothing else than \emph{observation channels} \citep{csiszar1972class}.

\subsubsection{Kernel Operations} \label{sec:operations}
Kernels can be combined through different operations. We introduce them here briefly, mainly inspired by \citep{kallenberg2017random,statistical2023johnston}, covering all the necessary properties for this work. We will use henceforth the notation $(X_i, \mc X_i)$ for a standard Borel measurable space. We remark that specifying the kernel action operator $\ka f $ for all measurable $f$ effectively defines a kernel as $\ka(x_1, B) \coloneqq \ka \chi_B(x_1)$ \citep[Remark 6.4]{cinlar2011ProbabilityAS}, where $\chi_B(x_2)$ is the indicator function for $x_2 \in X_2, \, B \in \mc X_2$.

The first set of operations defined here can be referred to as \emph{in-series operations}, given that the involved kernels are required to satisfy specific conditions on the spaces for which they are defined. These operations impose a \emph{more stringent set of feasibility conditions}.

\begin{property} \label[property]{p:comp} 
Given $\ka\colon X_1 \karrow X_2$ and $\lambda\colon X_2 \karrow X_3$, their \textbf{chain composition} is a kernel $\ka \circ \lambda\colon X_1 \karrow X_3$
uniquely determined by the following kernel action:
$$(\ka \circ \lambda) f (x_1)  \coloneqq  \int_{X_2} \ka(x_1,dx_2) \int_{X_3} \lambda(x_2,dx_3) f(x_3) \;,$$
where $f \colon X_3 \to \bb R$ is a positive $\mc X_3$-measurable function.
\end{property} 
    
\begin{property} \label[property]{p:prod} 
Given $\ka\colon X_1 \karrow X_2$ and $\lambda\colon X_1 \tm X_2 \karrow X_3$, their \textbf{product composition} is a kernel $\ka \tm \lambda\colon X_1 \karrow X_2 \tm X_3$ 
uniquely determined by the following kernel action:
$$(\ka \tm \lambda) f (x_1) \coloneqq  \int_{X_2} \ka(x_1, dx_2) \int_{X_3} \lambda((x_1,x_2),dx_3) \, f(x_2,x_3)\;,$$
for every $f$ positive $\mc X_2 \tm \mc X_3$-measurable.
\end{property} 

The operations defined above can naturally understood using well-known probability theory results.
Consider the trivial Markov kernel
$$\ka_\nu\colon \{*\} \karrow X_1, \enspace \nu \in \mc P(X_1).$$
In this setting, the operations \cref{p:comp} and \cref{p:prod} apply to $\ka_\nu$ when composed with a kernel $\lambda_1\colon X_1 \karrow X_2$ (for the chain composition) and $\lambda_2\colon \{*\} \tm X_1 \karrow X_2$ (for the product composition).

This allows us to write distribution-kernel combinations using the same notation as kernel-kernel ones, \ie, $\ka_\nu \circ \lambda_1$ and $\ka_\nu \tm \lambda_2$.%
\footnote{Strictly speaking, $\ka_\nu \circ \lambda_1 \in \mc M(\{ * \}, X_2)$, while $\lambda_1\nu \in \mc P(X_2)$. So this identification holds up to a suitable ``projection''. However, we avoid this level of technicality to keep the presentation clear and simple.}
Both of these constructions result in \emph{new probability measures}:
\begin{itemize}
    \item The composition $$\nu \circ \lambda_1 \coloneqq \ka_\nu \circ \lambda_1 \, \in \mc P (X_2)$$ is equivalent to the kernel action on probabilities $\lambda_1\nu$, and corresponds to the Law of Total Probability. We adopt the $\circ$ notation for it from now on. 
    \item The product $$\nu \tm \lambda_2 \coloneqq \ka_\nu \tm \lambda_2 \, \in \mc P(X_1 \tm X_2)$$ corresponds to the Bayesian decomposition of a joint probability into a marginal $\nu$ and a conditional probability $\lambda_2$.
\end{itemize}
Notice that $\lambda_2$ is essentially of the same type as $\lambda_1$, apart from a dummy variable over the singleton set $\{*\}$.  
In what follows, we will overload the notation and also write $\nu \tm \lambda_1 \, \in \mc P(X_1 \tm X_2)$.

The second set of operations defined here can be referred to as \emph{parallel operations}. Compared to in-series operations as in \cref{p:comp} and \cref{p:prod}, it allows for more flexible combinations of kernels.

\begin{property} \label[property]{p:super} 
Given $\ka\colon X_1 \karrow X_2$ and $\lambda\colon X_3 \karrow X_4$, their \textbf{superposition} is a kernel $\ka\otm\lambda \colon X_1 \tm X_3 \karrow X_2 \tm X_4$ 
uniquely determined by the following kernel action:
$$(\ka\otm\lambda) f (x_1,x_3) \coloneqq  \int_{X_2} \ka(x_1,dx_2) \int_{X_4} \lambda(x_3,dx_4) \, f(x_2,x_4)\;,$$ 
where $f\colon X_2 \tm X_4 \to \bb R$ is positive $\mc X_2 \tm \mc X_4$-measurable.
\end{property}

\begin{remark}
    Observe that no restriction is imposed on the parameter spaces to be equal, \eg, $X_1 = X_3$,  or Cartesian products with some space in common, \eg, $X_1 = Y_1 \tm Y_2, X_3 =  Y_1 \tm Y_3$ .
    When this happens, the actions of the two kernels ``superpose'' on the same space. It is possible for the superposition operation to produce a kernel
    $$(\ka \otm \lambda)f(x_1) := \int_{X_2} \ka(x_1,dx_2) \int_{X_3} \lambda(x_1,dx_3) \, f(x_2,x_3)\;,$$
    with $\ka \colon X_1 \karrow X_2 $ and $\lambda\colon X_1 \karrow X_3$, so that $\ka \otm \lambda \colon X_1 \karrow X_2 \tm X_3$.
    Another possible case is  $\ka' \colon X_1 \karrow X_2 $ and $\lambda' \colon X_1\tm X_2 \karrow X_3$, leading to
    $$\ka' \otm \lambda' = \ka' \tm \lambda',$$
    which makes the superposition a generalization of the product operation. In this case, we will use the $\tm$ symbol.
    However, in case we have more than one measure acting on the same space, the superposition integral would be ill-defined, making some combinations unfeasible. 
\end{remark}

Because of the above properties, we say that \cref{p:super} is the operation with the \emph{weakest feasibility conditions}, the set of rules to fulfill for a well-defined operation.

The last set of operations, introduced by us, can be described as a mid-way between the chain composition \cref{p:comp} and the superposition \cref{p:super}.  
\begin{property} \label[property]{p:part-chain}
    Given $\ka\colon X_1 \tm X_2 \karrow X_3$ and $\lambda\colon X_1 \tm X_3  \karrow X_4$, their \textbf{partial chain composition} is a kernel $\ka \circ_{X_3} \lambda\colon X_1 \tm X_2 \karrow X_4$ 
    uniquely determined by the following kernel action:
    $$(\ka \circ_{X_3} \lambda) f (x_1, x_2) \coloneqq \int_{X_3}  \ka((x_1,x_2),dx_3) \int_{X_4} \lambda((x_1,x_3),dx_4) \, f(x_4) \;,$$
    where $f\colon X_4 \to \bb R$ is a positive $\mc X_4$-measurable function.
\end{property}
Essentially, this operation only chains the kernels on the specified space, here $X_3$, while superposing them on the common parameter, $X_1$.

\subsection{Statistical Experiments and Supervised Learning}
\label[subsection]{sec:learning}

After establishing the notation for working with kernels, we are now ready to deploy the framework in the learning context.
The content presented here is connected to the literature on \emph{statistical experiments} and \emph{decision theory} \citep{torgersen1991comparison, shiryaev2000statistical}.
We summarize the key concepts crucial to our analysis and direct readers to relevant books for a more comprehensive perspective.

\subsubsection{The General Learning Problem} 
In statistical decision theory, a general learning problem can be viewed as a two-player game between \emph{Nature} and a \emph{decision-maker}.
Here, Nature represents an unknown process that generates the observed phenomena; the decision-maker observes the said phenomena and seeks to find the optimal action for each observation within the context of a given task.
Slightly more formally, Nature here stands for the (stochastic) act choosing an \emph{observation} $o \in O$ given some hidden \emph{state} $\theta \in \Theta$.
The stochastic process generating $o$ given $\theta$ is referred to as the \emph{experiment} $E$.

\begin{definition} \label[definition]{def:experiment}
    An \textbf{experiment} $E \colon \Theta \karrow O$ is a Markov kernel from the hidden state space to the observation space.
\end{definition}

The parameter space $\Theta$ and the observation space $O$ are fixed by the setting of the decision problem. We need to specify an additional set, the decision space $A$, to introduce the modeling of the decision-maker.
Having observed a phenomena via $E$, the decision-maker aims to construct a \emph{decision rule} $D$ mapping from the observation space $O$ to the action space $A$. The decision-making task can therefore be represented by the transition diagram
\begin{tikzcd}[arrows={decorate, decoration={snake,segment length = 1.5mm,amplitude=0.2mm}}, column sep = width("decision rule")]
\Theta 
\arrow[r, "E", "experiment"']
& O 
\arrow[r, "D", "decision\ rule"']  
& A,
\end{tikzcd}
where the decision rule is also modeled by a Markov kernel. 
Hence, it is interpreted as a \emph{stochastic rule} fixing a probability on the action space $A$ instead of the classical deterministic view.
In order to evaluate the performance of the decision maker with respect the optimal decision established by Nature, one introduces the concept of loss function and therefore of learning problem.

\begin{definition} \label[definition]{def:learn-prob}
    Consider the product space $(\Theta \tm O \tm A,\, \Omega \tm \mc O \tm \mc A)$, where $(\Theta, \Omega)$, $(O, \mc O)$, and $(A, \mc A)$ are standard Borel measurable spaces for parameters, observations and decisions.
    We refer to the space $(\Theta \tm O,\, \Theta \tm \mc O)$ as the \textbf{data space} on which we aim to learn.
    A \textbf{general learning problem} on such a product space is a pair $(\mc L, \mc C)$, where $\mc L$ denotes the learning context and $\mc C$ specifies the learning criterion. Specifically, the \textbf{learning context} is defined as $\mc L = (\ell , \mc H, P)$, where:
    \begin{itemize}
        \item $\ell\colon \mc P (A) \tm \Theta \to \bb R$ is a \textbf{loss function} in $L^0(\mc P (A) \tm \Theta, \bb R)$,
        \item $\mc H \subseteq \mc M (O,A)$ is a decision class, or \textbf{model class},
        \item and $P \coloneqq \pi_\theta \tm E$ is the \textbf{data-generating probability distribution}, with associated experiment $E\in \mc M(\Theta,O)$ modeling the stochastic observations, and prior distribution $\pi_\theta \in \mc P (\Theta)$ modeling the likelihood of the parameters to manifest. 
    \end{itemize} 
    The learning criterion $\mc C$ is chosen by the decision maker to evaluate their overall performance against Nature.
\end{definition}

We remark that many different choices of $\mc C$ are available, some of them better studied than others. Popular examples include expected risk and minimax risk.
We refrain to state a preference in this section, and defer it to the next one.

\subsubsection{Supervised Learning through Risk Minimization} 
In the specific setup of \emph{supervised learning}, the observation space $O$ is the attribute space $X \subset \bb R^d \, ,d \ge 1$, while both states $\Theta$ and actions $A$ correspond to the label space $Y$. 
Then, the experiment $E\colon Y \karrow X$ identifies a probability associated with the attribute $X$, given the state $Y$. 
Here we focus on the classification task that assumes the label space to be \emph{finite}, while no constraint is imposed on $X$ apart from being a compact subset of $\bb R^d$. 
We define here the formal framework for this setting, first introducing some useful additional notions, and then formally defining Bayes risk and a supervised learning problem.

\begin{definition}
\label{def:posterior}
    A \textbf{posterior kernel} for a data space $(\XY, \mc X \tm \mc Y)$ is a Markov kernel $F \in \mc M (X , Y)$ from the label space $(Y, \mc Y)$ to the attribute space $(X, \mc X)$.
\end{definition}

Notice that, as we already noticed when introducing Markov kernels, the product operation can be used to write the Bayes decomposition theorem for a join probability $P \in \mc P(\XY)$. 
Hence, given $F \in \mc M(X, Y)$ and $E\in \mc M(Y, X)$, we can write 
\begin{align*}
    P &= \pi_X \tm F = \pi_Y \tm E ,
\end{align*}
for some suitable marginal probabilities $\pi_X \in \mc P(X)$ and $\pi_Y\in \mc P(Y)$.\footnote{To be precise, $\pi_X \tm F \in \mc P(\XY)$ and $\pi_Y \tm E \in \mc P(Y \tm X)$ by \cref{p:prod}. The equality of the two decomposition holds up to a permutation. However, we avoid this level of technicality to keep the presentation clear and simple.}

\begin{definition}
    Let $(\XY, \mc X \tm \mc Y)$ be a data space.
    Given a loss $\ell \in L^0({\mc P}(Y)\tm Y, \bb R_{\ge 0})$, a model class $\mc H \subseteq \mc M (X,Y)$ and a joint probability distribution $P \coloneqq \pi_Y \tm E \in \mc P (\XY)$, with $\pi_Y \in \mc P(Y)$ and $E\in \mc M (Y,X)$. Then, the \textbf{Bayes risk} {\normalfont (\gls*{br})} is defined as
    \begin{align*}
        \br_{\ell, \mc H}(\pi_Y \tm E ) &\coloneqq \inf_{h \in \mc H } \risk_{\pi_Y \tm E}(\ell \circ h) \,, \\
        \risk_{ \pi_Y \tm E }(\ell \circ h ) &\coloneqq \, \bb E_{\rv Y \sim \pi_Y} \bb E_{\rv{X} \sim E_{\rv{Y}}} \ell(h_{\rv X}, \rv Y) \;. 
    \end{align*}
    Here, $\risk$ is known as the \textbf{risk}; the notation $h_{\rv X}$ and $E_{\rv Y}$ denotes evaluation of a kernel (such as $h$ or $E$) at a random variable (such as $\rv X$ or $\rv Y$) and will be used consistently throughout.
\end{definition}

\begin{definition} \label[definition]{def:sup-learn-prob}
    Let $(\XY, \mc X \tm \mc Y)$ be a data space, consisting of a finite set of labels $Y$ and a set of continuous attributes $X \subset \bb R^d$. A \textbf{supervised learning problem} on $(\XY, \mc X \tm \mc Y)$ is a general learning problem, where:
    \begin{itemize}
        \item the loss is $\ell \in L^0({\mc P}(Y)\tm Y, \bb R_{\ge 0})$,
        \item the model class is $\mc H \subseteq \mc M(X,Y)$,
        \item and the data-generating distribution is $P \coloneqq \pi_Y \tm E \in \mc P (\XY)$ with $\pi_Y \in \mc P(Y)$ and $E \in \mc M(Y,X)$.
    \end{itemize}  
    In this setting, the learning criterion is by default risk minimization, \ie, finding the optimal action $h \in \mc H$ that achieves the associated \emph{Bayes risk}.
    Thus, we refer to a supervised learning problem simply using $\mc L =(\ell, \mc H, P)$, with $\mc C$ implicitly given by risk minimization.
\end{definition}

The definition above fits in the general learning problem framework by considering the specific diagram
\begin{tikzcd}[arrows={decorate, decoration={snake,segment length = 1.5mm,amplitude=.2mm}}, column sep=.5cm]
Y
\arrow[r, "E"]
& X 
\arrow[r, "h"]  
& Y,
\end{tikzcd} 
where $h$ is a decision rule chosen in $\mc H$, therefore choosing a probability on $Y$ associated to a point in $X$.%
\footnote{This is, considering the hypothesis, or decision, as stochastic. Several techniques exist to obtain a deterministic labels from a stochastic decision rule, with different consequences \citep{cotter2019making}.}

For some cases, the formulation of learning problem in terms of the experiment, or loss and model class, can be restrictive; for this reason, we introduce some alternative ways of writing $\mc L$.  
This can be naturally justified by considering the following simple proposition.

\begin{proposition} 
    \label[proposition]{prop:notations-learning-problem}
     A supervised learning problem $\mc L = (\ell , \mc H, P = \pi_Y \tm E)$ on the data space $(\XY, \mc X \tm \mc Y)$ can be equivalently expressed
     \begin{enumerate}
        \item using the minimization set $\ell \circ \mc H \coloneqq \{ (x,y) \mapsto \ell (h(x), y) \,\mid\, h \in \mc H \},$
        \ie, as a couple $\mc L = (\mc F \coloneqq \ell \circ \mc H, P)$.
        \item using the posterior kernel, \ie, as $P = \pi_{X} \tm F$ for some prior $\pi_{X} \in \mc P (X)$ on the attribute space. 
        We will then refer to it as $\mc L = (\ell , \mc H, P =  \pi_{X} \tm F)$ on  $(\XY, \mc X \tm \mc Y)$;
        \item using the joint distribution $P$, \textbf{agnostic} regarding its factorization. 
        We will then refer to it as $\mc L = (\ell , \mc H, P)$ on the general data space space $(Z, \mc Z)$.
     \end{enumerate}
     We refer to $\mc L = (\ell , \mc H, P = \pi_Y \tm E)$ as \textbf{generative}, while to $\mc L = (\ell , \mc H, P =  \pi_{X} \tm F)$ as \textbf{discriminative}.
\end{proposition}

We remark that, in the literature, when $\mc H = \mc M (X,Y)$, we talk about unconstrained learning problem and unconstrained Bayes risk.
Since our focus in the following will exclusively be on constrained supervised learning problems, we will simply term them \emph{learning problems} and never make use of the other cases.

% By means of these two views of  learning problem, we can define two \gls*{cbr} through the following equalities:  
% \begin{align}
%         \text{Discriminative:} \ \  \bb E_{\rv X\sim\pi_{X}} \cbr_{\ell,\mc{H}} (F_{\rv X}) &= \bb E_{\rv X\sim\pi_{X}} \inf_{h_{\rv X} \in \mc{H}_{\rv X}}  \bb{E}_{\rv Y\sim F_{\rv X}}  \ell(h_{\rv X}, \rv Y) = \br_{\ell, \mc H}(P) \;,\label{eq:x-cbr}\\
%         \text{Generative:} \ \ \bb E_{\rv Y\sim\pi_{Y}}  \cbr_{\ell,\mc{H}} (E_{\rv Y}) &= \bb E_{\rv Y\sim\pi_{Y}} \inf_{h \in \mc{H}}  \bb{E}_{\rv X\sim E_{\rv Y}}  \ell(h_{\rv X}, \rv Y) = \br_{\ell, \mc H}(P) \notag \;,
% \end{align} 
% with the Generative version by definition equal to the constrained \gls*{br}, while for \cref{eq:x-cbr} we need to enforce the condition of at least one of the minima of the unconstrained \gls*{br} to be included in $\mc{H}$. 
% This assumption underlies all the non-corrupted learning problem we will talk about in the following.
% For our convenience, we ask the minima considered to be the $h$ \emph{matching} the $F$, the true posterior.

\section{A Taxonomy of Corruptions in Supervised Learning}
\label[section]{sec:taxonomy}
In this section, we formally define our conceptualization of corruption within the context of a learning problem, utilizing the mathematical tool of Markov kernels. 
Given the diverse forms corruptions can take, we categorize them through a novel taxonomy based on their input and output spaces--essentially classifying them by \emph{type}.
We demonstrate the exhaustiveness of this general framework, which facilitates the systematic study of the various corruption types and combinations. 
Finally, through a careful examination, we analyze the relationships between our taxonomy of corruption and existing corruption models, elucidating novel insights generated by our framework.

%=========================================

\subsection{Corruption Definition and Types}

We have defined in the previous section that a learning problem comprises three key components: the loss function $\ell$, the model class $\mc H$, and the probability distribution $P$ from which we draw the data. We now turn our attention to how such a problem may be corrupted.
%In the field of machine learning, considerable attention has been devoted by engineers and researchers to the task of designing suitable loss functions or model architectures; however, less effort has been put into data, given that they are often not responsible for collecting them but rather for processing them \citep{sambasivan2021everyone}. 

In contrast to the traditional concept of corruption in machine learning, which only focuses on data generation and is defined as \emph{distribution shift}--an alteration of the probability distribution to deviate from its original test counterpart--we argue that corruption can occur in any of the components. 
In this broader sense, opting for \emph{surrogate losses} can be regarded as a form of corruption to the original loss function. For instance, surrogate losses are often chosen in place of the 0-1 loss in the classification problems \citep{bartlett2006convexity}. 
Moreover, a \emph{misspecified model}, such as when the model class of choice, \eg, linear functions, does not include the true model, \eg, a quadratic function, can also be considered a form of corruption.
Therefore we define the general corruption as any alterations in $(\ell, \mc H, P)$.

\begin{definition}
\label[definition]{df:gen-corruption}
Let $(Z, \mc{Z}), (Z', \mc{Z}')$ be data spaces. A \textbf{general corruption} is a mapping sending a learning problem $\mc{L} = (\ell, \mc H, P)$ into another learning problem $\t{\mc{L}} = (\t{\ell}, \t{\mc H}, \t{P})$, where $\mc L , \t{\mc L}$ are defined on $(Z, \mc{Z}), (Z', \mc{Z}')$ respectively.
\end{definition}

To initiate a comprehensive taxonomy of corruption, we begin by examining a specific case where corruption is defined as a Markov kernel with fixed input and output probability spaces.\footnote{While Markov kernels have been utilized in formalizing corruption \citep{van2017theory, williamson2022information}, their primary foci were solely on label corruption, attribute corruption, or simple joint corruption. 
} This definition subsumes a significant portion of existing literature, including classical works on distribution shift and noisy data. As such, our attention now turns to this subcase, formally defined below, with the aim of establishing connections between our types of corruption and the diverse corruption models laid out in previous studies. This, in turn, suggests that future work must extend beyond this subcase, as we have identified certain examples that are not covered by this definition (see \cref{subsec:relation}).

\begin{definition}
\label[definition]{df:corruption}
A \textbf{Markovian corruption} is a general corruption that maps a learning problem $\mc L = (\ell, \mc H, P)$ defined on $(Z, \mc{Z})$ into another learning problem $\t{\mc L}$ on $(Z', \mc{Z}')$ through the action of a Markov kernel $\ka \colon Z \karrow Z'$, \ie, such that $\t{\mc L} = (\ell, \mc H, \t{P} = P \circ \ka)$. Two important subcases are:
\begin{enumerate}
    \item A \textbf{joint Markovian corruption}, which has $Z = Z' = \XY, \mc{Z} = \mc{Z}' = \mc X \tm \mc Y$;
    \item A \textbf{partial Markovian corruption}, which is such that $Z, Z'$ can differ, and may be $X, Y$, or $\XY$, with the associated $\sigma$-algebras varying accordingly.
\end{enumerate}
\end{definition}
We will use the term ``corruption'' throughout the remainder of the paper as shorthand for ``Markovian corruption''.

We remark that the definition above does not necessarily assume the Markov kernel $\ka$ to be known. We only require $\ka$ to exist, and for us to know the values assumed by the kernel action when evaluated on $P$, \ie, $\t P = P \circ \ka$. 
The kernel is therefore not uniquely identified by the corruption definition, since multiple Markov kernels can generate $\t P$ from $P$.
However, for the analysis carried on in the rest of the paper, we assume $\ka$ to be known. 

The rationale behind this choice for modeling corruption lies in viewing a Markov kernel, or \emph{observation channel} \citep{csiszar1972class}, as a point-wise description of the stochastic process that leads to an observed probability distribution. 
This process is determined by external conditions that, in some sense, limit our ability of ``seeing'' the truth (probabilistic world/Nature), consequently giving rise to corruptions (distorted data distribution).
%Given the focus of this work \lc{on changes in the probability distribution instead of loss and model class}, we will from now on refer to Markovian corruption simply as \defemph{corruption} . 

For formal statements, we abuse the kernel notation and refer to the corruption induced by a kernel as the kernel itself, \ie,  $\ka \colon Z \karrow Z' $, or equivalently $\ka \in \mc M (Z,Z')$ for some suitable sets $Z, Z'$.

\begin{figure}[p]
    \centering
    \scalebox{.95}{\tikzset{fellipse/.style={rectangle, draw=white, fill=orange!15, align=center}}
\tikzset{dellipse/.style={rectangle, draw, dashed, align=center}}

\begin{tikzpicture}[edge from parent/.style={draw,-latex}, sibling distance=3cm]
\node[fellipse] (topnode) {\footnotesize $\displaystyle \ka \in \mc M(X \tm Y,X \tm Y) $ \\ \footnotesize{joint} } 
child { node[fellipse] {\footnotesize $\displaystyle \ka \in \mc M(X \tm Y , Y) $ \\ \footnotesize{2-dependent}  } 
    child { node[fellipse] {\footnotesize $\displaystyle \ka \in \mc M(X , Y)$ \\ \footnotesize{ 1-dependent} } } 
    child { node[fellipse] {\footnotesize $\displaystyle \ka \in \mc M(Y , Y)$ \\ \footnotesize{simple} } }
    }
child { node[dellipse] {\footnotesize $\displaystyle \ka \in \mc M( Y , X \tm Y )$ \\ \footnotesize{1-param. joint} } }
child { node[dellipse] {\footnotesize $\displaystyle \ka \in \mc M( X , X \tm Y ) $ \\ \footnotesize{1-param. joint} } }
child { node[fellipse] {\footnotesize $\displaystyle \ka \in \mc M(X \tm Y , X) $ \\ \footnotesize{2-dependent}} 
    child { node[fellipse] {\footnotesize $\displaystyle \ka \in \mc M(Y , X) $ \\ \footnotesize{1-dependent} } } 
    child { node[fellipse] {\footnotesize $\displaystyle \ka \in \mc M(X , X) $  \\ \footnotesize{simple} } } 
    }
;
\end{tikzpicture}}
    \vspace{-.5ex}
    \caption{\emph{Hierarchy of partial corruption types}. 
    The partial corruption types are hierarchically organized based on their dependence on the instance $X$ and label $Y$ space, as depicted through a tree structure. At the root of the tree lies the most general form of corruption, where the domain and image spaces are both the joint one, \ie, $D(\ka)=I(\ka)=\XY$. 
    The arrows signify that a child node has its domain or image constant \wrt exactly one of the variables in its parent. Therefore, the children nodes can be expressed as subcases of their parent, but the parents generally cannot be expressed by only one of their children. 
    The partial corruption types that cannot be combined with others are shown in dotted boxes. Note that corner cases involving independence from all variables or identity kernels are excluded from this analysis.
    }
    \label{fig:corruptions}
\end{figure}

\begin{figure}[p]
    \centering
    \scalebox{.95}{\tikzset{nstyle/.style={rectangle, draw=white, fill=orange!15, align=center}}

\begin{tikzpicture}[edge from parent/.style={draw,-latex},sibling distance=4cm, level distance=2cm]
\node[nstyle] (topnode) at (0,3) {\footnotesize $\tau \colon X \tm Y \karrow X$ \\ \footnotesize $\otm $ \\ \footnotesize $\lambda \colon X \tm Y \karrow Y$ } 
    child { node[nstyle] {\footnotesize $\tau \colon Y \karrow X$ \\ \footnotesize $\otm $ \\ \footnotesize $\lambda \colon X \tm Y \karrow Y$ } 
        %child { node[nstyle] { 1D-$\t{X}$ $\otm$ 1D-$\t{Y}$ } } 
    } 
    child { node[nstyle] {\footnotesize $\tau \colon X \tm Y \karrow X$ \\ \footnotesize $\otm $ \\ \footnotesize $\lambda \colon  X  \karrow Y$ } 
        %child { node[nstyle] { 1D-$\t{Y}$ $\otm$ 1D-$\t{X}$ } } 
    } 
    child { node[nstyle] {\footnotesize $ \tau \colon X \karrow X$ \\ \footnotesize $\otm $ \\ \footnotesize $\lambda \colon X \tm Y \karrow Y$ } }
    child { node[nstyle] {\footnotesize $ \tau \colon X \tm Y \karrow X$ \\ \footnotesize $\otm $ \\ \footnotesize $\lambda \colon Y \karrow Y$ } } 
;
\node[nstyle](bottomnode) at (-4,-1) { \footnotesize $\tau \colon Y \karrow X$ \\ \footnotesize $\otm$ \\ \footnotesize $\lambda \colon X \karrow Y$ } 
;
\foreach  \x in {1,2}{
\draw[latex-] (bottomnode) -- (topnode-\x);
}
\node[nstyle](bottomnode) at (4,-1) { \footnotesize $\tau \colon X \karrow X$ \\ \footnotesize $\otm$ \\ \footnotesize $\lambda \colon Y \karrow Y$ } 
;
\foreach \x in {3,4}{
\draw[latex-] (bottomnode) -- (topnode-\x);
}

\end{tikzpicture}}
    \vspace{-1ex}
    \caption{\emph{Feasible combinations of partial corruptions}. 
    Joint corruptions, \ie, of type $\ka \colon X \tm  Y \karrow \XY$, are obtained by combining two compatible partial corruptions in \cref{fig:corruptions}. The graph structure is induced by that of the partial corruption types. Notice that we can only combine a partial corruption with $I(\tau) = X$ with another such that $I(\lambda) = Y$, following \cref{prop:fact}. Therefore, the arrows signify that both $\tau$ and $\lambda$ in a child node inherit their domains from the parent node with either $\tau$ or $\lambda$ constant \wrt exactly one of their domain variables.
    }
    \label{fig:combinations}
\end{figure}

\subsubsection{A New Taxonomy of Partial Corruptions} 
Partial corruptions can be classified in different ways based on the domain and image of their associated kernels. Starting from the most general corruption, \ie, the joint corruption on $\XY$ induced by $\ka \colon \XY \karrow \XY$, when one space (either $X$ or $Y$ space) is absent in the image or domain of the corruption kernel, we obtain a partial corruption.
In \cref{fig:corruptions}, we present all possible types of partial corruption, with the exception of those that are identical (Dirac delta kernels) or trivial ($D(\ka)=\{*\}$) kernels, as they can be seen as obvious subcases of other partial corruptions.
By construction, we can state:
\begin{proposition}\label[proposition]{prop:partial-completeness}
    There are no partial Markovian corruptions outside of those listed in \cref{fig:corruptions}.
\end{proposition}
In the same \cref{fig:corruptions}, we classify partial corruptions based on their \emph{signature type}, that is, which sets of $X$ and $Y$ constitute the domain and image of the corruption kernel.
Specifically, we employ the following nomenclature: \emph{joint} corruption when $D(\ka)=I(\ka)=\XY$; \emph{1-parameter joint} corruption when $D(\ka)=\XY$ and $I(\ka)$ is either $X$ or $Y$; \emph{2-dependent} when $D(\ka)=\XY$ and $I(\ka)$ is either $X$ or $Y$; \emph{1-dependent} when $D(\ka)=X$ and $I(\ka)=Y$ or the opposite; \emph{simple} corruption when $D(\ka)=I(\ka)\neq \XY$, so when they are either equal to $X$ or $Y$. 

\subsubsection{Constructing Joint Corruption as a Combination of Partial Ones}
We now enumerate all possible ways of constructing joint corruptions, \ie, of the type $\ka \colon \XY \karrow \XY$, by combining the nodes in \cref{fig:corruptions} through the superposition operation \cref{p:super}. 
To this end, we introduce an additional condition to be imposed on the combinations of partial corruptions.

\begin{definition}\label[definition]{def:one-step}
    A Markovian corruption, induced by a kernel $\ka \in \mc M (\XY,\XY)$, is said to be a \textbf{one-step Markovian (joint) corruption} if $\ka$ is formed as a superposition of two partial corruptions, \ie, $\tau \otm \lambda$, such that neither $\tau$ nor $\lambda$ can be further decomposed using the operations defined in
    \cref{p:comp,p:prod,p:super,p:part-chain}.
\end{definition}
This definition is intentionally crafted to capture the most fundamental forms of composable corruption: those that occur in a single step, \ie, without further factorizing the kernels. By identifying and analyzing these atomic components, we establish a clear and comprehensive taxonomy of combined corruptions.

\begin{remark}
    It is also possible for a corruption kernel $\ka \in \mc M(\XY, \XY)$ to not respect the one-step condition in \cref{def:one-step}. 
    This case can occur in different ways: a first option is that $\tau$ or $\lambda$ are obtained through combination of other kernels; or, it can happen that $\tau$ and $\lambda$ are not combined via superposition; lastly, it can also happen that a factorization \wrt to \cref{p:comp,p:prod,p:super,p:part-chain} does not exist for $\ka$.
    We use a single umbrella term for all such kernels, called \textbf{non-factorized joint corruptions}, and treat them as a distinct class in our analysis.
    In fact, what we are enforcing is a non-factorized representation of all non-one-step Markovian corruptions.
    Later in this section, we will also give examples of what can fall within this set of corruptions. While a full characterization of these more general scenarios is beyond the scope of this work, we will observe that studying one-step Markovian corruptions gives us many insights also on this distinct set of corruptions.
\end{remark}

Using the requirements introduced until now to shape the objects of interest, we prove the following statement.

\begin{proposition}\label[proposition]{prop:fact}
The set of feasible one-step Markovian corruptions $\ka = \tau \otm \lambda$ is such that $I(\tau)=X$ and $I(\lambda)=Y$.
\end{proposition}
\begin{proof}
    According to \cref{df:corruption}, we must map a joint probability distribution on $\XY$ into another joint one. 
    Hence, we must exclude the combinations of a simple corruption with a 1-dependent corruption since such a pairing cannot generate a joint corruption.
    Additionally, combinations such as $(\tau \otm \lambda) (x, d\t x, d \t y) = \tau(x, d\t x, d \t y) \otm \lambda (x, d \t y) $ or of more than 2 kernels are not allowed because, according to \cref{p:super}, the measure on the corrupted labels (or in general, space) would be ill-defined.
    By taking $\lambda$ and $\tau$ from the partial corruption in \cref{fig:corruptions}, enumerating all of their possible combinations, and checking which of them are feasible, we can see that only the ones with $I(\tau)=X$ and $I(\lambda)=Y$ or $I(\tau)=Y$ and $I(\lambda)=X$ respect the condition. 
    Therefore, fixing the notation so that $\tau$ is  an attribute corruption and $\lambda$ a label corruption, we get the proposition.
\end{proof}

The set of feasible combinations is depicted and hierarchically organized in \cref{fig:combinations}. 
\cref{prop:fact} formalizes a desirable property of corruption, allowing it to change the distribution on attribute $\rv X$ and the distribution on label $\rv Y$ in a distinguishable way, either independently or dependently. Therefore, corruptions with indistinguishable effects on label and attributes, such as 1-parameter joint ones, are incorporated in the class of joint non-factorized corruption, \ie, $\mc M(X\tm Y, X\tm Y)$.\footnote{Note that a 1-parameter joint corruption can be seen as a subcase of a joint one, as $\ka(x, y, d\t x d\t y) = \lambda(x, d\t x d\t y) \, \mathbf{1}(y)$, where $\mathbf{1}(y)$ only trivially depends on $y$ since it is the matrix with all entries equal to 1.} In the next section we will see this is an appropriate choice.

Lastly, we can state the following {characterization} result as a direct consequence of \cref{prop:fact}.
\begin{corollary}\label[corollary]{cor:one-step-completeness}
    There are no one-step Markovian corruptions outside of those listed in \cref{fig:combinations}, and therefore it constitutes a \defemph{complete} taxonomy.
\end{corollary}

\subsubsection{A Practical Example} \label{sec:example}

Here, we present an illustration of a one-step Markovian corruption (in the finite case) within a practical scenario to facilitate for the reader's understanding of corruption through kernels.
The provided example is adapted from \citep{fogliato2020fairness} which considers the prediction of recidivism in the criminal justice system--predict who goes on to commit future crimes.

Surveys have shown that ``in the case of drug crimes, whites are at least as likely as blacks to sell or use drugs; yet blacks are more than twice as likely to be arrested for drug-related offenses'' \citep{rothwell2014war}.
Given this, we consider modeling the observed outcome ``rearrest'', denoted as $\t{\rv Y} \in Y \coloneqq \{+1, -1\}$, as a corrupted version of the true outcome ``reoffense'', denoted as $\rv Y \in Y \coloneqq \{+1, -1\}$, depending on the attribute $\rv X \in X = \{b, w\}$. 
Specifically, the disparity between $\rv Y$ and $\t{\rv Y}$ can be captured by a higher probability of flipping the reoffense label ($\rv{Y}=+1$) to the no rearrest label ($\t{\rv{Y}}=-1$) for white population ($\rv{X}=w$) compared to the black population ($\rv{X}=b$):
\begin{align*}
    \alpha(w) > \alpha(b), \; \text{where} \; \alpha(x)\coloneqq p(\t{\rv{Y}}=-1 \mid \rv{Y}=+1, \rv{X}=x),
\end{align*}
where $p$ is a conditional probability density.

Moreover, we assume that the corruption arises solely from the hidden recidivists, and not from erroneous arrests of individuals not committing the offense. In other words, the probability of flipping the no reoffense label ($\rv{Y}=-1$) to the rearrest label ($\t{\rv{Y}}=+1$) is zero for both the black and white populations:
\begin{align*}
    \beta(w) = \beta(b) = 0, \; \text{where} \; \beta(x)\coloneqq p(\t{\rv{Y}}=+1 \mid \rv{Y}=-1, \rv{X}=x).
\end{align*}

A possible Markov kernel modeling the setting would be therefore of the type $\lambda \colon \XY \karrow Y$, exemplifying 2-dependent label corruption. 
More specifically, it can be written as a joint kernel $\delta_X \otm \lambda$ where $\delta_X \colon X \karrow X$.
Being defined in discrete probability spaces, both $\delta_X$ and $\lambda$ can be expressed as matrices with entries representing conditional probabilities. In particular, for clarity we rewrite $\lambda$ as its parameterized version $\lambda_{\rv X} \big\rvert_{{\rv X}=x} \colon Y \karrow Y$ for $x \in X$, obtaining: 
\begin{align*}
    \delta_X &\coloneqq 
    \begin{bmatrix}
        p'(\t{\rv{X}}=b \mid \rv{X}=b) & p'(\t{\rv{X}}=b \mid \rv{X}=w) \\ 
        p'(\t{\rv{X}}=w \mid \rv{X}=b) & p'(\t{\rv{X}}=w \mid \rv{X}=w)          
    \end{bmatrix} =
    \begin{bmatrix}
         1 & 0 \\ 
         0 & 1          
    \end{bmatrix}  
    \,, \\
    \lambda_{\rv X} \big\rvert_{{\rv X}=x} &\coloneqq 
    \begin{bmatrix}
        p(\t{\rv{Y}}=+1 \mid \rv{Y}=+1, \rv{X}=x)  & 0 \\ 
        p(\t{\rv{Y}}=-1 \mid \rv{Y}=+1, \rv{X}=x) & 1          
    \end{bmatrix}
    =
    \begin{bmatrix}
        1-\alpha(x)   & 0 \\ 
        \alpha(x) & 1          
    \end{bmatrix} 
    \, ,
\end{align*}
where the entries are determined according to the given problem setting, and $p'$ is a conditional probability density.
To illustrate, consider an example of $\alpha(b)=1/10$ and $\alpha(w)=1/5$, yielding the following expressions:
\begin{align*}
    \lambda_{\rv X} \big\rvert_{{\rv X}=b} = 
    \begin{bmatrix}
        1-\alpha(b)   & 0 \\ 
        \alpha(b) & 1          
    \end{bmatrix} =
    \begin{bmatrix}
         9/10   & 0 \\ 
         1/10   & 1          
    \end{bmatrix}\, \text{and} \,
    \lambda_{\rv X} \big\rvert_{{\rv X}=w} = 
    \begin{bmatrix}
        1-\alpha(w)   & 0 \\ 
        \alpha(w) & 1          
    \end{bmatrix} =
    \begin{bmatrix}
        4/5   & 0 \\ 
        1/5   & 1           
    \end{bmatrix}\,.
\end{align*}
From \cref{df:corruption} we know that defining a Markovian corruption requires specifying a learning problem. In particular, it is necessary to fix the clean probability distribution for us to observe the effect of the corruption kernel on it.
Therefore, we additionally consider
$$q = [1/4, 1/4, 1/4, 1/4 ]^{\top} \in \mc P (\XY) \,,$$
where the specific order is assumed to be
$$q \coloneqq [\, q(\rv{X}=b,\rv{Y}=+1),\, q(\rv{X}=b, \rv{Y}=-1),\, q(\rv{X}=w, \rv{Y}=+1),\, q(\rv{X}=w, \rv{Y}=-1) \,]^{\top} \,.$$

Note that in finite spaces, the superposition operation \cref{p:super} reduces to the Kronecker product, hence we write the joint corruption kernel $\delta_X \otm \lambda \colon \XY \karrow \XY$ as
\begin{align*}
    \delta_X \otimes \lambda =
    \begin{bmatrix}
        \arraycolsep=1.5pt\def\arraystretch{2.3}
        \begin{array}{c|c}
            1 \cdot \lambda_{\rv X}\big\rvert_{\rv{X}=b} & 0 \cdot \lambda_{\rv X}\big\rvert_{\rv{X}=w} \\ 
            \hline
            0 \cdot \lambda_{\rv X}\big\rvert_{\rv{X}=b} & 1 \cdot \lambda_{\rv X}\big\rvert_{\rv{X}=w}
        \end{array}
    \end{bmatrix}
    = 
    \begin{bmatrix}
        \arraycolsep=1.5pt\def\arraystretch{2.3}
        \begin{array}{c|c}
            \lambda_{\rv X}\big\rvert_{\rv{X}=b} & \mb{0}  \\ 
            \hline
            \mb{0}  &  \lambda_{\rv X}\big\rvert_{\rv{X}=w}
        \end{array}
    \end{bmatrix} =
    \begin{bmatrix}
            9/10 & 0 & 0   & 0   \\
            1/10 & 1 & 0   & 0   \\ 
            0   & 0   & 4/5 & 0 \\ 
            0   & 0   & 1/5 & 1
    \end{bmatrix}
    \,,
\end{align*}
which is a $4 \tm 4$ block diagonal matrix. Written in its probabilistic form, the entries of the matrix representation would hence be the probability of $\t{\rv{X}}=\t x, \t{\rv{Y}}=\t y \mid \rv{X}=x, \rv{Y}=y$.
Then, we can obtain the \emph{a corrupted joint probability} $\t q \in \mc P (\XY)$ in the following manner: 
\begin{align*}
    \t q = q \circ (\delta_X \otimes \lambda) & = 
    \begin{bmatrix}
        9/10 & 0 & 0   & 0   \\
        1/10 & 1 & 0   & 0   \\ 
        0   & 0   & 4/5 & 0 \\ 
        0   & 0   & 1/5 & 1
    \end{bmatrix}
    \begin{bmatrix}
        1/4  \\
        1/4  \\ 
        1/4  \\ 
        1/4 
    \end{bmatrix} 
    =
    \f1{40} \cdot
    \begin{bmatrix}
        9  \\
        11  \\ 
        8  \\ 
        12 
    \end{bmatrix} \,,
\end{align*}
In this finite case, the chain composition operation \cref{p:comp} reduces to matrix multiplication. 
After applying the corruption kernel $\delta_X \otimes \lambda$, the original clean learning problem, characterized by the joint probability $q$, is transformed into a distinct, corrupted problem governed by $\t q$.

%This example underscores the need to distinguish between a Markov kernel, as defined in \cref{def:markov-kernel}, and a Markovian corruption associated with a specific learning problem $\mc L$, as defined in \cref{df:corruption}. The former is defined independently of any learning task, while the latter depends explicitly on a clean data distribution specified by the learning problem.
%As a proof of concept, we have only compared the original and corrupted probability distributions, without consideration of the learning aspect--finding the optimal decision \wrt the Bayes risk measure. 
%In the upcoming \cref{sec:brchange}, we present a systematic analysis of the consequences of different corruptions on a supervised learning problem by examining how their Bayes risk is changed, accompanied by discussions on strategies for mitigating these consequences in \cref{sec:losscorrect}. 

\subsubsection{On the Exhaustiveness of Markovian Corruption}

We now know from \cref{cor:one-step-completeness} that the taxonomy of one-step corruptions is complete. However, we also noticed that one-step corruptions are not the only possible type of corruptions, because kernels can also come in different forms.
By leveraging a well-known property of Markov kernels---their bijection with coupling of probability spaces---we define an \emph{exhaustive} taxonomy.

\begin{definition}
    We say that a taxonomy of Markovian corruptions is \textbf{exhaustive} if, for every fixed pair of distributions $(P, \t P) \in \mc P (Z) \tm \mc P (Z) $, there exists a Markovian corruption from $\mc L$ to $\t{\mc L}$ \st $\t P = \ka \circ P$ for all loss $\ell$ and model class $\mc H$.
\end{definition}

\begin{proposition}\label[proposition]{prop:exhaustiveness}
    The set of feasible one-step {Markovian} joint corruptions, illustrated in \cref{fig:combinations}, 
    together with the set of non-factorized ones, constitutes an exhaustive taxonomy of Markovian corruptions.
\end{proposition}

\begin{proof} 
    A coupling is formally defined for two probability spaces $ (Z_1, \mc Z_1, P_1), \, (Z_2, \mc Z_2, P_2)$ as a probability space $ (Z_1 \tm Z_2, \mc Z_1 \tm \mc Z_2, P)$, such that the marginal probabilities associated to $P$ \wrt $Z_i, \; i \in \{1,2\}$, are the respective $P_i$ \citep[Definition 1.1]{wang2012coupling}. 
    By construction, Markov kernels with fixed input and output probabilities $P, \t{P}$ are in bijection with all the possible couplings existent on $Z \tm Z$ with two \emph{fixed} probability measures; for us, $P, \t{P}$ (see details in \cref{sec:inversion}). This, by definition, proves the exhaustiveness of the taxonomy, as it implies that every corruption that is \emph{not} one-step Markovian, still sends $P$ into $\t{P}$, has a non-factorized representation.
\end{proof}

\subsection{Scope and Contributions of Our Taxonomy}

\label{subsec:relation}
\begin{table*}[t]
    \centering
    {\footnotesize
    \caption{Illustration of the taxonomy with examples of existing corruption models. When only one kernel is indicated, missing variables remain unchanged. ``dep.'' is short for ``dependent''. Details and references in \cref{sec:related}.} 
    \label{tab:corr-model-sec3}
    \begin{tabular}{lll} % Modified the preamble to include three columns
        \toprule
        \textbf{\makecell[l]{Corruption name \\ in literature}} & \textbf{Corruption type} & \textbf{Kernel representation}\\
        \midrule
        Attribute noise & simple & $\tau \colon X \karrow X $ \\
        \midrule
        Style transfer & 1-dep. & $\tau \colon Y \karrow X$ \\
        \midrule
        Adversarial noise  & 2-dep. & $\tau \colon X \tm Y \karrow X$ \\
        \midrule
        Random classification noise & simple & $ \lambda \colon \{*\} \karrow Y$ \\
        \midrule
        Class-conditional noise & simple & $ \lambda \colon Y \karrow Y$ \\
        \midrule
        Instance-dependent noise &  1-dep. & $\lambda \colon X \karrow Y$ \\
        \midrule
        \makecell[l]{Instance- \& label-dependent \\ noise} & 2-dep. & $\lambda \colon X \tm Y \karrow Y$ \\
        \midrule
        Combined simple noise  & two simple combined & $(\tau \colon X \karrow X) \otm (\lambda \colon Y \karrow Y)$ \\
        \midrule
        Generalized target shift & two 2-dep. combined & \makecell[l]{$(\tau \colon X \tm Y \karrow X) \otm (\lambda \colon X \tm Y \karrow Y)$} \\
        \midrule
        Target shift & \makecell[l]{min. simple,\\ max. 1-dep. $\&$ 2-dep. combined} & \makecell[l]{$ \lambda \colon Y \karrow Y$, \\ $(\tau \colon Y \karrow X) \otm (\lambda \colon X \tm Y \karrow Y)$} \\
        \midrule
        Concept shift & \makecell[l]{min. simple,\\ max. two 2-dep. combined} & \makecell[l]{$ \lambda \colon Y \karrow Y$, \\ $(\tau \colon X \tm Y \karrow X) \otm (\lambda \colon X \tm Y \karrow Y)$} \\
        \midrule
        \makecell[l]{Covariate shift  \\ Sampling shift } & \makecell[l]{min. simple, \\ max. 2-dep. $\&$ 1-dep. combined} & \makecell[l]{$ \tau \colon X \karrow X$, \\ ($\tau \colon X \tm Y \karrow X) \otm (\lambda \colon X \karrow Y)$}  \\
        \midrule
        Concept drift & \makecell[l]{can be any type, including \\ the non-factorized one} & - \\
        \midrule
        \makecell[l]{Mutually contaminated \\distributions \&\\ Selection bias (w. absolute\\continuity)} & non-Markovian corruption & - \\
        \bottomrule
    \end{tabular}
    }
\end{table*}

As presented, our taxonomy introduces a unified, kernel-based framework for understanding and organizing data corruption models in machine learning. 
It focuses on one-step Markovian corruptions, formally defined in \cref{def:one-step}, and is proven to be complete (by itself) and exhaustive (adding non-factorized ones). The scope is broad enough to encompass a large variety of existing models, which we organize hierarchically based on their dependence on the instance and label spaces.

A primary contribution of our framework is the reformulation of existing corruption models from \cref{tab:corr-model-intro}, which are aligned with our taxonomy as depicted in \cref{tab:corr-model-sec3}.\footnote{Details about the relationships with these corruption instances are given in \cref{sec:related}, and discussions on the relationships with other data corruption taxonomies are given in \cref{sec:other-tax}.}
Prior categorizations, \eg, \citep{quinonero2008dataset}, typically rely on invariance-based definitions (as shown in \cref{tab:corr-model-intro}), that is, specifying which parts of the data distribution remains unchanged. By design, they do not support a hierarchical or compositional interpretation of corruption. In contrast, our taxonomy enables a comparative view of corruption models by analyzing the domain and image of their associated kernels. This allows us to order corruptions by type complexity:
a kernel with more dependencies induces a more intricate corruption, while simpler corruptions emerge as subcases of more complex ones.\footnote{However, the concept of ``complexity'' of a corruption should not be interpreted as anything more than a structural complexity, given by its type. We are not proving here any quantitative complexity result, but organizing corruptions hierarchically for \emph{qualitative} comparison.}
Such a view provides a theoretical foundation for the systematic analysis and resolution of corruptions; an instance how that can be developed is proposed in Sections \cref{sec:brchange} and \cref{sec:losscorrect}.

As a second point, our work reveals that some corruption models in \cref{tab:corr-model-intro} correspond to multiple corruption types or combinations of partial corruptions in our taxonomy. This is because their definitions in \cref{tab:corr-model-intro} often consider the corruption of either the probability in the $X$ space or the $Y$ space, leaving freedom for corrupting the other. In extreme cases like concept drift, which assumes corruption only in the joint distribution, it can be of any corruption type and may not even be factorized to partial corruptions. By observing this, we gain the insight that some corruption models are way more general than others, and can even be regarded as taxonomies of their own.
This additionally translates into the impossibility of having a one-on-one correspondence between our taxonomy and existing ones. We elaborate further on this point in \cref{sec:other-tax}.

Third, our framework helps identifying corruption types that have been taken for granted to be probabilistic in nature, but, according to our definition of probabilistic corruption within the Markovian framework, are not. 
These include complex cases that are non-one-step (\cref{def:one-step}) or even non-Markovian (\cref{df:corruption}), and therefore require different treatment. We discuss examples of such corruptions below.

\subsubsection{Multi-Step Corruption}
\label[subsubsection]{sec:multi-step}

In machine learning research, the corruption process typically involves two environments---training and test time---but other settings are also possible.
For example, scenarios involving more than two spaces arise when learning from multiple domains \citep{bendavid2010theory}, or in the presence of concept drift over time \citep{widmer1996learning, gama2014survey, Lu_2018}.
By relaxing certain assumptions, the applicability of our framework can be extended into such cases by combining one-step Markovian corruptions.
In these cases, kernels act in a ``sequential'' (\cref{p:comp}-\cref{p:prod}) or ``parallel'' (\cref{p:super}) manner, enabling the modeling of more complex patterns of corruption.

\tikzset{mycircle/.style={align=center}}

\begin{figure}[t]
    \centering
    \includegraphics[width=0.8\linewidth]{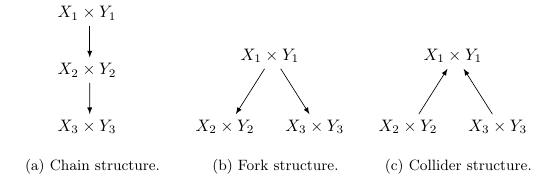}
    \caption{Possible non-degenerate relations among three probability spaces. Arrows represent a non-trivial Markov kernel $\kappa \colon X_i \times Y_i \rightsquigarrow X_j \times Y_j$.}
    \label{fig:3-env-corr}
\end{figure}

\textbf{Multi-Step Markovian Corruption.}
To illustrate how one-step corruptions give insights about non-one-step ones, consider a scenario with three environments $X_i \tm Y_i$, $i=1,2,3$. 
We aim to model Markovian corruptions occurring among these spaces and represent them using directed acyclic graphs, as usually done in causality literature \citep{pearl2016causal}.
In this representation, an arrow indicates the  possible presence of a non-trivial Markov kernel between two spaces, the absence of an arrow indicates that the two spaces must be related by a trivial kernel--in other words, they are independent.

We focus on three corruption configurations depicted in \cref{fig:3-env-corr}, excluding triangular structures with three arrows, as they lack a clear distinction between input and output spaces, making the corruption flow not interpretable.

The first configuration is the \emph{chain} structure shown in \cref{fig:3-env-corr}(a), where the spaces influence each other sequentially.
A concrete example is when $\ka_1 \colon X \karrow W$ models a feature extractor for the attribute space $X$, and $\ka_2 \colon W  \karrow X$ represents a corruption depending on the latent features only. 
Although this structure does not strictly satisfy our one-step corruption definition as per \cref{def:one-step}, it can still be represented in our framework by composing kernels as $\ka \coloneqq  \ka_1 \circ \ka_2 $, with each $\ka_i$ being a one-step Markovian corruption. 
Similar structures also occur in scenarios involving concept drift or online learning with corruption \citep{widmer1996learning, cesa2010online, Lu_2018}.

A second possibility is that the spaces relate according to the \emph{triangular} structures shown in \cref{fig:3-env-corr}(b) and (c). 
In particular, case (b) reflects assumptions made in settings combining data from different domains \citep{bendavid2010theory, van2017theory, redko2022survey}, where distinct observed distributions are assumed to arise from a common underlying clean distribution but corrupted differently depending on their environment. We can model this as a single Markov kernel obtained via superposition of other two, \ie, $\ka \coloneqq \ka_1 \otm \ka_2$, with $\ka_1 \in \mc M (X_1 \tm Y_1, X_2 \tm Y_2)$ and $\ka_2 \in \mc M (X_1 \tm Y_1, X_3 \tm Y_3)$. As for case (c), it can be used to model scenarios with merged (noisy or clean) datasets, which is a fairly common practice in robustness research, \eg, \citep{veit2017learning,fatras2022optimal}, as well as causality, \eg, \citep{gresele2022causal,garridomejia2024estimating}. The corruption would then be represented via $\ka \in \mc M(X_2 \tm Y_2 \tm X_3 \tm Y_3, X_1 \tm Y_1)$, with a decoupled joint probability as input, \eg, $P = P_2 \tm \ka_{P_3}$ where $\ka_{P_3}$ is a trivial kernel and $P_2,P_3$ are the clean underlying probabilities.

These three basic structures can be themselves combined to form more complex graphical models, capturing relationships among $n$ environments.
Since these are built from Markovian one-step corruptions or partial Markovian corruptions, our framework still offers insights into these more complex combinations, which motivates our focus on one-step corruptions.

\textbf{Multi-Step Non-Markovian Corruption.}
Consider two Markov kernels $\lambda \in \mc M (\XY \tm X, Y)$ and $\tau \in \mc M (X, X)$. Let $A$ be an element of the Borel sigma algebra on $\XY$. We can obtain a corrupted measure $\t P$ as 
\begin{align*} 
    \t{P}(A) &= (\t F \tm \t \pi)(A) \coloneqq [(P \circ_{\XY} \lambda) \tm (\pi \circ \tau)] (A) \\
    &= \int_{(\t x,\t y)\in A} [P \circ_{\XY} \lambda ] (\t{x},d\t{y}) \cdot [\pi \circ \tau] (d\t{x}) \\
    &= \int_{(\t x,\t y)\in A} \rup{ \int_{(x,y)\in \XY} \lambda(x,y,\t{x},d\t{y})\, P(dx,dy)} \cdot \rup{\int_{x'\in X} \tau(x',d\t{x}) \, \pi(dx')} \,,
\end{align*}
where $P$ is the joint clean probability and $\pi$ the associated clean $X$ marginal.
It is apparent that the domain space of $\lambda$ \defemph{does not respect the one-step assumption}: It assumes a coupling defining joint probability space on $(\XY\tm X)$, where the latter $X$ is meant to be equipped with the \emph{corrupted} marginal probability $\pi \circ \tau$, and $\XY$ has marginal $P$. This modeling choice amounts to a corruption that does not act in a ``parallel'' fashion on $X$ and $Y$, \ie, through $\otm$. The corruption of the label space happens on a second time step, depending on the outcome of an initial corruption phase carried out by $\tau$ and only involving $X$.

Notice that, in general, we cannot write the combined action of $\lambda$ and $\tau$ as a unique kernel $\ka$ acting on $P$; the available operations \cref{p:comp}, \cref{p:prod} and \cref{p:part-chain} do not permit this. Consequently, it cannot be expressed as a single graphical model, although each individual step can. However, we can write $\t{P}(A) = [\t{F} \tm \t{\pi}] (A) $, with $\t{\pi} \coloneqq \pi\circ\tau$ and $\t{F} \coloneqq P \circ_{\XY} \lambda$. This shows that the resulting corruption is \defemph{non-Markovian} in how it acts on the clean probability $P$, \emph{but has (almost) Markovian components.}%
\footnote{The ``almost'' refers to $\lambda$ acting on $P$ via partial chaining instead of standard chaining. This is very similar to our Markovian corruption definition, but it still is a slight relaxation.}

\subsubsection{Mutually Contaminated Distributions}

While being a term with less widespread recognition, \gls*{mcd} \citep{blanchard2014decontamination, menon2015learning, blanchard2016classification, katz2019decontamination} is a popular corruption model that has been studied in the literature under more familiar names, for example, \emph{learning from positive and unlabeled data} \citep{elkan2008learning, ward2009presence, du2014analysis, du2015convex, kiryo2017positive} in the binary class case, and \emph{learning from label proportions} \citep{quadrianto2008estimating, yu2014learning, liu2019learning, scott2020learning, tang2023multi}. Despite the popularity, less is understood about how \gls*{mcd} relates to other corruption models. Our framework offers new insights into such relationships and demonstrates how \gls*{mcd} extends beyond Markovian corruptions. 

To initiate this analysis, we first formally define \gls*{mcd} in the sense of \cref{tab:corr-model-intro}. Fix a measurable instance space $X$, and denote by $P$ a distribution over $X \tm [K]$ for $[K] \coloneqq \{1, 2, \cdots, K\}$ with random variables $(\rv X, \rv Y) \sim P$. %Let $k\in[K]$, $P_k \coloneqq \mathbb{P}(\rv X \mid \rv Y = k)$ be the class-conditional distribution, and $\pi_k \coloneqq P(\rv Y = k)$ be the base rate.

\begin{definition}[\citet{katz2019decontamination}] \label[definition]{df:mcd}
    Let $P(\rv X = x \mid \rv Y = k)$ be class-conditional distribution and $\pi_k \coloneqq P(\rv Y = k)$ be the base rate, both for $k\in [K]$.
    %Let $\mathbf{P}(x)=\big(P_k(x)\big)_{k\in [K]}$ be  clean class-conditional distributions and 
    Consider some mixing probabilities $\{\pi_{m,k}\}_{ m \in [M], k \in [K]}$, with $\pi_{m,k} \geq 0$ and $\sum_m \pi_{m,k} =1$.
    Then, \textbf{the} {\normalfont \gls*{mcd}} \textbf{corruption model} assumes that there is a general corruption from $(\ell,\mc H, P)$ to $(\ell,\mc H, \t P)$, such that the corrupted class-conditional distributions are of the form
    \begin{align*}
        \t P(\rv X = x \mid \rv Y = m) \coloneqq \sum_{k=1}^K \pi_{m,k} P(\rv X = x \mid \rv Y = k) \quad \forall x \in X\,,
    \end{align*}
    where $m \in [M]$ denotes the corrupted class.
\end{definition}

This definition has some clear differences with our definition of corruption. First, \cref{df:mcd} uses the class conditional probabilities $P(\rv X = x \mid \rv Y = k)$ instead of the joint probability. 
In our language, this means expressing corruption via the experiment $E$.
Secondly, their mixing probabilities defined a mixing matrix $\mathbf{\Pi} \coloneqq (\pi_{m,k})_{ m \in [M], k \in [K]}$ that is \emph{row-stochastic} instead of our column stochastic kernels.
We can therefore translate the \gls*{mcd} into our notation as in the following:
\begin{align}
    \t P (A) &= \sum_{Y}\int_A\int_X \delta_X(x,d\t x) \, \ka_M(\t{y},d{y}) \, E(y,dx) \, \t{\pi}(d\t y) \label{eq:mcd-corr} \\
    &= \int_A \left[ \big( \ka_M \circ (E \circ \delta_X) \big) \tm \t{\pi} \right] (d\t{x}, d\t{y})  \neq \int_A \left[ \big(\delta_X \otm \ka_M \big) \circ (\pi \tm E ) \right] (d\t{x}, d\t{y}) \,, \label{eq:mcd-compact}
\end{align}
where now $ \t P $ and $P$ are joint probabilities, $\ka_M(\t{y},d{y}) = \mathbf{\Pi}_{\t{y},y} \in \mc{M}(Y, Y)$, and $Y \coloneqq [\max(K, M)]$ to get a square matrix with added entries filled with zeros.
In particular, we remark that $\t{\pi}(d\t{y})$ is a marginal probability on the corrupted space, while ${\pi}(dy)$ is on the clean one. 
It is not specified by \citet{katz2019decontamination} how the corrupted marginal probability is obtained, nor whether it is the same as the one provided as input, \ie, as clean distribution.
Generally, we can always write the following relationship:
\begin{align}\label{eq:mdc-label-corr}
    \t{\pi}(d\t{y}) = \int_Y \lambda_M(\hat{y},d\t{y}) \, \pi({d\hat{y}}) \,,
\end{align}
where $\lambda_M \colon Y \karrow Y$, so the variable $\hat{y}$ is defined on the clean probability space $(Y, \mc Y, \pi)$, $\mc Y$ being a suitable $\sigma$-algebra.

\textbf{Mutually Contaminated Distributions Model is Non-Markovian.}
The formula derived above classifies \gls*{mcd} as multi-step, because plugging \cref{eq:mdc-label-corr} in \cref{eq:mcd-corr} violates \cref{def:one-step}. 
This gets even clearer when looking at \cref{eq:mcd-compact}: the right-hand side would imply that there exists a single kernel in $(\delta_X \otm \ka_M)(x,\t{y}, d\t{x}, dy) \in \mc{M}(\XY, X\tm Y)$ representing the \gls*{mcd} corruption scheme, but such representation is not possible because of how the \gls*{mcd} kernel acts on $E$ by definition. 
In addition, we remark that the existence of $(\delta_X \otm \ka_M)(x,\t{y}, d\t{x}, dy)$ would still \emph{not make a viable Markovian corruption} in the sense of \cref{df:corruption} because of the variables not being compatible with the probability $P$ for generating $\t P(d\t x, d\t y) = \big(P \circ (\delta_X \otm \ka_M)\big)(d\t x, d\t y) = \int_{\XY} P(dx, dy) \, \big(\delta_X \otm \ka_M \big)(x,\t{y}, d\t{x}, dy)$, the latter being an ill-posed integral since we have two measures on $Y$. 

\citet{menon2015learning} assume it plausible to have a corrupted label marginal totally unrelated to the original clean one; we model this case as a degenerate kernel constantly equal to the output probability, \ie, $\lambda_M(\hat{y},d\t{y}) = \t{\pi}(d\t{y})$.
The other extreme case is for the corrupted and clean marginals to not differ, and in such a case we are still in the presence of a corruption that is not one-step.
That because, having $\lambda_M(\hat{y},d\t{y}) = \delta_Y(\hat{y}, d\t{y})$ we write the marginal 
$$\pi(d\t{y}) = \int_Y \lambda_M(\hat{y},d\t{y}) \, \pi({d\hat{y}}) = \int_Y \delta_Y(\hat{y},d\t{y}) \, \pi({d\hat{y}}) \,.$$

\textbf{Comparison with Class Conditional Noise.} We can lastly look at the comparison of \gls*{mcd} with \gls*{ccn} to understand more in depth its non-Markovian nature. 
Clearly we cannot reduce \gls*{mcd} to \gls*{ccn}, as we have already shown in the above.
However, \citet{menon2015learning} prove in their Section 2.3 that \gls*{ccn} can be mapped to the \gls*{mcd} model in the binary case, and claim that \gls*{ccn} is a special case of \gls*{mcd}. Their argument can be trivially extended to multi-class setting by taking
$$[\ka_M]_{ij} \coloneqq \frac{[\lambda_C]_{ij}\t{\pi}_j}{\sum_{j=1}^{|Y|} [\lambda_C]_{ij}\t{\pi}_j} \,,$$
where $\lambda_C$ is the Markov kernel associated to \gls*{ccn}, and $\delta_X \otm \lambda_C$ would be its joint form.
In plain words, the usual definition of \gls*{ccn} via $\lambda_C$ can be manipulated such that the $\ka_M$ will subsume the label corruption, and the marginal corruption of the \gls*{mcd} is assumed to be a delta, \ie, $\lambda_M =\delta_Y$. However, it would still act on the clean probability as a non-Markovian corruption, as we have proved above. We therefore gain a new insight on \gls*{mcd}: It cannot be thought as a generalization of \gls*{ccn}; the two models can be equivalent in certain regimes of their parameters, but they are in general non-comparable noise models when written in their respective original definition.

\subsubsection{Selection Bias} \label{sec:selectionbias}
Another corruption model that has been widely studied in the literature is selection bias. Over the years, multiple definitions have been proposed, as we briefly discuss in \cref{sec:related} in comparison with covariate shift. 
We show that, under its classical formulation based on the Radon–Nikodym derivative, selection bias cannot be captured within the Markovian corruption framework. In contrast, under a probabilistic formulation, it does fall within this family. These two definitions are therefore incompatible and cannot be jointly assumed in a single theoretical analysis. This illustrates the utility of our kernel-based taxonomy, which classifies corruptions by their probabilistic nature and provides a unified basis for understanding them.
% We demonstrate in the following that selection bias cannot be subsumed by the Markovian corruption framework, when considered with its classical formulation with the Radon–Nikodym derivative. \lc{On the other hand, using a common probabilistic definition, it falls within the family of Markovian corruptions. We conclude that the two definitions are not compatible, and therefore cannot be assumed in the same theoretical analysis.}

\begin{definition}[Chapter 3.2, \citet{quinonero2008dataset}] \label[definition]{df:sbias}
    Let $Z \subseteq \bb R^d$ and the Borel $\sigma$-algebra $\mc Z$ on $Z$ form a measurable space. Consider a clean probability space $(Z, \mc Z, P)$ and a corrupted one $(Z, \mc Z, \t P)$, from which we aim to learn.
    We define \textbf{selection bias} as a general corruption such that $\mc L = (\ell, \mc H, P)$ and $\t{\mc L} = (\ell, \mc H, \t P)$, and that fulfills the following conditions:
    \begin{enumerate}
        \item Support condition, or absolute continuity of the measures:
            $\exists \ \alpha \in L^1(Z, \mc Z, P) $ \st $ \enspace \t P(A) = \int_A \alpha(z) P(dz) \enspace  \forall A \in \mc Z$,
            where $\alpha$ is a (almost surely unique) Radon–Nikodym derivative;
        \item Selection condition:
            $ \sup_{z\in Z} \alpha(z) < + \infty$.
    \end{enumerate}
\end{definition}

\textbf{Non-Markovian Definition of Selection Bias.} Clearly, selection bias can in principle include different instances of our taxonomy, since its type is not specified by its characterizing conditions.
We now try to understand if it meets the requirement for being Markovian in the first place. Comparing it with the action of a general $\ka \in \mc M(Z,Z)$ on the input probability $P$, we get the condition
$$\int_A \int_Z \ka(z, d\t z) P(dz) = \int_A \alpha(z) P(dz) \quad \forall A \in \mc Z\;.$$
It is easy to check that the kernel satisfying the condition is $\ka(z, d\t z) \coloneqq \delta_{z}(d\t z) \alpha(z)$, which respects the definition of kernel, but does not fulfill the Markov property, \ie, $\ka(z, Z)=1$ for all $z\in Z$, unless $\alpha(z) = 1 \ \forall z \in Z$.
This kernel is defined such that $P$ is corrupted into $\t P$, but it does not preserve mass for every input probability measure, therefore it is not what we are looking for to say that selection bias is a Markovian corruption. Is this $\ka$ the only possible guess?

Consider a general $\ka \in \mc M(Z,Z)$. It can be rewritten through its density \wrt a suitable measure, \ie,
$$\t P(A) = \int_A \int_Z \ka(z, d\t z)\, P(d z) = \int_A \int_Z k(z, \t z) \, \nu(d\t z) \, P(d z) \quad \forall A \in \mc Z \,,$$
and defining $\beta(\t z) \coloneqq \int_Z k(z, \t z) \, P(d z)$, we obtain
$\t P(A) = \int_A \beta(\t z) \, \nu(d\t z) \ \ \forall A \in \mc Z \,.$
Imposing $\ka$ to act as selection bias, we get $\beta(z) = \alpha(z) \ \forall z \in Z$ and $\nu = P \in \mc P (Z),\ \mu \coloneqq \f{\nu + P}2 $ -a.e. On the other hand, the Markov condition asks
$$\int_Z k(z, \t z) \, \nu(d\t z) = \int_Z k(z, \t z) \, P(d\t z)  = 1 \quad \forall z \in Z  \quad \Rightarrow \quad \beta(z) = \alpha(z) = 1 \quad \forall z \in Z\,.$$
Hence, we reached a contradiction and proved that \emph{selection bias cannot be directly represented as a Markov kernel} if we impose it to be acting on probabilities \emph{exactly} as the Radon–Nikodym derivative $\alpha$.
Obviously, there exists a Markovian corruption relating $P$ and $\t P$, since they are probability measures and our exhaustiveness argument holds. Therefore, we can represent selection bias via a Markovian corruption. However, that would not reflect the ``natural'' definition of selection bias that acts through the weighting function $\alpha$.

\textbf{Markovian Definition of Selection Bias.}
Interestingly, in the same Chapter 3.2 of \citet{quinonero2008dataset}, the authors also make use of the probabilistic definition of selection bias, as originally introduced by \citet{rubin1976inference}. More precisely, they define the probability $P_{tr}(\rv X=x,\rv Y=y\mid \rv S=1)$ as the one generating the training samples, where $\rv S$ is some Bernoulli selection variable that determines whether a data point is excluded or included (\ie, corrupted).
The test data are instead drawn from the uncorrupted distribution $P \in \mc P(\XY)$. In other words, they assume two joint probability spaces on $(\XY\tm \{0,1\})$: one clean, where $\rv S$ and the data are independent, and the marginal data distribution is $P$; one with dependence, and the data distribution is some $\t P$. It follows that these objects exist:

\begin{enumerate}
    \item $\pi \in \mc P(\{0,1\})$, the marginal of $\rv S$ \wrt the clean joint probability on $(\XY\tm \{0,1\})$;
    \item $\ka_{\pi} \colon \XY \karrow \{0,1\}$, a trivial kernel that assumes the value $\pi$ regardless of the point in $\XY$ it is evaluated on.%
    \footnote{This is equivalent to the definition of trivial kernel we gave in \cref{def:trivial-kernel}.} 
    This is the kernel description of the random sampling case, where $\rv S$ is independent of $(\rv X, \rv Y)$; 
    \item $\ka_{tr} \colon \{0,1\} \karrow \XY$, the kernel representation of the the biased training probability $P_{tr}$, \ie, 
    \begin{align*}
        \int_{A \subseteq \XY} \ka_{tr}(s=1,dx,dy)  = P_{tr}\Big((\rv X, \rv Y) \in A \,\mid\, \rv S =1\Big) 
    \end{align*} 
\end{enumerate}

Thus, we can write
\begin{align*}
    \t P(A) \coloneqq (P \circ \ka_{\pi} \circ \ka_{tr})(A)
     =:  (P \circ \ka)(A) ,
\end{align*}
for $A \subseteq \XY$. $\t P$ is the training distribution and $P$ the test one. This is therefore a Markovian representation of selection bias, and it is used as definition of the involved probability as if it was interchangeable with \cref{df:sbias}. 
Having proved that \cref{df:sbias} leads to a non-Markovian corruption, we know the two definitions of selection bias considered here are at least not using same ``representation'' for the corruption process. 
In addition, a Markovian corruption is generally not guaranteed to generate a corrupted distribution respecting the absolute continuity condition. 
This particular kernel $\ka = \ka_\pi \circ \ka_{tr}$ is defined through the degenerate kernel $\ka_\pi$, and therefore (the support of) $\t P$ does not at all depend on (the support of) $P$.
The corruption process is only dependent on $S$ and it relationship with $(\t X, \t Y)$ in a latent variable model.
We conclude that the two definitions are not equivalent, and should be used in conjunction only upon careful examination.

\section{Consequences of Corruption: Data Processing Equalities}
\label[section]{sec:brchange}
Having identified all the types of one-step Markovian corruptions, a natural subsequent question is how to systematically compare their effects. 
Recently, in \citep{williamson2022information}, Data Processing Equality results have been studied within the supervised learning framework and from an information-theoretic point of view. They have been inspired by the theory of comparison of statistical experiments \citep{blackwell1951comparison, torgersen1991comparison} and the information-theoretic Data Processing Inequality \citep{polyanskiy2025information}, which relates Bayes risk after and before corruption.
Their work adapts the theory to machine learning, including more realistic assumptions such as a restricted model class $\mc H$, a fixed loss of interest $\ell$ and a fixed prior distribution. Together with an experiment $E$, the fixed prior uniquely identifies the joint distribution $P \in \mc P(X\tm Y)$. 
More formally, their equalities are of the form 
$$\br_{{\ell \circ \mc H}}[\pi \tm \t E] = \br_{\widetilde{\ell \circ \mc H }}[ \pi \tm E]  \;,$$ 
where the notation $\widetilde{(\cdot)}$ indicates the action of some Markov kernel changing an object. 
In their framework, the quantity $\t E$ is a corrupted experiment in $\mc M (Y,X)$, which is computed either as $E \circ \tau\,, \tau \in \mc M(X,X)$ or as $\lambda \circ E\,, \lambda \in \mc M(Y,Y)$. 
We recall that the minimization set is defined as $\ell \circ \mc H \coloneqq \{\, (x,y) \mapsto \ell(h_x,y) \, | \, h \in \mc{H} \subseteq \mc M ( X, Y ) \, \} $ (as introduced in \cref{prop:notations-learning-problem}), and its corrupted counterpart $\widetilde{\ell \circ \mc H }$ is obtained as the action of the kernel $\tau$ or $\lambda$ on the set $\ell \circ \mc H$--the same kernel acting on $E$ and determining $\t E$. We will formally give this statement later in \cref{thm:br-s-s}.

The equality trivially induces an equivalence relation on the space of all possible learning problems, given the bijection between Markov kernels and couplings described in the previous section.
In its general form, we write it as
$$( {\ell \circ \mc H}, \pi \tm \t E)\brequiv (\widetilde{\ell \circ \mc H}, \pi \tm E) \;.$$

\citet{williamson2022information} only considered corruption acting on the sole experiment by composition, specifically they use what is referred to here as simple $X$ and $Y$ corruption.
In our contributions we also adopt the equality approach, but relate the clean and corrupted \emph{learning problems} through Bayes risk. 
We prove equivalences that formally characterize how problems are affected by different kinds of joint corruption; the kinds are identified by our taxonomy.
This class of results cannot be directly considered as part of the comparison of experiments theory, as we are not providing any inequality results and are considering a different setting. Rather, we complement that line of work by identifying when problems are equivalent in terms of appropriate measures of risk, and by analyzing how the structure of these equivalences changes depending on the type of corruption applied.

While the equality of Bayes risks is a direct consequence of formalizing corruptions via Markov kernels, we give explicit formulas describing how the corruptions act on prior and posterior distributions, as well as on the minimization set $\ell \circ \mc H$.
Accordingly, the main goal of this section is to present qualitative results in terms of conserved ``entropy''\footnote{For readers interested in how exactly Bayes risk relates to entropy, we refer to \citet{grunwald2004game,williamson2022information,polyanskiy2025information}.} between corrupted and clean learning problems, and establish a bridge between the problems themselves. 
These results also lay the basis for the loss-correction framework in Section \cref{sec:losscorrect}, as they enable a precise differentiation of the loss correction techniques based on the corruption type.

\subsection{Preliminary Properties}

When introducing Markov kernels in \cref{def:markov-kernel}, we allowed it to be defined on different input and output spaces. However, we also align with a more classical view of kernels as related to Markov chains, considering them as modification of the same set of objects while rearranging the probability measure defined on it. Hence, a corruption from $X\tm Y$ to $Y$ has to be considered as a \textit{parameterized version} of the corruption on $Y$, where the parameter is $x$. For this reason, we also introduced the operation \cref{p:part-chain}, which allows chaining while keeping a specified free parameter.
A degenerate sub-case takes place when we deal with a kernel from $X$ to $Y$.
We will make use of the notation $\ka_y$, $\ka_x$ to express the parameterization, which is a shortcut for $\ka_{\rv{Y}=y}=\ka_y$, $\ka_{\rv{X}=x}=\ka_x$ respectively.

Markov kernels prove themselves as useful modeling choice not only because of their interpretation and flexibility; they also have the property of preserving expectations under certain modifications. 
For this result to hold, one should assume a certain setting for the considered learning problems. A possibility is to take a more geometrical approach, \eg, the one described in Section 19.2 of \citet{aliprantis2006infinite}. Here we choose a less involved one to present---but very similar in the imposed restrictions--and introduce one key assumption on the loss function: we require it to be \emph{bounded}.
This requirement restricts the definition of loss function we gave in \cref{sec:learning}; positivity and boundedness together ensure the Fubini-Tonelli's theorem to be applied safely in the following proofs.
We can now give a formal statement of the aforementioned property of kernels.

\begin{lemma}[Data Processing Equality in Terms of Risk]
\label[lemma]{lemma:kernel-is-adjoint-to-risk}
Consider a bounded loss $\ell$, a model $h \in \mc M (X,Y)$, and a probability distribution $P$ on $(\XY, \mc X \tm \mc Y)$ standard Borel. Let $\ka$ be a joint corruption kernel in $\mc M (\XY, \XY)$. Then,
    \begin{align*}
        \risk_P [ \ka (\ell \circ h )] =  \risk_{P\circ\ka}  [\ell \circ h ].
    \end{align*}
\end{lemma}
\begin{proof}
    We know that $\risk_P [f] \coloneqq \int f dP $ and the kernel actions definition from \cref{sec:actions}. 
    Being the loss and kernel both bounded and positive, Fubini-Tonelli's theorem holds, and
    \begin{align*}
        \int( \ell \circ h) dP 
        &= \int_{(x,y)} \int_{(\t x,\t y)} \ka(x,y, d\t x, d\t y) \, \ell(h(\t x),\t y) \, P(dx, dy)  \\
        &= \int_{(\t x,\t y)} \ell(h(\t x),\t y) \int_{(x,y)} \ka(x,y, d\t x, d\t y) \, P(dx, dy)  \\
        &= \int_{(\t x,\t y)} \ell(h(\t x),\t y)  ( P \circ \ka )(d\t x, d\t y) .
    \end{align*}
\end{proof}
This result is central to establishing the Data Processing Equality for constrained Bayes risk in \cref{subsection:novel-dpe}. What remains unclear is how the \emph{form} of equality changes, specifically in terms of the clean distribution $P$ and the minimization set of $\ell\circ\mc H$, under different types of corruption.

Lastly, we specify that in all the following statements the joint corruption action on the learning problem is written as the superposition $\tau \otm \lambda$, where $I(\tau) = X$ and $I(\lambda) = Y$. Their full signature will be provided in each theorem.
Also, we use the notation $\ka \mc F \coloneqq \{ \ka f, \ \forall \, f \in \mc F \}$ for the action of a kernel on a compatible set of functions.

\begin{remark}
    The following section will focus on Bayes risk because of its connection with information theory:
    The \gls*{br} equalities (or, Data Processing Equalities) prove changes in the entropy measure induced by some ``learnable information'', \ie, the maximum amount of information contained in some distribution \wrt a learning problem. 
    However, \cref{lemma:kernel-is-adjoint-to-risk} shows that the results are valid also for a sub-optimal hypothesis in $\mc{H}$. The equivalences proved in the following sections can also be seen as risk induced ones (opposed to Bayes risk ones).
\end{remark}

%=======================================

\subsection{Existing Result: Data Processing Equalities for Simple Corruption}
As a first step, we can show that our framework subsumes the existing result proved by \citet{williamson2022information}. 
We give our own proof in \cref{sec:proof4}.
In our taxonomy, their ``combined noise'' corresponds to our ``combined simple corruption''. 

\begin{proposition}[Combined simple noise,  \citet{williamson2022information}]
\label[proposition]{thm:br-s-s}
Let $\ell$ be a bounded loss function.
Consider the clean learning problem $(\ell, \mc{H}, P)$, $E\colon Y \karrow X$ its associated experiment such that $P = \pi_Y \tm E$ for a suitable $\pi_Y$, and $F\colon X \karrow Y$ its associated posterior such that $P = \pi_X \tm F$ for a suitable $\pi_X$.
Let $\cemph{ (\tau\colon X\karrow X) \otm (\lambda\colon Y \karrow Y)}$ be a corruption acting on this problem. Then, the kernel action on $P$ can be rewritten as
\be \label{eq:br-s-s}
 P \circ (\cemph{\tau} \otm \cemph{\lambda})  = ( \pi_Y \tm E) \circ (\cemph{\tau} \otm \cemph{\lambda})
 = \pi_Y \circ \big[ (E \circ \cemph{\tau}) \otm \cemph{\lambda} \big] 
\ee
or, equivalently,
\be \label{eq:br-s-s-F}
   P \circ (\cemph{\tau}\otm\cemph{\lambda}) = ( \pi_X \tm F ) \circ (\cemph{\tau}\otm\cemph{\lambda})
    = \pi_X \circ \big[ \cemph{\tau} \otm (F \circ \cemph{\lambda} )  \big]  .
\ee
Then, the \gls*{br} Data Processing Equality, 
\be \label{eq:br-s-s-equivalence}
  \Big( \ell \circ \mc{H},  P \circ (\cemph{\tau} \otm \cemph{\lambda})  \Big) \brequiv
  \Big( \cemph{\tau}(\cemph{\lambda}\ell \circ \mc{H} ) , P \Big) , \nonumber
\ee
holds such that the functions contained in the new minimization set are defined as
$$ \tau (\lambda\ell \circ \mc{H}) \coloneqq \{ (x, y) \mapsto [ \tau (\lambda\ell_{y} \circ h) ](x) , h \in \mc H \} .$$
\end{proposition}

\textbf{A Fundamental Difference Between Label and Attribute Corruption.}
Starting from the simpler result in \cref{thm:br-s-s}, we can easily observe some properties of corruption that are conserved also in the next cases.
Consider its hypotheses, \ie, $ \tau \otm \lambda = \delta_X \otm \lambda$ being a corruption acting only on labels. The formula in \cref{eq:br-s-s-F} therefore tells:
$${\br_{\ell \circ \mc{H}}\big[ \pi_X \circ \big(  \delta_X \otm (F \circ \cemph{\lambda} )  \big) \big]= \br_{(\cemph{\lambda\ell} \circ \mc{H}) }  \big[ \pi_X \tm F \big] } .$$
When looking at the right-hand side, we see that the $\lambda$ component only modifies the loss function, and leaves the model class untouched.
On the other hand, when considering $ \tau \otm \lambda = \tau \otm \delta_Y$, we obtain
$${\br_{\ell \circ \mc{H}}\big[ \pi_Y \circ \big( ( E \circ \cemph{\tau} ) \otm \delta_Y \big) \big]= \br_{\cemph{\tau(\ell \circ \mc{H})} }  \big[\pi_Y \tm E \big] } \;,$$
and in this case notice that the action of $\ka = \tau \otm \lambda$ affects the whole minimization set when considering the Bayes risk on the clean distribution.
These simple label (Markovian) corruptions are \emph{equivalent in Bayes risk to loss corruptions}, which is non-Markovian in the sense of \cref{df:corruption}, although induced by a Markov kernel.

%=======================================

\subsection{Novel Data Processing Equalities for Other Corruptions}
\label{subsection:novel-dpe}

We now present the results for each of the remaining corruption combinations in \cref{fig:combinations}. From now on, we will refer to the following formula as \defemph{\gls*{br} Data Processing Equality}:
\begin{align}\label{eq:br-equiv-formula}
    \Big( \ell \circ \mc{H},  P \circ (\cemph{\tau} \otm \cemph{\lambda})  \Big) \brequiv
  \Big( \cemph{\tau}(\cemph{\lambda}\ell \circ \mc{H} ) , P \Big) .
\end{align}
The following two results are a strict generalization of \cref{thm:br-s-s}.

\begin{theorem}[2-dependent $\tau$, simple $\lambda$]
\label{thm:br-s-2dx}
Let $\ell$ be a bounded loss function.
Consider the learning problem $(\ell, \mc{H}, P)$ and suppose $E\colon Y \karrow X$ is its associated experiment such that $P = \pi_Y \tm E$ for a suitable $\pi_Y$. 
Let $\cemph{ (\tau\colon X \tm Y \karrow X) \otm (\lambda\colon Y \karrow Y)} $ be a corruption acting on this problem. Then,
\be 
P \circ (\cemph{\tau}\otm\cemph{\lambda}) = ( \pi_Y \tm E ) \circ (\cemph{\tau}\otm\cemph{\lambda}) =
 \pi_Y \circ \big[ ( E \circ_X \cemph{\tau}) \otm \cemph{\lambda} \big] 
 \nonumber
\ee
and the \gls*{br} Data Processing Equality in \cref{eq:br-equiv-formula} holds such that the functions contained in the new minimization set are defined as
$$ \tau (\lambda\ell \circ \mc{H}) \coloneqq \{ (x, y) \mapsto [ \tau (\lambda\ell_{y} \circ h) ](x,y) , h \in \mc H \} .$$
\end{theorem}

Here in \cref{thm:br-s-2dx} we have shown the \gls*{br} equality for the experiment $E$. % in line with the Comparison of Experiments \citep{torgersen1991comparison} and Information Equalities literature \citep{williamson2022information}.
However, for some corruptions the equalities cannot be stated with $E$ and the generative formulation of a learning problem, unless ignoring the joint corruption factorization formula. 
We hence use the posterior kernel $F$, \ie, the discriminative formulation of a learning problem, and gain more insights about the minimization set.

\begin{theorem}[simple $\tau$, 2-dependent $\lambda$]
\label{thm:br-s-2dy}
Let $\ell$ be a bounded loss function.
Consider the learning problem $(\ell, \mc{H}, P)$ and suppose $F\colon X \karrow Y$ is its associated posterior such that $P = \pi_X \tm F$ for a suitable $\pi_X$. 
Let $\cemph{(\tau\colon X  \karrow X) \otm (\lambda\colon X\tm Y \karrow Y)}$ be a corruption acting on this problem. Then, the kernel action on $P$ can be written as
\be 
    P \circ (\cemph{\tau}\otm\cemph{\lambda})  = ( \pi_X \tm F ) \circ (\cemph{\tau}\otm\cemph{\lambda}) =
    \pi_X \circ \big[ \cemph{\tau} \otm ( F \circ_Y \cemph{\lambda} ) \big] , \nonumber
\ee
and the \gls*{br} Data Processing Equality in \cref{eq:br-equiv-formula} holds such that the functions contained in the new minimization set are defined as
$$ \tau (\lambda\ell \circ \mc{H}) \coloneqq \{ (x, y) \mapsto [ \tau (\lambda\ell_{(x,y)} \circ h ) ](x) , h \in \mc H \} .$$ 
\end{theorem}

We can notice, thanks to \cref{thm:br-s-2dx,,thm:br-s-2dy}, that when corruption involves dependent structures in the factorization, the loss function or the whole minimization set are modified in a parameterized, \emph{dependent} way. 
Consider, for instance, the action of $\lambda\colon X \tm Y \karrow Y$ on the minimization set, when $\tau = \delta_X$. 
By definition, it generates the measurable functions
$$\cemph{\lambda \ell} \circ \mc{H}
= \{ (x,y) \mapsto (\cemph{\lambda \ell}) (h_x, \cemph{x, y})  \,|\, h \in \mc H\} = \{ (x,y) \mapsto (\cemph{\lambda\ell_{(x,y)}} \circ h )(x)\} ,$$ 
which is a strong change in the definition of the induced loss function $\t\ell\colon \mc P (Y)\tm X \tm Y \to \bb R_{\ge0}$ considered, although a still valid choice; for example, it has been employed in \citet{steinwart2008svm}.
We additionally underline here that \emph{corruptions on $Y$ only affect the loss function and do not touch the model class, even in the dependent case.}

The next theorems cover the factorizations involving 1-dependent corruptions. 
In the first case, we are again forced to use either $E$ or $F$, depending on the involved factors.
We group the two results into one theorem for brevity.

\begin{theorem}[1-dependent, 2-dependent]
\label{thm:br-1d-2d}
Let $\ell$ be a bounded loss function. Consider the clean learning problem $(\ell, \mc{H}, P)$, suppose $E\colon Y \karrow X$ is its associated experiment such that $P = \pi_Y \tm E$ for a suitable $\pi_Y$, and $F\colon X \karrow Y$ its associated posterior such that $P = \pi_X \tm F$ for a suitable $\pi_X$. 
\begin{enumerate}
    \item Let $\cemph{(\tau\colon Y \karrow X) \otm (\lambda\colon X \tm Y \karrow Y)}$ be a corruption acting on the problem. Then,
    \be
    P \circ ( \cemph{\tau}\otm \cemph{\lambda} )  = ( \pi_Y \tm E ) \circ ( \cemph{\tau}\otm \cemph{\lambda} ) =
    \pi_Y \circ \big[ \cemph{\tau} \otm ( E \circ_X \cemph{\lambda} ) \big] 
    \label{eq:br-1dx-2dy}
    \ee
    The \gls*{br} Data Processing Equality in \cref{eq:br-equiv-formula} holds such that the functions contained in the new minimization set are defined as
    $$ \tau (\lambda\ell \circ \mc{H}) \coloneqq \{ (x, y) \mapsto [ \tau (\lambda\ell_{(x,y)} \circ h ) ](y) , h \in \mc H \} .$$ 
    \item Let $\cemph{(\tau\colon X \tm Y \karrow X) \otm (\lambda\colon X \karrow Y)}$ be a corruption acting on the problem. Then,
    \be
    P \circ (\cemph{\tau} \otm \cemph{\lambda})  = (\pi_X \tm F ) \circ ( \cemph{\tau}\otm \cemph{\lambda} ) =
    \pi_X \circ \big[ ( F \circ_Y \cemph{\tau} ) \otm \cemph{\lambda}  \big] 
    \nonumber
    \ee 
    The \gls*{br} Data Processing Equality in \cref{eq:br-equiv-formula} holds such that the functions contained in the new minimization set are defined as
    $$ \tau (\lambda\ell \circ \mc{H}) \coloneqq \{ (x, y) \mapsto [ \tau (\lambda\ell_{x} \circ h ) ](x,y) , h \in \mc H \} .$$ 
\end{enumerate}
\end{theorem}

Since the 1-dependent $\ka$ and $\lambda$ combination is a subcase of both previous corruptions, we can prove the result as a simple corollary. Notice that this implies both $E$ and $F$ formulations to hold.

\begin{corollary}[1-dependent $\tau$, general $\lambda$]
\label[corollary]{thm:br-1d-1d}
Let $\ell$ be a bounded loss function. Consider the clean learning problem $(\ell, \mc{H}, P)$, $E\colon Y \karrow X$ its associated experiment such that $P = \pi_Y \tm E$ for a suitable $\pi_Y$, and $F\colon X \karrow Y$ its associated posterior such that $P = \pi_X \tm F$ for a suitable $\pi_X$. 
Let $\cemph{(\tau\colon Y \karrow X) \otm (\lambda\colon X \karrow Y)}$ be a corruption acting on the problem. Then, 
\be
    P \circ ( \cemph{\tau}\otm \cemph{\lambda} )  =   ( \pi_Y \tm E ) \circ ( \cemph{\tau}\otm \cemph{\lambda} ) =
    \pi_Y \circ \big[ \cemph{\tau} \otm ( E \circ \cemph{\lambda} ) \big] 
    \nonumber .
\ee
or, equivalently,
\be
    P \circ ( \cemph{\tau} \otm \cemph{\lambda} ) = (\pi_X \tm F ) \circ ( \cemph{\tau} \otm \cemph{\lambda} ) =
    \pi_X \circ \big[ ( F \circ \cemph{\tau} ) \otm \cemph{\lambda} \big] .
    \label{eq:br-1dy-2dx}
\ee 
The \gls*{br} Data Processing Equality in \cref{eq:br-equiv-formula} holds such that the functions contained in the new minimization set are defined as
    $$ \tau (\lambda\ell \circ \mc{H}) \coloneqq \{ (x, y) \mapsto [ \tau (\lambda\ell_{x} \circ h ) ](y) , h \in \mc H \} .$$ 
\end{corollary} 

In all the Theorems involving a 1-dependent corruption, the minimization set is heavily modified. 
To better understand how, we take a closer look at the functions contained in the clean and corrupted minimization sets.
First, we slightly rework the notation for the minimization set by considering the loss function $\ell(\cdot,y)$ as a parameterized one, \ie, $\ell_y \colon \mc P(Y) \to \bb R_{\ge 0}$; 
then, the set $\ell \circ \mc H \coloneqq \{\, (x,y) \mapsto \ell(h_x,y) \, | \, h \in \mc{H} \subseteq \mc M ( X, Y ) \, \} $ can be equivalently rewritten as $\{\, (x,y) \mapsto (\ell_y \circ h)(x) \, | \, h \in \mc{H} \subseteq \mc M ( X, Y ) \, \}$.

In \cref{eq:br-1dx-2dy}, we have again the kernel $\lambda \in \mc M (X \tm Y,Y)$ acting on the loss; hence, we obtain $\t{\ell}_{(x,y)} = \t{\ell}(\cdot ,x, y) \coloneqq (\lambda \ell)(\cdot ,x, y)$. Therefore, we are again inducing a more general notion of loss, namely $\t\ell\colon \mc P (Y)\tm X \tm Y \to \bb R_{\ge0}$.
Additionally, the whole composition with the model $h$, \ie, $\big( \t{\ell}_{(x,y)} \circ h \big)(\t x)$, is modified by the action of $\tau \in \mc M(Y, X)$, which ``swaps'' the input $\t x \in X$ with $y \in Y$ in addition to modifying the function itself.
Combining them together, we get the new minimization set containing functions of the form $f(x,y) = \big[ \tau \big( \t{\ell}_{(x,y)} \circ h \big) \big](y) $, which is not anymore comparable with the initial form $\ell_{\t y}\circ h(\t x)$, nor interpretable as a performance evaluation for the model $h$.

A similar strong modification is observed for the minimization set in \cref{eq:br-1dy-2dx}, which contains functions of the form $f(x,y) = \big[\tau\big( \t{\ell}_x \circ h \big)\big](y) \coloneqq \big[\tau\big( (\lambda\ell)_x \circ h \big)\big](y) $. {That is caused both by the action of $\lambda \in \mc M (X,Y)$ on $\ell(\cdot, y)$, which results in a new loss function $(\lambda\ell)_x(\cdot) \coloneqq \lambda\ell(\cdot, x)$, as well as the action of $\tau$ on $\ell \circ h$.}

The final result of the factorization, involving $\tau\colon X \tm Y \karrow X$ and $\lambda\colon X \tm Y \karrow Y$, yields a negative implication as detailed in the following.

\begin{theorem}[2-dependent $\ka$, general $\lambda$]
\label{thm:br-2-2}
    Let $\ell$ be a bounded loss function. Consider the clean learning problem $(\ell, \mc{H}, P)$, and let $\cemph{(\tau\colon X \tm Y \karrow X) \otm (\lambda\colon X \tm Y \karrow Y)}$ be a corruption acting on the problem. 
    Then:
    \begin{enumerate}
        \item the action of such corruption on the joint probability $P$ is equivalent to the one of the non-factorized joint corruption;
        \item the \gls*{br} Data Processing Equality in \cref{eq:br-equiv-formula} holds;
        \item the functions contained in the new minimization set are defined as
        $$ \tau (\lambda\ell \circ \mc{H}) \coloneqq \big\{ (x, y) \mapsto [ \tau (\lambda\ell_{(x,y)} \circ h ) ](x,y) , h \in \mc H \big\} .$$
    \end{enumerate}
\end{theorem}

This result is due to the full dependence on the joint space $X \tm Y$ for both $\tau$ and $\lambda$, making it impossible in general to derive a meaningful decomposition of the action on $P$ via \cref{p:comp}, \cref{p:prod} and \cref{p:part-chain}.
However, we can still distinguish the effect of $\lambda$ on the loss, as achieved in all previous cases, and of $\tau$ on the full minimization set.
For a detailed analysis and proof, see \cref{sec:proof4}.

\section{Loss-correction Approaches to Corruption Mitigation}
\label[section]{sec:losscorrect}
We now leverage our corruption framework and the derived Data Processing Equalities to reason about the fundamental question:

\begin{center}
    \emph{Can corruption be mitigated to guarantee accurate learning from corrupted data?}
\end{center}
In our chosen formalization, the training data are drawn from a corrupted distribution $\t P := P\circ\ka$, while evaluation is performed on data drawn from the original clean distribution $P$.
We define \defemph{accurate learning} as the property of a learning problem for which the optimal model, obtained minimizing risk over a constrained model class using corrupted training data, \emph{coincides} with the model obtained training on clean data with the same model class.

While prior work has already theoretically explored corruption mitigation, often by constructing corrected loss functions tailored to specific types of corruption, such methods are typically limited to a specific corruption model. 
To overcome this limitation, we introduce the notion of \defemph{Bayesian inverse} of a corruption kernel, which enables a principled and generalized form of loss correction that, in theory, applies systematically to all one-step Markovian corruptions captured in our taxonomy.%\footnote{We do not consider non-factorized corruptions, as they lead to trivial outcomes.}

\subsection{Existing Work on Corruption-corrected Learning for Label Corruption}

A vast amount of theoretical research on corruption-corrected learning has been carried in the fields of \emph{learning with noisy labels} and \emph{learning under distribution shift}.
Their goal has been to achieve \emph{unbiased learning} from biased data,\footnote{Existing literature defines unbiased learning via unbiasedness of the empirical risk estimator. This notion is trivially implied by our \cref{eq:ure}.} where ``biased data'' means corrupted data in our context, while ``unbiased'' refers to the use of a corrected loss $\tilde{\ell}$ that yields %an unbiased estimate of the clean risk:
\begin{align}
\label{eq:ure}
\risk_{\t P} [ \tilde{\ell} \circ h] =  \risk_{P}  [\ell \circ h ],
\end{align}
with $\risk_P [f] \coloneqq \int f dP $.
This equality can be obtained as a direct consequence of \cref{lemma:kernel-is-adjoint-to-risk} when an appropriate loss correction is applied.
When the Bayes risk is attained for some $h^* \in \mc{H}$, such unbiasedness directly implies accurate learning, formally written as 
\begin{align}
\label{eq:loss-correction-goal}
   h^* \in \argmin_{h\in\mc{H}}{\risk}_{\t P}[\tilde{\ell} \circ h]  \quad \text{and} \quad h^* \in \argmin_{h\in\mc{H}} \risk_{P}  [\ell \circ h ],
\end{align}
and therefore constitutes a stronger requirement.
We refer to the family of approaches for achieving the above goals collectively as \gls*{cl}.

Here, we examine two well-established approaches to unbiased learning through such loss correction, which serves as comparisons to our framework. We will then highlight their limitations and propose a generalized correction scheme grounded in Data Processing Equalities to address them. 

\subsubsection{Reconstruction-based Method}
The first method achieves loss correction by means of a \defemph{reconstruction matrix} \citep{van2017theory, patrini2017making, natarajan2013learning}, which is derived from a Markovian corruption $\lambda\colon Y \karrow Y$ assumed to be reconstructible.
In the finite space case, reconstructibility means that the corruption kernel matrix admits a left inverse $\lambda^*$ (so $\lambda^*\lambda=I$ on functions over $Y$).
The loss correction is then defined as 
\begin{align}
\label{eq:loss-correction-reconstruction}
    \t{\ell}(h_x,\t{y}) \coloneqq \lambda^*\ell(h_x,\tilde{y}) \coloneqq \sum_y \lambda^*_{\t y y}\ell(h_x,y), \enspace \lambda\colon Y \karrow Y,
\end{align}
where $\lambda^*$ is the reconstruction matrix and $\lambda^*\ell$ is the matrix acting on the loss vector, indexed by the label.
This yields unbiased learning as per \cref{eq:ure}, since for all $x$,
\begin{align*}
    \mathbb{E}_{\t{ \rv Y} \sim F \circ\lambda}[\tilde{\ell}(h_x,\t{ \rv Y})]
    &\overset{L\ref{lemma:kernel-is-adjoint-to-risk}}{=} \mathbb{E}_{{\rv Y} \sim F}\lambda(\tilde{\ell} \circ h_x) (\rv Y)\\
    &\overset{\eqref{eq:loss-correction-reconstruction}}{=}\mathbb{E}_{{\rv Y} \sim F}\lambda^*\lambda({\ell} \circ h_x)( \rv Y)\\
    &=\mathbb{E}_{{\rv Y} \sim F}[\ell(h_x,{ \rv Y})].
\end{align*}
This method is called \emph{backward correction} by \citet{patrini2017making}, and the \emph{method of unbiased estimators} by \citet{natarajan2013learning}.

We note that such a reconstruction matrix $\lambda^*$ is in general not a Markov kernel and may contain negative entries, which can make the corrected loss negative and cause problems for optimization.
Moreover, although the underlying framework in these works is similar to ours, they can only handle simple $Y$ corruption, \ie, $\delta_X \otm \lambda,\, \lambda\colon Y \karrow Y$, and do not generalize to more complex joint corruption cases.

\subsubsection{Importance-weighting-based Method}
A second line of work corrects loss functions through \gls*{iw} \citep{shimodaira2000improving,cortes2010learning,sugiyama2012machine}, originally developed for \emph{covariate shift}, where the input distribution is corrupted by $\tau$ such that $I(\tau)=X$ while the conditional distribution $F$ remains unchanged.
Under model misspecification \citep{white1981consequences},%
\footnote{By contrast, when the model is well specified, standard empirical risk minimization remains consistent under covariate shift and \gls*{iw} provides no benefit.}
\gls*{iw} provides a principled correction by requiring the clean data distribution to be absolutely continuous \wrt the corrupted one, \ie, $P \ll \t P$. The corrected loss takes the form
\begin{align}
\label{eq:loss-correction-iw}
   \t{\ell}(h(\mb{x}), {\t y}) \coloneqq w(\mb{x})\, \ell(h(\mb{x}), {\t y}), \; w(\mb{x}) \coloneqq \f{dP}{d\t P}(\mb{x}),
\end{align}
where $w(\mb{x})$ is called the importance weight, typically expressed via densities \wrt the Lebesgue measure.
This guarantees the unbiased learning goal in \cref{eq:ure}.

The key limitations of \gls*{iw} are that it applies directly only to covariate shift, and that it depends critically on {the assumptions of} absolute continuity as well as model misspecification.
Although recent work extends \gls*{iw} to joint corruptions by weighting with $w(x,y) \coloneqq \f{dP}{d\t P}(x,y)$ over the joint distribution \citep{liu2015classification,fang2020rethinking, fang2023generalizing}, these methods remain restricted by the absolute continuity requirement (at least on the overlapping support of $P$ and $\t P$), which is essential for the importance weights to be well-defined.

At first glance, the \gls*{iw} correction formula \cref{eq:loss-correction-iw} looks similar to the ones we derive later, however, it is not a subcase of our kernel-based loss correction.
\gls*{iw} operates by reweighting losses through the Radon-Nikodym derivatives $\f{dP}{d\t P}$, whereas our framework performs correction through kernel inversion.
The two mechanisms are fundamentally different, as Radon–Nikodym derivatives exist only under absolute continuity assumptions, while kernel-based loss corrections are free from such restrictions and can systematically handle general corruptions beyond that.
In addition, Radon–Nikodym derivatives cannot be represented as Markov kernels, as we have seen in \cref{sec:selectionbias}.
% \Nan{It is mentioned one is not a subcase of the other in the beginning. Maybe we can add this sentence in the middle of the paragraph?} \Laura{I think it works better here, with an added ``In addition''.}

\subsection{A Generalized Framework for Corruption-Corrected Learning}
In \gls*{cl}, existing approaches such as reconstruction-based or importance-weighting based methods are either designed for specific types of corruption, or rely on restrictive assumptions, and therefore cannot resolve the question posed above in a systematic or general way.
To this end, we introduce the concept of the Bayesian inverse of a Markov kernel, which enables a systematic analysis of \gls*{cl} across the wide range of corruptions in our taxonomy.

\subsubsection{Bayesian Inverse of a Markov Kernel}

To study the \gls*{cl} problem within our framework, we first define a principled way to reverse the corruption process.
We introduce here the Bayesian inverse of a Markov kernel \citep{dahlqvist2016bayesian,cho2019disintegration}, which preserves the Markov property and yields a mathematically well-defined mechanism for inverting corruptions.

\begin{definition}
\label[definition]{def:inv}
    Let $P\in\mc P(Z_1)$ be a probability distribution, and $\ka \in \mc M(Z_1, Z_2)$ be a Markov kernel with the property $P \circ \ka=\t P$ for some $\t P \in \mc P(Z_2)$. The \emph{Bayesian inverse} of $\ka$ is defined as the Markov kernel $\ka^\dag \in \mc M(Z_2, Z_1)$ such that it induces together with $\t P$ the same coupling induced by $\ka$ and $P$, \ie, 
    $$(P \tm \ka)(A \tm B) = (\t P \tm \ka^{\dag})(A \tm B), \ \forall \, A\tm B \in \mc Z_1 \tm \mc Z_2 .$$
\end{definition}
By taking $A$ or $B$ equal to the $Z_2$ or $Z_1$, we respectively get the property of $\ka^\dag$ and $\ka$ being a ``weak'' inverse of each other. 
\begin{proposition}
\label[proposition]{prop:binv-is-cleaning-kernel}
    Let $\ka \colon Z_1 \karrow Z_2$ be a Markov kernel with the property $\t P = P \circ \ka$ for $P \in \mc P (Z_1), \t P \in \mc P (Z_2)$, and $\ka^\dag$ its Bayesian inverse. Then, it reverses the action on the fixed input and output probabilities, \ie,
    $$P(A) = (\t P \circ \ka^{\dag})(A) , \ \forall \, A \in \mc Z_2.$$
\end{proposition}

\begin{remark}
    Recalling how we have defined the posterior kernel $F\in\mc M(X,Y)$ in \cref{def:posterior}, \ie, given $\pi_X \in \mc P(X)$ and $\pi_Y \in \mc P(Y)$,
    \begin{align}
        \pi_X \tm F = \pi_Y \tm E = P \in \mc P(\XY),
    \end{align}
    we notice that $F = E^\dag$.
\end{remark}

We will refer to the Bayesian inverse of the corruption kernel as the \defemph{cleaning kernel}.
In general, the Bayesian inverse is not unique, since it corresponds to a class of equivalence induced by the probability measures on $Z_1$ and $Z_2$.
However, we are always sure it exists given the assumption of using standard Borel measure spaces when defining Markov kernels (more details in \cref{sec:inversion}). This is a weaker existence condition \wrt the one given for the reconstruction matrix.%
\footnote{Such a  weaker condition comes at the price of the Bayesian inverse kernel being a \defemph{typed inverse}, meaning that to compute $\ka^\dag$, we need not only the kernel $\ka$ but also the initial probability $P$; having a different $P$ induces a different Bayesian inverse. This point becomes clearer when considering the discrete case, as explained in Remark~\ref{remark:bayesian-inverse-exists}.}

\begin{remark}
\label{remark:bayesian-inverse-exists}
    In the discrete case, the Bayesian inverse always exists and is defined by Bayes rule. 
    Furthermore, it is uniquely defined $\t P$-a.s. 
    This easy to see by unfolding \cref{def:inv} into:
    $$\int_B \ka(z_1, A)\, P(d z_2) = \int_A \ka^\dag(z_2, B)\, \t P(d z_2) \quad \forall A \in \mc Z_2, \, \forall B \in \mc Z_1.$$
    This formulation extends the discrete Bayes' rule $P(z_2 \mid z_1)P(z_1) = P(z_1 \mid z_2)P( z_2) \, {\forall z_1,  z_2 }$.
    Indeed, in the discrete case, the Bayesian inverse always exists and can be expressed as 
    $$\ka^\dag(z_1\mid z_2) \coloneqq \frac{P(z_1)\ka(z_2 \mid z_1)}{\t P(z_2)} \quad  \enspace \forall z_1,  z_2 \enspace  \text{\st} \enspace \t P( z_2) \neq 0.$$
    This formula ensures the uniqueness of $\ka^{\dag}$ within the support of $\t P$, as all components are unique. 
    However, outside the support when $\t P$ is zero, the uniqueness may not hold, requiring a non-fixed value for $ z_2 \in Z_2$ where $\t P(z_2) = 0$.
\end{remark}

The Bayesian inversion operation has the following desirable property of preserving the expectations.

\begin{proposition}[Inversed Data Processing Equality in Terms of Risk]
    \label[proposition]{prop:binv-exp-preserv}
    Consider a learning problem $(\ell,\mc H, P)$ on $(\XY, \mc X \tm \mc Y)$, a corruption $\ka \in \mc M (\XY, \XY)$, and a function $f\in \ell \circ \mc M (X, Y)$. Let $\ell$ be a bounded loss function. Then,
    \be \label{eq:dpe-risk-binv}
        \risk_{P} [f(\rv Z)] = \risk_{P \circ \ka} [\ka^{\dag}f(\t{\rv Z})] ,
    \ee
    and 
    \be  \label{eq:dpe-br-binv}
        \br_{\ell \circ \mc H}\big( P \big) = \br_{\ka^\dag(\ell \circ \mc H)}\big( P \circ \ka \big),
    \ee
    where $\ka^\dag$ is the cleaning kernel and $(\ka^\dag(\ell \circ \mc H), P \circ \ka)$ the corruption-corrected problem.
\end{proposition}
\begin{proof}
    The first claim follows by applying \cref{lemma:kernel-is-adjoint-to-risk} for $P \circ \ka \circ \ka^\dag = P$, where the equality holds because of \cref{prop:binv-is-cleaning-kernel}. The second is proved by taking the infimum over $\ell \circ \mc H$ of both sides of \cref{eq:dpe-risk-binv}.
\end{proof}

\begin{remark}\label[remark]{remark:factorization}
    Notice that \cref{prop:binv-exp-preserv} above does not imply $ \ell \circ h^* = \ka^\dag (\ell \circ h^*)$, but only that their risks coincide, \ie, $\bb E_P (\ell \circ h^*) = \bb E_{P\circ\ka}[\ka^\dag (\ell \circ h^*)]$. 
    Under certain conditions on the learning problem, there exists some $h^*, \t h^* \in \mc H$ such that $\ell \circ h^* = \ka^\dag (\ell \circ \t h^*)$, but it is not necessarily the case that $\t h^*=h^*$.
\end{remark}

In the following we will make use of these facts assuming $Z_1=Z_2=X \tm Y \eqqcolon Z$ to match the setting of our taxonomy of corruption.

\subsubsection{Label Corruption Correction with Bayesian Inverse}

Similarly to the reconstruction-matrix and importance-weighting approaches, we enforce the condition in \cref{eq:ure} and derive a principled loss correction method based on the Bayesian inverse $\ka^{\dag}$ of the corruption kernel $\ka$.

We first illustrate this in the case of simple label corruption $\lambda\colon Y \karrow Y$.
Consider its associated cleaning kernel sending $\t{\mc L}=(\t{\ell}, \mc H, \t P)$ to the clean one $\mc L = (\ell, \mc H, P)$, \ie, the Bayesian inverse $\lambda^{\dag} \colon Y \karrow Y$. 
By considering a corruption $\delta_X \otm \lambda$ and its inverse $\delta_X \otm \lambda^{\dag}$, \cref{prop:binv-exp-preserv} allows us to write

\begin{align*}
\risk_{\t P}(\lambda^{\dag} \ell \circ h) = \risk_{P}(\ell \circ h), \quad \text{where~} {\t P} := P \circ (\delta_X \otm \lambda), \quad \forall \, h \in \mc H \,.
\end{align*}
This directly yields the loss correction formula:
\begin{align}
\label{eq:loss-correction-bayesian}
    \t{\ell}(h_x,\t{y})\coloneqq \lambda^{\dag}\ell(h_x,\tilde{y}), \quad \lambda^{\dag}\colon Y \karrow Y.
\end{align}
The corrected loss satisfies the unbiased learning criterion in \cref{eq:ure}, and hence also achieves the accurate learning goal in \cref{eq:loss-correction-goal}.

Notably, in this simple label corruption case, our kernel-based correction in \cref{eq:loss-correction-bayesian} resembles the reconstruction-matrix correction in \cref{eq:loss-correction-reconstruction}. However, the two differ fundamentally, as illustrated below.

\begin{example}
Following \citet[Sec.~4.1.2]{van2017theory}, consider symmetric label corruption in binary classification setting where the clean distribution is given by $P = (p_1, p_2)^\top$. Suppose $\sigma\in(0,0.5)$;
let the corruption kernel be
\begin{align*}\lambda = 
    \begin{bmatrix}
        \sigma & 1-\sigma \\
        1-\sigma & \sigma  \\ 
    \end{bmatrix}.
\end{align*}
Its associated reconstruction matrix exists, and it amounts to
\begin{align*}\lambda^* = \f1{1-2\sigma} \cdot
    \begin{bmatrix}
        1-\sigma & -\sigma \\
        -\sigma & 1-\sigma  \\ 
    \end{bmatrix},
\end{align*}
while the Bayesian inverse takes the form
\begin{align*}\lambda^\dag =
    \begin{bmatrix}
        \f{p_1(1-\sigma)}{p_1(1-\sigma) + p_2\sigma}  & \f{p_1\sigma}{p_2(1-\sigma) + p_1\sigma} \\
        \f{p_2\sigma}{p_1(1-\sigma) + p_2\sigma} & \f{p_2(1-\sigma)}{p_2(1-\sigma) + p_1\sigma}  \\ 
    \end{bmatrix}.
\end{align*}
The two objects are clearly different, with the Bayesian inverse approach requiring the knowledge of the clean probability values in order to compute the cleaning matrix. 
\end{example}

We note that in simple corruption settings, such as $\lambda \colon Y \karrow Y$ as discussed above, the Bayesian inverse naturally takes the form $\lambda^\dagger: Y \karrow Y$. However, for a joint corruption kernel of the form $\ka = \tau \otm \lambda$, the Bayesian inverse does not, in general, distribute over $\otm$; that is, one typically has $\ka^\dagger \neq \tau^\dagger \otm \lambda^\dagger$. Consequently, in what follows, we only assume that $\ka^\dagger$ is \defemph{one-step Markovian} and therefore admits the form $\tau \otm \lambda$.

We now extend this paradigm to the $\lambda \colon X \tm Y \karrow Y$ case, as from \cref{thm:br-s-2dy} we see that also in the dependent case, label corruption only affects the loss function. 

\begin{theorem} \label{thm:losscorr-label}
    Let $( \ell, \mc H, P)$ be a clean learning problem with $\ell$ being a bounded loss function, and $\ka=\delta_X \otm \lambda$ a one-step Markovian corruption on it. Let $\ka^{\dag}$ be the cleaning kernel inverting $\ka$, such that $(\ka^{\dag}( \ell \circ \mc H), P \circ \ka)$ is its associated corrected problem.
    When $\lambda \in \mc M(Y,Y)$, we have 
    $$ \t{\ell}(h(\t x), \t y) \coloneqq (\lambda \ell)\, (h(\t x), \t y) \quad \forall \, (\t x, \t y) \in X \tm Y \,, $$
    while, when $\lambda \in \mc M(\XY,Y)$ we have a more general notion of loss, \ie,
    $$ \t{\ell}(h(\t x), \t x, \t y) \coloneqq (\lambda \ell)\, (h(\t x), \t x, \t y) \quad \forall \, (\t x, \t y) \in \XY \,, $$
    with $\ell \colon \mc P(Y) \tm \XY \to \bb R_{\ge 0}$.  Similarly, when $\lambda \in \mc M(X,Y)$,
    $$ \t{\ell}(h(\t x), \t x) \coloneqq (\lambda \ell)\, (h(\t x), \t x) \quad \forall \, (\t x, \t y) \in \XY \,, $$
    with $\ell \colon \mc P(Y) \tm X \to \bb R_{\ge 0}$.  
\end{theorem}

\begin{proof}
    Let us consider a general function in the set $\ka^\dag (\ell \circ \mc H)$. Assuming $\ka^{\dag}=\delta \otm \lambda$, $\delta(\t x, dx) \in \mc M(X,X)$, it will take the form
    \begin{align*}
        \ka^\dag (\ell \circ h)(\t x, \t y) = (\delta\otm\lambda)(\ell \circ h) (\t x, \t y) \coloneqq 
        \sum_{y \in Y} \int_{x \in X} \ell(h(x),y) \, \delta(\t x, dx)  \,\lambda(\t x, \t y, dy) \;. 
    \end{align*}
    
    Now let $\lambda$ be a 2-dependent label corruption, \ie, in $\mc M(\XY, Y)$. We can define the loss correction $\t\ell$ as
    \begin{align*}
        (h(\t{x}), \t x, \t y) \mapsto \t{\ell}(h(\t{x}), \t x, \t{y}) 
        &\coloneqq \sum_{y \in Y} \int_{x \in X} \ell(h(x),y) \, \delta( \t x, dx) \, \lambda( \t x, \t y, dy) \\
        &= \int_{x \in X} (\lambda\ell)(h( x),\t x, \t y)\, \delta(\t x, dx)   
        = (\lambda\ell)(h(\t x),\t x, \t y) ,
    \end{align*}
    where $(h(\t{x}), \t x, \t y) \in \mc P(Y) \tm X \tm Y$.
    Hence, the case $\lambda(\t x, \t y, dy)$ and its subcases $\lambda(\t x, dy)$, $\lambda(\t y, dy)$ combined with an identity kernel on $X$ do not change the hypothesis function. 
    %Notice also that by property of Markov kernels and their transitions, $\t \ell$ is still measurable, positive and bounded.
    Lastly, by definition of Markov transition, we have that when $\lambda \in \mc M(Y,Y)$ 
    \begin{align}
        \t{\ell}(h(\t{x}), \t{y}) &\coloneqq \sum_{y \in Y} \int_{x \in X} \ell(h(x),y) \, \delta( \t x, dx) \, \lambda(\t y, dy) 
        = (\lambda\ell)(h(\t x),\t y),
    \end{align}
    which respects the classical definition of loss function, as $(h(\t{x}), \t y) \in \mc P(Y) \tm Y$.
    This proves the thesis, as by construction $\t{\ell}(h(\t{x}), \t x, \t{y}) \in \ka^\dag (\ell \circ \mc H)$. 
    \cref{prop:binv-exp-preserv} implies that $\ka^\dag (\ell \circ \mc H)$ makes \cref{eq:ure} hold.
\end{proof}

For the dependent label corruption cases, we need to relax the definition of loss function, and allow it to take $x$ as an additional input. This generalization is not unprecedented; for example, it has been employed in \citet{steinwart2008svm}.

\subsubsection{Generalized Corruption-Corrected Learning with Bayesian Inverse}
For corruptions that involve more than just label corruption, we cannot obtain loss corrections using the same proof technique as \cref{thm:losscorr-label}.
This is because, when applying the risk conservation formula in \cref{prop:binv-exp-preserv}, the model class itself is also affected by the corruption kernel.
Formally, assume $\ka^{\dag} = \tau \otm \lambda$ with $I(\tau)=X$ and $I(\lambda)=Y$. Then, \cref{eq:ure} becomes
\begin{align} \label{eq:ure-gcl}
    \risk_{P} ( \ell \circ h ) = \risk_{P\circ \ka} \Big[ \ka^\dag ( \ell \circ h ) \Big] = \risk_{P\circ \ka } \Big[ (\tau \otm \lambda) ( \ell \circ h )\Big], \quad \forall h\in \mc M(X,Y),
\end{align}
which does not necessarily imply \cref{eq:loss-correction-goal}.
In \cref{eq:ure-gcl}, the corruption effect on loss and model class is in general \emph{indistinguishable}; we are not able to rewrite the set $(\tau \otm \lambda)(\ell \circ \mc H)$ as the composition $\t{\ell} \circ \t{\mc H}$, let alone $\t{\ell} \circ {\mc H}$ as per the \gls*{cl} case.
Nevertheless, our framework still allows for some new understanding on how attribute corruption may be corrected.

To this end, we formalize a weakened version of the \gls*{cl} paradigm, requiring to find a loss correction formula $\t{\ell}$ that depends on $\ell$, $h$ and $\ka^{\dag}$. That is the \gls*{gcl} paradigm, defined as the condition
\begin{align}\label{eq:glc-goal}
    h^* \in \argmin_{h\in\mc{H}}{\risk}_{\t P}[\tilde{\ell} \circ h]  \quad \text{and} \quad h^* \in \argmin_{h\in\mc{H}} \risk_{P}  [\ell \circ h ]
\end{align}
with the additional \emph{factorization requirement} of 
\begin{align}\label{eq:glc-req}
\t \ell ( h, x, y) = \ka^\dag (\ell \circ h)(x,y), \quad \forall h \in \mc M(X,Y), x\in X, y \in Y,
\end{align}
where $\t \ell$ is a \emph{generalized loss} $\t \ell \colon \mc H \tm X \tm Y \to \bb R_{\ge 0}$.\footnote{As for the requirements of the factorization on the whole set of Markov kernels, it can be weakened for instance to $\mc H$ if we know that the minimizer is contained in the set.}

A key assumption made in the above paradigm is the existence of a minimizer for the problems $(\ell, \mc H, P)$ and $(\t \ell, \mc H, P)$; hence, for it to be used, a practitioner would need to impose additional conditions. 
We will discuss the feasibility of the paradigm after presenting the loss correction formulas. For now, we impose an additional assumption and show that it is enough for the implying well-posedness of \gls*{gcl}.

\begin{assumption} \label{a:minimizer}
    Given the clean learning problem $(\ell, \mc H, P)$, there exists a minimizer $h^*\in \mc H$ that attains the Bayes risk value associated to it, \ie,
    $\br_{\ell\circ\mc H}(P) = \risk_{P}[\ell \circ h^*].$
\end{assumption}

\begin{lemma}\label{lemma:correction-formula-implies-gcl}
    Let $(\ell, \mc H, P)$ be a clean learning problem respecting \cref{a:minimizer}, and $\ka = \tau \otm \lambda$ a Markovian corruption kernel in $\mc M(\XY,\XY)$ on it. Denote the risk minimizer for the clean problem as $h^*\in \mc H$. Then, if \cref{eq:glc-req} is fulfilled by some $\t\ell$, $h^*$ is also the minimizer of the \gls*{gcl}-corrected problem $(\t\ell, \mc H, P \circ \ka)$, as per \cref{eq:glc-goal}.
\end{lemma}
\begin{proof}
    Consider the problem $(\ka^\dag(\ell\circ\mc H), P\circ \ka)$ and assume that there is some (possibly generalized) loss function $\t\ell$ such that \cref{eq:glc-req}.  By \cref{prop:binv-exp-preserv} and \cref{eq:glc-req},
    \begin{align}
        \risk_{P} ( \ell \circ h ) &\overset{P\ref{prop:binv-exp-preserv}}{=} \risk_{P \circ \ka} \Big[ (\tau \otm \lambda) ( \ell \circ h )\Big] \overset{\eqref{eq:glc-req}}{=} \risk_{P \circ \ka} ( \t\ell \circ h ),
    \end{align}
    which, taking infimum on the right- and left-most equations, implies the thesis.
\end{proof}

Hence, finding loss correction formulas respecting \cref{eq:glc-req} and \cref{a:minimizer} allows us to use the \gls*{gcl} correction paradigm.

%\textbf{Loss Correction Formulas.} 
We now state and prove the correction results for all the corruption case lying within the \gls*{gcl} paradigm.
Recall that the notation $f \# \mu$ stands for the push-forward probability measure of the distribution $\mu$ through the function $f$, defined as $(f \# \mu)(A) \coloneqq \mu(f^{-1}(A))$ for a suitable set $A$.
In the following we will use such notation for kernels.
By definition of kernel, $\tau$ \st $I(\tau)=X$ is a measure when fixing $\t x \in X$, and considering the kernel $h \in \mc M(X,Y)$ as its associated function $h \colon X \to \mc P(Y)$ yields a measurable function; hence, we can write 
$$(h \# \tau)(\t x) (A) \coloneqq \tau (\t x, h^{-1}(A)), \ A \subset \mc P (Y),$$ 
that is a family of distributions defined on a set of probability measures on $Y$, evaluated on $A$ and indexed by $\t x$.  
Since it is indexed and induced by Markov kernels, we can see it as a posterior probability on the set $\mc P (Y)$, given $\t x$.

\begin{theorem} \label{thm:losscorr-gen}
    Let $( \ell, \mc H, P)$ be a clean learning problem with $\ell$ being a bounded loss function and such that \cref{a:minimizer} holds. 
    Let $\ka^{\dag} = \tau \otm \lambda \in \mc M (\XY,\XY)$ be the one-step Markovian cleaning kernel inverting $\ka$, such that $(\ka^{\dag}( \ell \circ \mc H), P \circ \ka)$ is its associated corrected problem.
    Then, we can find a generalized loss $\t \ell\colon \mc H \tm \XY\to \bb R_{\ge0}$ respecting the \gls*{gcl} paradigm. In particular:
    \vspace{4pt}
    \begin{enumerate} 
        \setlength{\itemsep}{4pt}
        
        \item When $\ka^{\dag}$ is of the form $(\tau \colon X \karrow X)$ $\otm$ $(\lambda\colon Y \karrow Y)$, or $(\tau \colon X \karrow X)$ $\otm$ $(\lambda\colon X \tm Y \karrow Y)$, or $(\tau \colon X \tm Y \karrow X)$ $\otm$ $(\lambda\colon Y \karrow Y)$, we have 
        \begin{equation*}
            \t{\ell}(h, \t x, \t y) \coloneqq \bb E_{\rv u \sim (h \# \cemph{\tau})(\t x)} [ \cemph{\lambda}\ell\, (\rv u, \t y) ] \quad \forall \, ( \t x, \t y) \in X \tm Y \,.
        \end{equation*}
        When both corruptions are simple, the $\lambda\ell$ formula remains unchanged.
        When $\lambda$ is 2-dependent, it induces $\cemph{\lambda} \ell \, (\rv u, \t x, \t y)$.
        Lastly, we get $({h \# \cemph{\tau}})(\t x)$ to be replaced by $({h \# \cemph{\tau}})(\t x, \t y)$ when $\tau$ is 2-dependent.
        
        \item When $\ka^{\dag}$ is of the form $(\tau \colon Y \karrow X)$ $\otm$ $(\lambda\colon X  \karrow Y)$, we have
        \begin{equation*}
            \t{\ell}(h, \t x, \t y) \coloneqq  \bb E_{\rv u \sim (h \# \cemph{\tau})(\t y)} [ \cemph{\lambda}\ell \, (\rv u , \t x ) ] \quad \forall \, (\t x, \t y) \in X \tm Y \,.
        \end{equation*} 
        
        \item When $\ka^{\dag}$ is of the form $(\tau \colon Y \karrow X)$ $\otm$ $(\lambda\colon X \tm Y \karrow Y)$, or $(\tau \colon X \tm Y \karrow X)$ $\otm$ $(\lambda\colon X  \karrow Y)$, we respectively have
        \begin{align*}
            \t{\ell}(h, \t x, \t y) &\coloneqq  \bb E_{\rv u \sim (h \# \cemph{\tau})(\t y)} [ \cemph{\lambda} \ell \, (\rv u, \t x, \t y) ] \quad
            \forall \, (\t x, \t y) \in X \tm Y\,; \\
            \t{\ell}(h, \t x, \t y) &\coloneqq \bb E_{\rv u \sim (h \# \cemph{\tau})(\t x, \t y)} [ \cemph{\lambda} \ell \, (\rv u, \t x) ] \quad
            \forall \, (\t x, \t y) \in X \tm Y\,.
        \end{align*} 

        \item When  $\ka^{\dag}$ is of the form $(\tau \colon X \tm Y \karrow X)$ $\otm$ $(\lambda\colon X \tm Y  \karrow Y)$, we have
        \begin{equation*}
            \t{\ell}(h, \t x, \t y) \coloneqq  \bb E_{\rv u \sim (h \# \cemph{\tau})(\t x, \t y)} [ \cemph{\lambda}\ell \, (\rv u , \t x , \t y) ] \quad \forall \, (\t x, \t y) \in X \tm Y \,.
        \end{equation*} 
    \end{enumerate}
\end{theorem}

\begin{proof}
Let us consider a general function in the set $\ka^\dag (\ell \circ \mc H)$,
$$\ka^\dag (\ell \circ h) \coloneqq \sum_{y \in Y} \int_{x \in X} \ell(h(x),y) \,\ka^{\dag}(\t x, \t y, dx, dy) = \sum_{y \in Y} \int_{x \in X} \ell(h(x),y) \,(\tau \otm \lambda)(\t x, \t y, dx, dy) \;. $$
By \cref{lemma:correction-formula-implies-gcl}, we are ensured that finding a loss correction formula $\t\ell$ that respects \cref{eq:glc-req} is enough for ensuring that such loss respects the \gls*{gcl} paradigm. Hence, we consider here the function in the sets $\ka^\dag(\ell\circ\mc H)$ and define an ad-hoc $\t\ell$ depending on how $\ka^\dag$ affects $\ell\circ\mc H$.

We have already seen in \cref{thm:losscorr-label} how loss correction formulas are identified by a $\lambda$ that is simple, 1- or 2- dependent. For this reason, we compute step-by-step the simplest case of attribute corruption (a simple $\tau$) combined with a more complex label corruption (2-dependent $\lambda$), and understand how attribute corruption influences the loss correction formulas. All remaining cases can be computed with small variations.

Consider the $\ka^{\dag}$ from such that $\ka^\dag=\tau\otm\lambda$ with $\tau \in \mc M(X,X)$ and $\lambda \in \mc M(\XY,Y)$. We can define the loss correction $\t\ell$ as
\begin{align}
    \t{\ell}(h, \t x, \t{y}) &\coloneqq \sum_{y \in Y} \int_{x \in X} \ell(h(x),y) \, \tau(\t x, dx) \, \lambda(\t x, \t y, dy) \nonumber \\
    &= \sum_{y \in Y} \int_{x \in h(X)} \ell(u,y) \, \tau ( \t x, h^{-1}(du)) \, \lambda( \t x, \t y, dy)  \nonumber\\
    &= \int_{x \in h(X)} (\lambda\ell)_{\t x}(u,\t{y}) \, \tau ( \t x, h^{-1}(du)) \label{eq:pushf} ,
\end{align}
where $u = u(dy) \in \mc P(Y)$ and $h(X)\coloneqq \{ h(x) \mid x \in X\}$. 
The following equality holds:
$$\bb E_{\rv u \sim \tau (\t x, h^{-1}( \cdot )) }[ \rv u] = \int_{x \in h^*(X)} u \, \tau (\t x, h^{-1}(du)) = (h \# \tau)(\t x) \in  \mc P (Y) ,$$ 
by definition of $\mc H$ as a subset of $\mc M(X, Y)$ and using the definition of $h \# \tau$.
We remark that $\tau (\t x, (h)^{-1}(du))$ is then a probability in $\mc P(\mc P(Y))$.
Hence we can rewrite \cref{eq:pushf} as
\begin{align}
    \t{\ell}(h, \t x, \t{y}) &\coloneqq \int_{u \in \mc P(Y)}  \, (\lambda\ell)_{\t x}(u,\t{y}) \, \tau ( \t x, h^{-1}(du)) = \bb E_{\rv u \sim (h \# \tau)(\t x) }[(\lambda\ell)(\rv u, \t x, \t y)] .  \nonumber
\end{align}
This is the same correction formula found in point 1.

As for more dependent attribute corruptions,  \ie, $\tau(\t x,  \t y, dx) $, the action on the hypothesis will be dependent from $\t{y}$. Therefore we obtain
$\t{\ell}(h, \t x, \t{y}) =  \bb E_{\tau (\t x, \t y, h^{-1}( \cdot )) }[ (\lambda\ell)(\rv u, \t x, \t y)] \;,$
where only the simple label corruption can be considered, given the missing result for the \gls*{br} equality in the $D(\tau) = D(\lambda) = \XY$ case. 
Hence, for the remaining points 2, 3 and 4, we follow the same procedure deployed in the above, using the action formula of dependent corruptions as described in the proof of \cref{thm:br-1d-2d,thm:br-2-2}, and obtain the thesis by fulfilling the \gls*{gcl} requirement in \cref{eq:glc-req}.
\end{proof}

\subsubsection{Discussion of the {\normalfont \textsc{gcl}} Paradigm and Its Correction Formulas} 
\cref{thm:losscorr-gen} provides a constructive proof for the existence of \gls*{gcl} corrected losses, \ie, that respect both \cref{eq:glc-goal} and \cref{eq:glc-req}, granted that \cref{a:minimizer} holds.
What is therefore shown is that the \gls*{gcl} paradigm is closely related to the \gls*{cl} one, but it is only defined for certain classes of learning problem, as it imposes the existence of a minimizer \st the condition in \cref{eq:glc-goal} is well-posed. 
Within our setup, we could fulfill the assumption by requiring the loss $\ell$ not only to be positive, measurable and bounded but also \emph{continuous} on $\mc P(Y)$ in the Euclidean topology, and $\mc H$ to be \emph{compact}. This would imply continuity of the functional $h \in \mc H \mapsto \risk_{P}[\ell(h (x), y)]$, and therefore the existence of a minimizer on the compact set $\mc H$. 

While \gls*{gcl} is on one hand adding more restrictive conditions, it is also weakening some of the \gls*{cl} requirements: it asks the corrected loss to be only a generalized loss, which is then used in the factorization condition \cref{eq:glc-req}.
Having a generalized loss is necessary for the attribute corruption, as standard losses may fail to mitigate this case. The following example showcases one of such scenarios.

\begin{example}
\label{ex:non-inj-model-gcl}
Consider a simple neural network with one hidden layer of two ReLU neurons and a softmax output,
\begin{align*}
    h &\colon X \to \mc P(Y), \qquad
    h(\mb{x}) \coloneqq \mathrm{softmax}\big( W_2 \, \sigma_{\text{ReLU}}(W_1 \mb{x} + b_1) + b_2 \big),
\end{align*}
where $\mb{x} \in X \subseteq \bb R^2$, $y \in Y=\{0,1\}$,
$W_1 \in \bb R^{2 \tm 2}$, $b_1 \in \bb R^2$,
$W_2 \in \bb R^{2 \tm 2}$, $b_2 \in \bb R^2$,
and $\sigma_{\text{ReLU}}(z)=\max(0,z)$.
Writing the hidden activations $n_j$ and logits explicitly, we get
\begin{align*}
    n_j(\mb{x}) &= \max\{0, \langle [W_1]_j, \mb{x} \rangle + b_{1,j}\}, \quad j=1,2, \\
    z_i(\mb{x}) &= \langle [W_2]_i, \mb{n}(\mb{x}) \rangle + b_{2,i}, \\
    h(\mb{x})_i &= \frac{e^{z_i(\mb{x})}}{\sum_{j=1}^{2} e^{z_j(\mb{x})}}.
\end{align*}
Here $[W_1]_j$ denotes the $j$-th row of $W_1$ and $[W_2]_i$ denotes the $i$-th row of $W_2$. 
For simplicity and without loss of generality, we can choose $b_{2,i}=0$ and the weights $W_1$ to be always positive.

For our argument, we want to consider the points in $X \subset \bb R^2$ for which the ReLU activation makes the model $h(x)$ \emph{non-injective}: that happens if both pre-activations $\langle [W_1]_j, \mb{x} \rangle + b_{1,j}$ are negative, \ie, for the set of inputs
\begin{align*}
    \mc S_h \coloneqq 
    \Biggl\{ \mb{x} = (x_1, x_2) \in \mathbb{R}^2 \;\Big|\; 
    x_2 < \min\Biggl(
    \frac{-b_{1,1} - [W_1]_{1,1} x_1}{[W_1]_{1,2}}, \;\;
    \frac{-b_{1,2} - [W_1]_{2,1} x_1}{[W_1]_{2,2}}
    \Biggr) \Biggr\}.
\end{align*}
Hence, we get that $n_1(\mb{x}) = n_2(\mb{x}) = 0$, and therefore $h(\mb{x}) = (0.5, 0.5)$.
This implies that
\begin{align*}
    \ell(h(\mb{x}_1), y) = \ell(h(\mb{x}_2), y), \quad \forall y \in Y, \ \forall \mb{x}_i \in \mc S_h .
\end{align*}
However, if we apply a deterministic attribute corruption kernel, \ie, $\ka^\dag \coloneqq \delta_{f(\mb{x})} \otm \delta$ with $f\colon X \to X$ measurable and $\delta \in \mc M(Y,Y)$, we have that 
\begin{align*}
    \ka^\dag (\ell\circ \mc H) = \Big\{ \ell(h(f(\mb{x})), y) \,\mid\, h \in \mc H \Big\}.
\end{align*}
Imposing $\t\ell (h(\mb{x}),y) = \ell(h(f(\mb{x})), y) $ as per the \gls*{cl} definition (\ie, stretching \cref{eq:loss-correction-bayesian} to the attribute corruption case), with $\t\ell\colon \mc P(Y) \tm Y \to \bb R_{\ge 0}$, implies that 
\begin{align*}
    \ell\Big(h(f(\mb{x}_2)), y\Big) = \t\ell\Big(h(\mb{x}_2),y\Big) 
    = \t\ell\Big(h(\mb{x}_1),y\Big) 
    = \ell\Big(h(f(\mb{x}_1)), y\Big),
\end{align*} 
for $\mb{x}_1, \mb{x}_2 \in \mc S_h$ and for all $y\in Y$. 
This yields a contradiction. Indeed, since $h(x)$ is constant on $\mathcal S_h$, any standard corrected loss of the form $\tilde \ell(h(x),y)$ must assign the same value to all $x \in \mathcal S_h$. 
However, one can construct a measurable transformation $f$ (for instance, a suitable translation depending on $h$) such that %$f(\mb{x}_1)$ exits the inactive region while $f(\mb{x}_2)$ remains inside it. Consequently, 
$h(f(\mb{x}_1)) \neq h(f(\mb{x}_2))$, which implies
$$
\ell(h(f(\mb{x}_1)),y) \neq \ell(h(f(\mb{x}_2)),y).
$$
This contradicts the requirement that the corrected loss depends only on $h(x)$. Therefore, in this case a standard corrected loss $\t\ell\colon \mc P(Y) \tm Y \to \bb R_{\ge 0}$ cannot effectively mitigate the consequences of this attribute corruption. 
The correction necessarily depends on $h$, which justifies the need for generalized losses in the \gls*{gcl} framework.
\end{example} 

The factorization requirement in \cref{eq:glc-req} is the key property that defines \gls*{gcl}, as it implies
$$
\risk_{P \circ \ka}(\t \ell \circ h^*) = \risk_{P \circ \ka} \left[\ka^\dag (\ell \circ h^*)\right],
$$
and by \cref{prop:binv-exp-preserv},
$$
\risk_{P \circ \ka} (\t \ell \circ h^*) = \risk_{P}( \ell \circ h^*).
$$
Since the factorization holds for every $h \in \mc H$, it follows that $h^*$ is also a minimizer for the factorized corrected problem $(\t \ell, \mc H, P \circ \ka)$. Investigating which conditions on the learning problem ensure the factorization property is beyond the scope of this analysis. 
The conclusion to be taken from our loss correction formulas is not that they provide a new practical tool, but rather that they demonstrate how, even under optimistic assumptions, loss correction techniques become far more complex and involved when dealing with attribute corruption kernels.

Indeed, all the novel correction formulas found in this chapter are more complex than the ones defined in previous works \citep{van2017theory,patrini2017making}, which only consider a label corruption scenario similar to our \gls*{cl} for simple label corruption. 
Our version, \cref{thm:losscorr-label}, further extends the setting by including the dependent label corruption.
The second set of results, included in \cref{thm:losscorr-gen}, are instead fulfilling the weaker conditions imposed by the \gls*{gcl} framework, but only for a subset of learning problems for which an optimal model exists.
Our loss correction formulas offer us an important insight: under a Bayes risk point of view, \emph{there is another fundamental difference between label and attribute corruption}. 
They induce distinct corrupted learning settings, and traditional loss correction does not ensure unbiased learning in the sense of \gls*{cl} in the presence of the latter. 

\section{Conclusions}
We proposed a comprehensive and unified framework for general corruption, extending its definition also to model class and loss function changes. 
We did so by using Markov kernels, and systematically studying corruption in three key aspects: typology, consequence, and correction. 
The choice of working with Markov kernels enables the use of information-theoretical tools, and provides an alternative interpretation of corruption as an \emph{observation channel} through which we get to see our data distribution. 
This mathematical modelization allows one to consider data as a dynamic element of a learning problem, as opposed to the view of data as static facts and true representations of reality.

We established a new taxonomy for Markovian corruption of learning problems, yielding qualitative comparisons between corruption types in terms of their hierarchy. 
To gain a deeper understanding of corruption, we analyzed their consequences by proving Data Processing Equalities for Bayes risk. 
Given different possible factorizations of a corruption of the joint space, the learning problem is affected in different ways. 
Furthermore, we applied the equalities for obtaining loss correction formulas. 
Such an application is rather conventional, and usually leads to a proposed mitigation for the specific model considered. 
This work does not propose any mitigation algorithm, but analyzes the fundamental difference between label and attribute corruption. 
The Data Processing Equality results together with the analysis carried out in Section \cref{sec:losscorrect} lead us to the following conclusions:
\begin{itemize}
    \item Label and attribute corruption differ in how they change the learning problem. The former does not influence the model class; the latter changes model class and loss function in a way that generally cannot be disentangled.
    \item Classical corruption-corrected learning (\gls*{cl}) is not an adequate paradigm to study general corruption. For cases involving non-identical attribute corruption, we introduce a more general framework named generalized corruption-corrected learning (\gls*{gcl}).
    \item Loss correction formulas for attribute corruptions involve the notion of generalized loss and an expectation over the set of all $h \# \tau$ predictions. This suggests a \emph{negative} result, as standard loss correction techniques may not guarantee accurate learning when dealing with general attribute corruption.
\end{itemize}

\subsection{Limitations and Future Work}
We considered data as probability distributions, implicitly assuming that each dataset has an associated probabilistic generative process. For many applications of machine learning, such an assumption is not warranted.
Corruption is being induced by a Markov kernel, under the strong assumption of having full access to their actions. 
We note that in some cases Markov kernels can be estimated from corrupted data \citep{liu2015classification,scott2015rate}, but this question is in general still open and needs further investigation. 
The consequences of corruption are analyzed through Bayes risk without accounting for sampling or imperfect optimization. 
Bridging the gap between the distributional-level and the sample-level results would be the next step for this study, which requires tailored ad-hoc analyses.
Other directions for making this framework more practically usable include developing quantitative methods to compare corruption severity and investigating the effects of optimization algorithms on the analysis. 

From a more theoretical point of view, future work includes investigating the non-Markovian and multi-step classes of corruptions. As we pointed out in \cref{sec:taxonomy}, model misspecification lies within the general corruption class, and might be studied alone or as an additional corruption ``chained'' to a Markovian one. Similarly, changes in loss function can be analyzed further. 
Additionally, the topic of non-probabilistic corruption \citep{meng2022comments,boyd2023we}, only superficially touched in the present work, needs deeper analysis. 
It is unclear whether the current theoretical tools, deployed when dealing with distributional changes, are enough for characterizing and potentially mitigating their consequences on learning problems.

% Acknowledgements and Disclosure of Funding should go at the end, before appendices and references

\acks{This work was partially funded by the Deutsche Forschungsgemeinschaft (DFG, German Research Foundation) under Germany’s Excellence Strategy — EXC number 2064/1 — Project number 390727645, as well as by the German Federal Ministry of Education and Research (BMBF): T\"{u}bingen AI Center, FKZ: 01IS18039A.
The authors thank the International Max Planck Research School for Intelligent Systems (IMPRS-IS) for supporting Laura Iacovissi. Additional thanks to Armando J. Cabrera Pacheco, Nicol\`{o} Zottino and Jack Brady for helpful discussions, as well as Christian Fr\"{o}hlich, Wenkai Xu and one anonymous reviewer for giving thorough and valuable feedback on earlier versions.}

% Manual newpage inserted to improve layout of sample file - not needed in general before appendices/bibliography.

\appendix

\newpage
\section{Glossary}
\printunsrtglossary[sort=def,style=long,nonumberlist,title=]

\section{Summary of Actions and Consequences of Corruption}
% \newgeometry{heightrounded, showframe, left=3cm, bottom=0.1cm, top=1in}
% https://tex.stackexchange.com/questions/347327/section-title-above-table-in-landscape-mode

\begingroup
\vfill
\centering % uncomment to centre the table horizontally as well as vertically
\begin{sideways}
    \setlength{\tabcolsep}{8pt}
    \begin{threeparttable}
    \label[table]{tab:action}
    \caption{Corruption types, Bayes risk equalities, and loss correction formulas.
    Each corruption kernel $\kappa$ is factorized as $\tau$ (acting on inputs $X$) and $\lambda$ (acting on labels $Y$), modifying both the data distribution $P$ and the minimization set $\ell \circ \mathcal{H}$.
    The corresponding Data Processing Equality for Bayes risk is $\Big( \ell \circ \mc{H},  P \circ (\cemph{\tau} \otm \cemph{\lambda})  \Big) \brequiv \Big( \cemph{\tau}(\cemph{\lambda}\ell \circ \mc{H} ) , P \Big)$ (\cref{sec:brchange}), and the resulting loss correction formulas (\cref{sec:losscorrect}) are summarized. 
    All operators are expressed using the kernel operations defined in \cref{sec:operations}: \cref{p:comp} chain composition ($\circ$), \cref{p:prod} product composition ($\tm$), \cref{p:super} superposition ($\otm$), and \cref{p:part-chain} partial chain composition ($\circ_{(\cdot)}$).}
        \begin{tabular}{cccc}
        \toprule
        \thead{Corruption type \\ $\ka \coloneqq \tau \otm \lambda$} & \thead{Corruptions action \\ $P\to \t P \coloneqq P \circ (\cemph{\tau} \otm \cemph{\lambda})$} & \thead{Corruption action $\tau (\lambda\ell \circ \mc{H})$ on the minimization\\ set $\ell \circ \mc H \coloneqq \{\, (x,y) \mapsto \ell(h_x,y) \, | \, h \in \mc{H}\}$} & \thead{Loss correction formula\\ $\t{\ell}(h, \t x, \t y), \, \forall \, ( \t x, \t y) \in X \tm Y$} \\
        \midrule 
        $\thead{\cemph{ (\tau \colon X\karrow X) \otm} \\ \cemph{(\lambda \colon Y \karrow Y)}}$ 
        & \thead{$\t P = \pi_Y \circ \big[ (E \circ \cemph{\tau}) \otm \cemph{\lambda} \big]$ \\ $\t P = \pi_X \circ \big[ \cemph{\tau} \otm (F \circ \cemph{\lambda} )  \big]$} & {$ \{ (x, y) \mapsto [ \tau (\lambda\ell_{y} \circ h) ](x) , h \in \mc H \}$}
        & {$\bb E_{\rv u \sim (h \# \cemph{\tau})(\t x)} [ \cemph{\lambda}\ell\, (\rv u, \t y) ]$}
        \\
        \midrule 
        $\thead{\cemph{ (\tau \colon X \tm Y \karrow X) \otm}\\ \cemph{(\lambda \colon Y \karrow Y)}} $
        & \thead{$\pi_Y \circ \big[ ( E \circ_X \cemph{\tau}) \otm \cemph{\lambda} \big]$} & $ \{ (x, y) \mapsto [ \tau (\lambda\ell_{y} \circ h) ](x,y) , h \in \mc H \}$
        & {$\bb E_{\rv u \sim (h \# \cemph{\tau})(\t x, \t y)} [ \cemph{\lambda}\ell\, (\rv u, \t y) ]$}
        \\
        \midrule
        $\thead{\cemph{(\tau \colon X  \karrow X) \otm} \\ \cemph{ (\lambda \colon X \tm Y \karrow Y)}}$
        & $\pi_X \circ \big[ \cemph{\tau} \otm ( F \circ_Y \cemph{\lambda} ) \big]$ & $ \{ (x, y) \mapsto [ \tau (\lambda\ell_{(x,y)} \circ h ) ](x) , h \in \mc H \}$
        & {$\bb E_{\rv u \sim (h \# \cemph{\tau})(\t x)} [ \cemph{\lambda}\ell\, (\rv u, \t x, \t y) ]$}
        \\
        \midrule
        $\thead{\cemph{(\tau \colon Y \karrow X) \otm} \\ \cemph{(\lambda \colon X \tm Y \karrow Y)}}$ & $\pi_Y \circ \big[ \cemph{\tau} \otm ( E \circ_X \cemph{\lambda} ) \big]$ & $ \{ (x, y) \mapsto [ \tau (\lambda\ell_{(x,y)} \circ h ) ](y) , h \in \mc H \}$ & {$\bb E_{\rv u \sim (h \# \cemph{\tau})(\t y)} [ \cemph{\lambda} \ell \, (\rv u, \t x, \t y) ]$}
        \\
        \midrule
        $\thead{\cemph{(\tau \colon X \tm Y \karrow X) \otm} \\ \cemph{(\lambda \colon X \karrow Y)}}$ & $\pi_X \circ \big[ ( F \circ_Y \cemph{\tau} ) \otm \cemph{\lambda}  \big]$ & $ \{ (x, y) \mapsto [ \tau (\lambda\ell_{x} \circ h ) ](x,y) , h \in \mc H \}$ & {$\bb E_{\rv u \sim (h \# \cemph{\tau})(\t x, \t y)} [ \cemph{\lambda} \ell \, (\rv u, \t x) ]$}
        \\
        \midrule
        $\thead{\cemph{(\tau \colon Y \karrow X) \otm}\\ \cemph{(\lambda \colon X \karrow Y)}}$ & \thead{$\pi_Y \circ \big[ \cemph{\tau} \otm ( E \circ \cemph{\lambda} ) \big]$ \\ $\pi_X \circ \big[ ( F \circ \cemph{\tau} ) \otm \cemph{\lambda} \big]$} & $ \{ (x, y) \mapsto [ \tau (\lambda\ell_{x} \circ h ) ](y) , h \in \mc H \}$ 
        & {$\bb E_{\rv u \sim (h \# \cemph{\tau})(\t y)} [ \cemph{\lambda}\ell \, (\rv u , \t x ) ]$}
        \\
        \midrule
        \thead{$\cemph{(\tau \colon X \tm Y \karrow X) \otm}$ \\ $\cemph{(\lambda \colon X \tm Y \karrow Y)}$} & $P \circ (\cemph{\tau}\otm\cemph{\lambda})$ & $\{ (x, y) \mapsto [ \tau (\lambda\ell_{(x,y)} \circ h ) ](x,y) , h \in \mc H \}$ & $\bb E_{\rv u \sim (h \# \cemph{\tau})(\t x, \t y)} [ \cemph{\lambda}\ell \, (\rv u , \t x , \t y) ]$
        \\
        \bottomrule
        \end{tabular}
    \end{threeparttable}
\end{sideways}
\vfill
\endgroup

%exhaustiveness is needed for the following paragraph*, so it should be the last in this section
\section{Related Existing Paradigms}
\label[section]{sec:related}

A Markov kernel-based taxonomy is substantially different from previous work. Therefore, in this section, we carefully examine how existing corruption models fit into our taxonomy. This involves reformulating them as specific instances of Markov corruptions, thereby unveiling their relationships within the corruption hierarchy presented in \cref{fig:corruptions}a. 

The primary challenge stems from the lack of consistency across the literature; different authors sometimes refer to the same corruption process with different names or use the same name to denote different settings. 
For instance, classical studies on concept drift \citep{widmer1996learning, Lu_2018} generally define it as a mismatch in the joint distributions between two different learning environments, \eg, training and test times. Meanwhile, works such as in \citet{moreno2012unifying} characterize it further by necessitating unchanged attribute or label priors.

We attempt a partial unification of the corruption models we are aware of by establishing connections as depicted in \cref{tab:corr-model-sec3}, while additional technical intricacies regarding correspondences and relationships are elucidated in subsequent sections.

% Check if they are all included in the notation table!
% As a reminder, we review the commonly used notations.
% Let $E\colon Y \karrow X$ be an experiment and $F: X \karrow Y$ be a posterior kernel. The clean distribution $P$ can be represented either in a discriminative manner as $\pi_{X} \tm F$ or in a generative manner as $\pi_{Y} \tm E$. However, we cannot observe samples drawn from the clean distribution $P$, but observe samples from some corrupted distribution $\tilde{P}$. The corruption is generally represented as $\ka_{Z\t{Z}}$, where the variables $z=(x,y) \in Z$ are referred to as parameters and $\t z=(\t x, \t y) \in Z$ are referred to as corrupted variables, written in the kernel formulation as differentials $d\t z = d\t x d\t y$ . $\delta_{Z\t{Z}}$ denotes a kernel induced by the Dirac delta measure from $(Z,\mc{Z})$ to $(Z,\mc{Z})$.

\subsection{Simple Corruptions}

The most well-known and widely studied corruptions in the literature are the simple cases, where the corruption solely acts on the feature space $X$ or the label space $Y$. We discuss in the following various examples of simple corruptions, \ie, in the sets $\mc M (X,X)$ and $\mc M(Y,Y)$, as defined in \cref{fig:corruptions}.

\subsubsection{Attribute Noise} The problem of attribute noise concerns errors that are introduced into the observations of attribute $\rv X$, leaving the labels untouched \citep{shackelford1988learning, goldman1995can, zhu2004class,williamson2022information}.
Widely studied examples of such errors include erroneous attribute values and missing attribute values.
Instead of observing $(\rv X, \rv Y)$, in the first case, one can only observe a distorted version of $\rv X$, e.g.~$(\rv X+\rv N, \rv Y)$ with some independent noise random variable $\rv N \ind \rv X$;
in the second case, one's observation of $\rv X$ contains missing values.

Let $\mathbf{X}=(x_{ij})_{1\leq i \leq n, 1\leq j \leq d}$ be the complete input matrix, with $|X|=n$, and $\mathbf{M}=(m_{ij})_{1\leq i \leq n, 1\leq j \leq d}$ be the associated missingness indicator matrix such that $m_{ij}=1$ if $x_{ij}$ is observed and $m_{ij}=0$ if $x_{ij}$ is missing.
Then the corresponding observed input matrix is $\mathbf{X}_{o}=\mathbf{X}\odot\mathbf{M}$ and its missing counterpart is $\mathbf{X}_{m}=\mathbf{X}-\mathbf{X}_{o}$, where $\odot$ denotes Hadamard product.
The missing value mechanisms are further categorized into three types based on their dependencies \citep{rubin1976inference, little2019statistical}:\footnote{Assume the rows $x_i$, $m_i$ are assigned a joint distribution. and $\rv X$ and $\rv M$ are treated as random variables.}
\begin{itemize}
    \item Missing completely at random (\textsc{mcar}): the cause of missingness is entirely random, \ie, $p(\rv{M} \mid \rv{X})=p(\rv{M})$ does not depend on $\rv{X}_{o}$ or $\rv{X}_{m}$. This corresponds to having a trivial Markov kernel acting on the clean distribution, $\tau \colon \{*\} \karrow X \equiv \mu \in \mc P (X)$.
    \item Missing not at random (\textsc{mnar}): the cause of missingness depends on both observed variables and missing variables, \ie, $p(\rv{M} \mid \rv{X})=p(\rv{M} \mid \rv{X}_{o}, \rv{X}_{m})$. This case corresponds to our non-trivial $\tau \colon X \karrow X$.
    \item Missing at random (\textsc{mar}): the cause of missingness depends on observed variables but not on missing variables, \ie, $p(\rv{M} \mid \rv{X})=p(\rv{M} \mid \rv{X}_{o})$. This case is a sub-case of the non-trivial $\tau \colon X \karrow X$, which is not directly specifiable by our taxonomy because of the different premises it is built on.
\end{itemize}

\subsubsection{Class-Conditional Noise (\textsc{ccn})}
The problem of \gls*{ccn} arises in situations where, instead of observing the clean labels, one can only observe corrupted labels that have been flipped with a label-dependent probability, while the marginal distribution of the instance remains unchanged \citep{natarajan2013learning, patrini2017making, van2017theory, williamson2022information}.
\gls*{ccn} is an example of {simple label corruption, $\mc M (Y,Y)$,} that can be formulated as a corrupted posterior. 
For classification tasks, $Y$ is assumed to be a finite space.
Therefore the corruption {$\lambda \colon Y \karrow Y$} can be represented by a column-stochastic matrix $\mathbf{T}=(\rho_{ij})_{1\leq i \leq |Y|, 1\leq j \leq |Y|}$ which specifies the probability of the clean label $\rv Y=j$ being flipped to the corrupted label $\t{\rv Y}=i$, \ie, $\forall i,j, \, \rho_{ij}=p(\t{\rv Y}=i \mid \rv Y=j)$.
The corrupted joint distribution can be rewritten as $\t P=\sum_{Y}\, p(\t{\rv Y} \mid \rv Y)\, p(\rv{Y} \mid \rv{X})\, p(\rv{X})$.
In the literature, $\mathbf{T}$ is known as the noise transition matrix with its elements $\rho_{ij}$ referred to as the noise rates, and is useful for designing loss correction approaches (our results in \cref{sec:losscorrect} significantly generalize existing loss correction results in \gls*{ccn} to our broad class of simple, dependent and combined corruptions) \citep{patrini2017making}. 
Prior to the proposal of the \gls*{ccn} model, early studies primarily focused on a symmetric subcase of $\mathbf{T}$ in binary classification, known as random classification noise (\textsc{rcn}) \citep{angluin1988learning, blum1998combining, van2015learning}. 
Note that in \textsc{rcn}, the output of the corruption {$\lambda \colon Y \karrow Y$} remains constant \wrt its parameters. 
Recently, some variants of \gls*{ccn} have been further developed, for example, in \citet{ishida2017learning, ishida2019complementary}, complementary labels are modeled via a symmetric $\mathbf{T}$ whose diagonal elements are all equal to zero.

\subsection{Dependent Corruptions}
Although simple corruptions have been well studied and understood, more complexities arise in dependent cases, yet they receive relatively less attention and understanding. {We discuss in the following examples of the dependent corruptions in the sets $\mc M (Y, X)$, $\mc M (X,Y)$, $\mc M (X \tm Y, X)$ and $\mc M (X \tm Y,Y)$, as defined in \cref{fig:corruptions}a.}

\subsubsection{Style Transfer} 
Style transfer refers to the process of migrating the artistic style of a given image to the content of another image \citep{gatys2015neural, johnson2016perceptual}. The primary objective is to recreate the second image with the designated style of the first image. In recent developments, it has also been applied to audio signals \citep{grinstein2018audio}.
If we represent the style of the first image by $\rv Y$, and the second image and the reconstructed image as $\rv X$ and $\t{\rv X}$ respectively, style transfer serves as an illustrative example of {$\tau \colon Y \karrow X $} ``corruption''.
Note that the aim here is to \emph{learn how to corrupt} instead of learning in the presence of corruption. We mention this connection because our framework can also be used also with different purposes, but underline that our \gls*{br} results are not applicable to this case.
The process of style transfer can be formulated as a corrupted posterior.

\subsubsection{Adversarial Noise}
In contrast to additive random attribute noise, adversarial noise is specifically crafted by adversaries for each instance with the intent of changing the models' prediction of the correct label \citep{szegedy2013intriguing, goodfellow2014explaining, papernot2016limitations, kurakin2018adversarial, hendrycks2021natural}.
Such adversarial examples raise significant security concerns as they can be utilized to attack machine learning systems, even in scenarios where the adversary has no access to the underlying model.
The adversarial noise is an example of {$\tau \in \mc M(X \tm Y, X)$} corruption that can be formulated as a corrupted experiment.

\subsubsection{Instance-Dependent Noise ({\normalfont \textsc{idn}})}
As a counterpart to \gls*{ccn}, the problem of \textsc{idn} arises in situations where, instead of observing the clean labels, one can only observe corrupted labels that have been flipped with an instance-dependent (but not label-dependent) probability \citep{ghosh2015making, menon2018learning}.
It is a special case of the \textsc{iln} noise model, which we will describe later.
\textsc{idn} is an example of {$\lambda \in \mc M (X,Y)$} corruption that can be formulated as a corrupted experiment. 

\subsubsection{Instance- and Label-Dependent Noise ({\normalfont \textsc{iln}})}
\textsc{iln} is the most general label noise model, which arises in situations where, instead of observing clean labels, one can only observe corrupted labels that have been flipped with an instance- and label-dependent probability \citep{menon2018learning, cheng2020learning, yao2021instance, wang2021tackling}.
\textsc{iln} is an example of {$\lambda \in \mc M (X\tm Y,Y)$} corruption that can be formulated as a corrupted posterior.
Compared to the matrix representation $\mathbf{T}$ of the \gls*{ccn} corruption $\ka_{Y\t{Y}}$, the \textsc{iln} corruption $\ka_{XY\t{Y}}$ can be represented by a matrix-valued function of the instance $\mathbf{T}(x)=(\rho_{ij}(x))_{1\leq i \leq|\t Y|, 1\leq j \leq |Y|}$ which specifies the probability that the instance $\rv X=x$ with the clean label $\rv Y=j$ being flipped to the corrupted label $\t{\rv Y}=i$, \ie, $\forall i,j, \, \rho_{ij}(x)=p(\t{\rv Y}=i \mid \rv Y=j, \rv X=x)$.
Some subcases of \textsc{iln} have also been studied in the literature, for example, the boundary-consistent noise, which considers a label flip probability based on a score function of the instance and label. The score aligns with the underlying class-posterior probability function, resulting in instances closer to the optimal decision boundary having a higher chance of its label being flipped \citep{du2015modelling}.

\subsection{Combined Corruptions}
\label[section]{subsec:comb-corr-examples}
Given the simple and dependent corruptions, we can combine them to generate 2-parameter joint corruptions, { \ie, $\mc M ( X \tm  Y , X \tm Y )$.}
Below, we discuss some examples of combined noise models illustrated in \cref{fig:corruptions}b.

\subsubsection{Combined Simple Noise}
The simplest combined corruption is the combined simple noise, where the observations of attribute $\rv X$ are subject to some errors and the observed labels $\rv Y$ are flipped with a label-dependent probability \citep{williamson2022information}.
Combined simple noise is an example of {$\tau \colon X\karrow X \otm \lambda \colon Y \karrow Y$} corruption that can be formulated as a corrupted experiment.

\subsubsection{Target Shift}
In the literature, target shift, also known as prior probability shift, refers to the situation where the prior probability $p(\rv Y)$ is changed while the conditional distribution $p(\rv X \mid \rv Y)$ remains invariant across training and test domains \citep{japkowicz2002class, he2009learning, buda2018systematic, lipton2018detecting}.
The definition is established by assuming certain invariance from a generative perspective of the learning problem, that is, considering it as a corruption of the experiment according to $P = \pi_Y \tm E$.
However, when examining the learning problem from a discriminative perspective, the change in $p(\rv Y)$ may cause changes in both $p(\rv X)$ and $p(\rv Y \mid \rv X)$ due to the Bayes rule.
Existing frameworks for the categorization of target shift do not capture these implications, as they are based on the notion of invariance from a single perspective of the $E$ direction.
In contrast, our framework categorizes corruptions based on their dependencies and therefore is advantageous by offering dual perspectives from both the $E$ and $F$ directions.
Specifically, target shift is a subcase of $\tau \colon Y \karrow X$ $\otm$ $\lambda \colon X \tm Y \karrow Y$ corruption and can be formulated either as a corrupted experiment or as a corrupted posterior.
The corrupted distribution is given by $\t P = ( \pi_Y \tm E ) \circ (\tau \otm \lambda) $ or $\t P = ( \pi_X \tm F) \circ (\tau \otm \lambda) $.

\subsubsection{Covariate Shift}
In the literature, covariate shift refers to the situation where the marginal distribution $p(\rv X)$ is changed while the class-posterior probability $p(\rv Y \mid \rv X)$ remains invariant across training and test domains \citep{shimodaira2000improving, quinonero2008dataset, sugiyama2012machine, zhang2020one}.
Similarly to target shift, the definition is based on assuming invariance from the discriminative perspective of the learning problem, treating it as a corruption of the posterior using $P = \pi_X \tm F$.
However, when viewed from a generative perspective, changes in $p(\rv X)$ may lead to changes in $p(\rv Y)$ and $p(\rv X \mid \rv Y)$ due to the Bayes rule.
Covariate shift is a subcase of $\tau \colon X \tm Y \karrow X$ $\otm$ $\lambda \colon X \karrow Y$ corruption and can be formulated either as a corrupted posterior or as a corrupted experiment. 
The corrupted distribution is given by $\t P = ( \pi_Y \tm E ) \circ (\tau \otm \lambda) $ or $\t P = ( \pi_X \tm F) \circ (\tau \otm \lambda) $.

It is important to clarify that while covariate shift is sometimes used interchangeably with sample selection bias in certain literature, the two are not synonymous.
This point is also mentioned by the author of the original covariate shift paper \citep{shimodaira2000improving} in the book by \citet[Chapter 11]{quinonero2008dataset}: they claim covariate shift to be a special form of selection bias when the latter is taken under assumption of missing at random, and in general, selection bias without such a structure is difficult. 
However, based on our definition of selection bias in \cref{df:sbias}, it is not true that covariate shift is a special form of selection bias. Nonetheless, various definitions exist in the literature and they can relate in different ways. 

We introduce here a classical definition of selection bias, which leads to the one we gave in the main text, see \citep[Chapter 3.2]{quinonero2008dataset}.
Let $\rv S$ be a binary selection variable deciding whether a datum is included in the training set ($\rv S=1$) or excluded from it ($\rv S=0$).
The corrupted distribution by selection bias can be expressed as $\t P (\rv X, \rv Y)=P(\rv X, \rv Y \mid \rv S=1)$.
By assuming the missing at random structure, where $\rv S$ is independent of $\rv Y$ given $\rv X$: $P(\rv S \mid \rv X, \rv Y) = P(\rv S \mid \rv X)$, we recover covariate shift where $P(\rv X \mid \rv S=1) \neq P(\rv X)$ and $P(\rv Y \mid \rv X, \rv S) = P(\rv Y \mid \rv X)$.

Note that covariate shift is only harmful when the model class is misspecified \citep{shimodaira2000improving}. 
This issue is typically addressed through importance-weighted empirical risk minimization--weighting the training losses according to the ratio of the test and training input densities \citep{sugiyama2012machine,fang2023generalizing}. 
In such context, the additional assumption of $P \ll \t P$ is required so to obtain the weighted risk on the training set to be equal to the risk on the test set. 
This assumption is therefore in contrast with \cref{df:sbias}, requiring for selection bias the support condition $\t P \ll P$.

More generally, selection bias necessitates both the support condition and the selection condition with bounded $\frac{dP}{d\t{P}}(x_i,y_i) \, \forall i \in [n]$, which are stronger than the original definition of covariate shift assuming only the change of marginal distribution $p(\rv X)$ and the invariance of the class-posterior probability $p(\rv Y \mid \rv X)$.
As a result, there exist covariate shift scenarios that cannot be attributed to selection bias when $\t{P} \ll P$ is not the case.

\subsubsection{Generalized Target Shift}
In the literature, generalized target shift refers to the situation where the prior probability $p(\rv Y)$ and the conditional distribution $p(\rv X \mid \rv Y)$ both change across training and test domains, however, with some invariance assumptions in the latent space \citep{zhang2013domain, gong2016domain, yu2020label}.
Generalized target shift is a subcase of $\tau \colon X \tm Y \karrow X$ $\otm$ $\lambda \colon X \tm Y \karrow Y$ corruption that can be formulated as a corrupted experiment.
Note that simplified sub-examples can also manifest as a generalized target shift; however, it is important to avoid degenerating into the basic $\tau \colon X \karrow X$ corruption, as it would violate the requirement of corrupting the label distribution.

\subsubsection{Concept Drift, Concept Shift, and Sampling Shift}
Concept drift refers to the situation where data evolves over time, leading to different categorizations depending on the nature of the change. Typically, concept drift between time point $t_0$ and $t_1$ is characterized by $p_{t_0}(\rv X, \rv Y) \neq p_{t_1}(\rv X, \rv Y)$ \citep{widmer1996learning, gama2014survey, Lu_2018}. In our words, $p_{t_1}(\rv X, \rv Y)$ can be seen as a corrupted version of $p_{t_0}(\rv X, \rv Y)$. Given its generality, this case can be associated with every corruption in our framework; therefore, the most general correspondence is the $\tau \colon X \tm Y \karrow X$ $\otm$ $\lambda \colon X \tm Y \karrow Y$ joint Markov kernel.

There are two types of concept drifts popular in the literature:
\begin{itemize}
    \item Concept shift \citep{vorburger2006entropy, widmer1993effective, salganicoff1997tolerating}: in this case, $p(\rv Y \mid \rv X)$ changes over time, and such changes can occur with or without changes on $p(\rv X)$, often referred to as concept shift; in our framework, this is a subcase of $\tau \colon X \tm Y \karrow X$ $\otm$ $\lambda \colon X \tm Y \karrow Y$ corruption. More details in \cref{tab:trad-tax}.
    \item Sampling shift \citep{tsymbal2004problem, widmer1993effective, salganicoff1997tolerating}: here, $p(\rv X)$ changes over time while $p(\rv Y \mid \rv X)$ remains invariant, also known as virtual drift; in our framework, this is a subcase of $\tau \colon X \tm Y \karrow X$ $\otm$ $\lambda \colon X \karrow Y$ corruption. More details are provided in \cref{tab:trad-tax}.
\end{itemize}

However, in the literature, concept drift is also defined with more invariance assumptions. For example, in \citet{moreno2012unifying}, they define concept drift as $p(\rv Y \mid \rv X)$ changing while $p(\rv X)$ remains invariant or $p(\rv X \mid \rv Y)$ changing while $p(\rv Y)$ remains invariant. Similar to instance- and label-dependent noise and covariate shift, they are examples of $\lambda \colon X \tm Y \karrow Y$ corruption and $\tau \colon X \tm Y \karrow X \otm \lambda \colon Y \karrow Y$ corruption that involve more corrupted spaces at different time points.

\section{Comparison with Other Taxonomies}
\label[section]{sec:other-tax}

We notice that most of the taxonomies available in the literature are based on the notion of invariance, inducing taxonomies very different from ours.
We here connect our work to other categorization paradigms for distribution shifts, although without claiming it to be a comprehensive review.

We divide taxonomies in two main groups: the traditional ones, focusing on identifying which probability in the set $\{ \pi_Y, \pi_X, E, F\}$ is forced to be left invariant and which one is forced to be corrupted  \citep{moreno2012unifying}, and the causal ones, where a causal graph structure is associated to the corruption process \citep{zhang2020domain} and hidden structures are possibly involved so that some latent feature is left unchanged by the corruption \citep{kull2014patterns, subbaswamy2022unifying}. 
Notice that in none of the cited works the corrupted distribution is assumed to have a specific form or to be ``close enough'' to the clean one. 
We do not review these other cases, because they are too far from our point of view and objective.

\begin{table*}[t]
    \centering
    \caption{Traditional taxonomies resume.}
    {\footnotesize
    \begin{tabular}{cccccc} 
        \toprule
        Corrupted & Invariant & \thead{Name in \\ \citep{moreno2012unifying}} & \thead{DAG in \\ \citep{kull2014patterns}} & Ours \\
        \midrule
        \thead{at least one \\ among \\  $\{ \pi_X, F, E \}$, \\ according to \\ compatibility} & $\pi_Y$ & \thead{concept shift\footnotemark \\when $\rv Y \to \rv X$ } & 
        \tikz[baseline=-5mm,edge from parent/.style={draw,-latex}, sibling distance=.8cm, level distance=.8cm]{% 
            \node (topnode) at (0,0) { $\rv D$ } 
                child { 
                    node (x) { $\rv X$ } 
                } 
                child {
                    node (y) { $\rv Y$ }
                    edge from parent[draw=none] 
                }
            ;
            \draw[latex-] (x) --+ (y) ;
        }% 
        & \thead{subcase of \\$\ka \colon X \karrow X$} 
        \\
        \midrule
        \thead{at least one \\ among \\  $\{ \pi_Y, F, E \}$, \\ according  to \\ compatibility} & $\pi_X$ & \thead{concept shift\footnotemark\\ when $\rv X \to \rv Y$ } & 
        \tikz[baseline=-5mm,edge from parent/.style={draw,-latex}, sibling distance=.8cm, level distance=.8cm]{% 
            \node (topnode) at (0,0) { $\rv D$ } 
                child { 
                    node (x) { $\rv X$ } 
                    edge from parent[draw=none] 
                } 
                child {
                    node (y) { $\rv Y$ }
                }
            ;
            \draw[-latex] (x) --+ (y) ;
        }% 
        & \thead{subcase of \\$\lambda \colon Y \karrow Y$}\\
        \midrule
        \thead{at least $\pi_Y$, \\ causing $\pi_X$ \\ or  $F$ \\ to change} & $E$ & \thead{prior probability \\shift\footnotemark ~when \\$\rv Y \to \rv X$ } &
        \tikz[baseline=-5mm,edge from parent/.style={draw,-latex}, sibling distance=.8cm, level distance=.8cm]{% 
            \node (topnode) at (0,0) { $\rv D$ } 
                child { 
                    node (x) { $\rv X$ }
                    edge from parent[draw=none] 
                } 
                child {
                    node (y) { $\rv Y$ }
                }
            ;
            \draw[latex-] (x) --+ (y) ;
        }%
        & \thead{at least a \\ $\lambda \colon Y \karrow Y$ \\ subcase, at most \\ $\ka \colon X \tm Y \karrow X$\\$ \otm $\\$\lambda \colon X \tm Y \karrow Y$}\\
        \midrule
        \thead{at least $\pi_X$, \\ causing $\pi_Y$ \\ or  $E$ \\to change} & $F$ & \thead{covariate shift\footnotemark \\when $\rv X \to \rv Y$ } & 
        \tikz[baseline=-5mm,edge from parent/.style={draw,-latex}, sibling distance=.8cm, level distance=.8cm]{% 
            \node (topnode) at (0,0) { $\rv D$ } 
                child { 
                    node (x) { $\rv X$ } 
                } 
                child {
                    node (y) { $\rv Y$ }
                    edge from parent[draw=none] 
                }
            ;
            \draw[-latex] (x) --+ (y) ;
        }% 
        &  \thead{at least a \\ $\ka \colon X \karrow X$ \\ subcase, at most \\ $\ka \colon X \tm Y \karrow X$\\$ \otm $\\$\lambda \colon X \tm Y \karrow Y$ } \\
        \bottomrule
    \end{tabular}
}

    \label{tab:trad-tax}
\end{table*}
\footnotetext[16]{as sole attribute noise: \citep{shackelford1988learning, goldman1995can, zhu2004class, williamson2022information}}
\footnotetext[17]{as sole class-conditional noise: \citep{angluin1988learning, blum1998combining, natarajan2013learning, patrini2017making, van2017theory, williamson2022information}; in general: \citep{yamazaki2007asymptotic,alaiz2008assessing}} 
\footnotetext[18]{or label shift, or class imbalance: \citep{japkowicz2002class, he2009learning, buda2018systematic, lipton2018detecting, tang2022invariant}}
\footnotetext[19]{\citep{shimodaira2000improving, quinonero2008dataset, sugiyama2012machine, zhang2020one}}

\subsection{Traditional and Causal Taxonomies} 
Focusing on the first case, a complete \emph{traditional taxonomy} has four types of possible corruptions. 
Taking into account which marginal or conditional probability is forced to be corrupted, we obtain a finite number of corruption subcases of these four macro-types. 
However, the different cases obtained may overlap, as it is schematically shown in \cref{tab:trad-tax}.  
The cases that have a clear correspondence with ours are the ones leaving invariant a marginal distribution, generating simple noises. 
All the other cases cannot be directly mapped into our taxonomy, so we explicitly write the range of corruption types covered by them.

As for \emph{causal taxonomies}, based on causal graphs, they are more difficult to describe in a unified way since different applications lead to different notations. We then avoid doing so, and limit ourselves to qualitatively compare them with our work.

A common trend is to identify the current space we live in with a variable $\rv D$, the \emph{domain} or \emph{environment}, possibly taking values in $\bb{N}$. 
This variable is then included in the causal graph indicating on what it is acting, as done in the examples in \cref{tab:trad-tax}.
In the case described by \cref{df:corruption} we restrict it to take values in $\{0,1\}$, the clean and corrupted environments.
This representation is again missing some our corruptions, since it is only possible to encode $X$ and $Y$ changing across domains and not whether other environments influence the current one.
The shifts in \citet[Fig.~3]{kull2014patterns} involving hidden variables (concept shift subcases) resemble our idea of a ``latent process'' influencing the current environment, but still fail to cover all the possible cross-domain influence in \cref{fig:corruptions}. 
An additional limitation of the causal approach lies in the causal assumption itself; we are forced, in this setting, to only consider one conditional probability between $E$ and $F$ to be a valid representation of the generative process, while in our framework we are not inherently forced to make this choice. 
We although can still make use of causal information in case it is available, as well as have more favorable causal relationship between $X$ and $Y$ depending on which corruption type we want to analyze. This is apparent in the Data Processing Equalities we prove.

In both the described classes of taxonomies, it not natural nor simple to define a hierarchy of corruptions. 
In particular, in the traditional taxonomy the specification of what is corrupted leaves room for other components to be forced to be influenced, creating overlaps between cases.
As for composing them, a DAG representation of corruption model can facilitate their chaining. Nevertheless, feasibility rules are rather complex and unclear to understand, given the overlapping nature of the corruptions and identifiability problems for causal representations \citep{pearl2009causality}.

\section{Bayesian Inversion in Category Theory}
\label[section]{sec:inversion}

In this section, we provide a more formal definition of the Bayesian inverse of a Markov kernel, based on some existing results from category theory applied to Bayesian learning \citep{dahlqvist2016bayesian}. In fact, Bayesian update is exactly kernel inversion. These results guarantee the valid and proper utilization of the inverse kernel in the current paper. Before delving into the details, we introduce relevant categorical concepts, establishing the necessary background to proceed. For a comprehensive overview of category theory, we recommend interested readers to refer to \citep{mac2013categories}.

\subsection{Categorical Concepts}
To begin, let \gls*{mes} be the category of measurable spaces with measurable maps as morphisms, and \gls*{pol} be the category of \emph{Polish spaces}, i.e., separable metric spaces for which a complete metric exists, with continuous maps as morphisms.
The functor $\mc B : \Pol \to \Mes$ associates any Polish space to the measurable space with the same underlying set equipped with the Borel $\sigma$-algebra, and interprets continuous maps as measurable ones. Measurable spaces in the range of $\mc B$ are \emph{standard Borel spaces}, which are important because the \emph{regular conditional probabilities} are known to exist in them, but not in general \citep{faden1985existence}. Therefore, they will be used as the building block of the \gls*{krn} category in the subsequent Bayesian inversion theorem.

The \emph{Giry monad} is the monad on a category of suitable spaces which sends each suitable space $X$ to the space of suitable probability measures on $X$. 
In this case, the set of suitable spaces is the one of the \gls*{mes} category induced by the functor $\mc B$. 
To define it more formally, we now consider the triple $(\mc P , \mu, \delta)$:
\begin{itemize}
    \item the functor $\mc P$ is such that we assign to every space $X$ in \gls*{mes} the set of all probability measures on $X$, $\mc P(X)$. 
    This is equipped with the smallest $\sigma$-algebra that makes the evaluation function $ev_{B}: \mc P(X) \to [0,1] = P \mapsto P(B)$ measurable, for $B$ a measurable subset in $X$;
    \item the multiplication of the monad, $\mu: \mc P^2 \Rightarrow \mc P$, is defined by $$\mu_X(Q)(B)=\int_{q\in {\mc P}(X)}ev_{B}(q)dQ\,;$$
    \item the unit of the monad, $\delta: Id \Rightarrow \mc P$, sends a point $x \in X$ to the Dirac measure at $x$.
\end{itemize}
This equips the endofunctor $\mc P: \Mes \to \Mes$ into a monad, that is, the Giry monad $\G \coloneqq (\mc P, \mu, \delta)$ on measurable spaces.

The Kleisli category of $\G$, denoted by $\mc K\ell$, has the same objects as $\Mes$, and the morphism $\ka \colon X \karrow Y$ in $\mc K\ell$ is a kernel $\ka \colon X \to {\mc P}(Y)$ in $\Mes$. The Kleisli composition of kernel $\ka \colon X \karrow Y$ with $\lambda \colon Y \karrow Z$ is given by $\lambda \circ_{\G} \ka= \mu_{Z} \circ {\mc P}(\lambda) \circ \ka$. 
The action of the functor $\mc P$ on a kernel results, by definition, in the push-forward operator $\mc P (\ka)(\cdot) \coloneqq (\cdot) \circ \ka^{-1}$, defined on a suitable space of probabilities.
Hence, $\circ_{\G}$ is the same as the chain composition we defined in \cref{p:comp}.

\begin{table*}[t]
    \centering
    {\footnotesize
    \caption{Comparison of categorical concepts in \citet{dahlqvist2016bayesian} and probabilistic concepts in this paper.}
    \label{tab:category}
    \begin{tabular}{cc} 
        \toprule
        Categorical & Probabilistic\\
        \midrule
        Kleisli category of Giry monad $\G$, $K\ell$ & \thead{measurable spaces as objects and Markov kernels as arrows} \\
        \midrule
        arrows in category $1 \downarrow \mc K\ell$ & \thead{$\mc M (X,Y)$ where $X$ and $Y$ have marginals $p$ and $q$, respectively} \\
        \midrule
        arrows in category $1 \downarrow F$ & \thead{subset of the above $\mc M (X,Y)$ with \\measure-preserving maps induced by identical kernels $\delta$} \\
        \midrule
        Kleisli composition $\circ_\G$ & \thead{chain composition $\circ$ in \cref{p:comp} with transitional kernels} \\
        \midrule
        $\alpha^{X}_{Y} : \text{Hom}_\Krn(X,-) \to \Gamma(X,-)$ & \thead{product composition $\tm$ in \cref{p:prod} with a kernel and a probability} \\
        \bottomrule
    \end{tabular}
    }
\end{table*}

\subsection{The Bayesian Inversion Theorem}
\label{subsec:category-inversion}

\citet{dahlqvist2016bayesian} investigates how and when the Bayesian inversion of the Markov kernel is defined, both directly on the category of measurable spaces, and indirectly by considering the associated linear operators (i.e., Markov transition, see \citet{cinlar2011ProbabilityAS}). 
Below, we only introduce the first result of the Bayesian inversion theorem, given the focus of Markov kernels we have in the current paper, and then describe the pseudo-inversion operation in \cref{p:part-chain} in a more formal way.

The category of Markov kernel considered here is the one of \textbf{\textit{typed kernel}}. 
Their definition is tied to a fixed probability $p$ on $X$ and a fixed probability $q$ on $Y$, so that $\ka \circ_\G p = q$, instead of being characterized for every probability on $X$ and every reachable output. 
In general, one can define Markov kernels as operators on the space of probabilities; that is not our interest, as we tie the concept of corruption to a specific couple on the clean and corrupted distribution.
This remark is also crucial for understanding our notion of exhaustiveness in \cref{sec:exhaustive}.

The key object for building the inversion operation is the \gls*{krn} category, similar to our notion of space of Markov kernels $\mc M(X,Y)$, but with an equivalence relation acting on it. We describe its construction in the following steps.
\begin{enumerate}
    \item  Let $F \colon \Mes \to \mc K\ell$ be the functor embedding \gls*{mes} into $\mc K\ell$ which acts identically on spaces and maps measurable arrows $\ka \colon X \to Y$ to Kleisli arrows $F(\ka) = \delta_Y \circ \ka$. 
    This means that $F(\ka)$ only allows one possible jump at each $x$ in $X$, with $\delta_Y$ an identical jump (i.e., a deterministic kernel).

    \item It further induces the category $1 \downarrow F$ of probabilities $p: 1 \karrow {\mc P}(X)$, denoted by $(X,p)$, and morphisms $\ka \colon (X,p) \karrow_{\delta} (Y,q) $ as degenerate arrows $F(\ka): X \karrow Y$ s.t. $q = F(\ka) \circ_\G p = {\ttP} (\ka)(p)=p \circ \ka^{-1}$. In more familiar terms, this is saying that $q$ is the push-forward of $p$ along $\ka$. $1 \downarrow F$ includes all measure-preserving maps induced by degenerate arrows.

    \item When the arrows are not degenerate, we obtain the supercategory $1 \downarrow \mc K\ell$ with the same objects. Specifically, in this category, an arrow from $(X,p)$ to $(Y,q)$ is any Kleisli arrow $\ka \colon X \karrow Y$ s.t. $q = \ka \circ_\G p$, and the arrows are what we denoted as $\mc M (X,Y)$, where $X$ has marginal probability $p$ and $Y$ has marginal probability $q$.

    \item Markov kernels cannot be inverted as they are, because of their \emph{non-singularity}. Lemma 3 in 
    \citet{dahlqvist2016bayesian} characterizes it by proving that for a kernel $\ka \colon (X,p) \karrow (Y,q) $ there are $p$-negligibly many points jumping to $q$-negligible sets.
\end{enumerate}

Once the non-singularity is understood, we can define an equivalence relation on $1 \downarrow \mc K\ell$ that allows a well-posed definition of the inverse kernel.

\begin{definition}\label[definition]{def:equiv-of-kerns}
    For all objects $(X,p), (Y,q)$, $R_{(X,p), (Y,q)}$ is the smallest equivalence relation on $Hom_{1 \downarrow \mc K\ell}(X,Y)$  such that
    $$(\ka,\ka') \in R_{(X,p), (Y,q)} \enspace \Leftrightarrow \enspace \ka = \ka' \enspace p\text{-a.s.} $$
\end{definition}

They prove $R$ to be a congruence relation on $1 \downarrow \mc K\ell$ in their Lemma 4. This congruence relation allows us to define the quotient category, with proper morphisms.

\begin{definition}
    The category {\normalfont \gls*{krn}} is the quotient category $(1 \downarrow \mc K\ell) / R \;$.
\end{definition}

Having defined the category, we have to build the functions that are going to constitute the Bayesian inversion operator, \ie, a bijection between $\text{Hom}_\Krn((X,p),(Y,q))$ and $\text{Hom}_\Krn((Y,q),(X,p))$. 
There are two mappings between the \gls*{krn} category and the space of couplings associated to $(X,p),(Y,q)$. 
The first is equivalent to the product composition we defined in \cref{p:prod} applied to a kernel (\ie, conditional probability) and a probability, and is formally written as
$$\alpha^{X}_{Y} : \text{Hom}_\Krn((X,p),(Y,q)) \to \Gamma((X,p),(Y,q)) \enspace \text{s.t.} \enspace \alpha^{X}_{Y}(\ka)(B_X \tm B_Y) \coloneqq \int_{x \in B_X} \ka(x)(B_Y) dp \,,$$
with $\Gamma((X,p),(Y,q)) \subset {\mc P}(X \tm Y)$ the typed couplings associated to the marginals $(X,p),(Y,q)$.
The second is defined as its inverse operation, and it is decomposing a joint probability along a fixed marginal distribution (aka, disintegrating it), \ie,
\begin{align*}
     & D^{X}_{Y} : \Gamma((X,p),(Y,q)) \to \text{Hom}_\Krn((X,p),(Y,q)) \\ 
     & \text{s.t.} \quad D^{X}_{Y}(\gamma) \coloneqq \ttP(\pi_Y) \circ \pi_X^{\dag},  \ \gamma \in  \Gamma((X,p),(Y,q)) ,\\
     & \text{and} \quad \gamma(B_X \tm B_Y) \coloneqq \int_{x \in B_X} D^{X}_{Y}(\gamma) (x)(B_Y) dp \,,
\end{align*}
with $(\cdot)^{\dag}$: adjoint operator.
As one is the inverse of the other, they are both obviously bijective and the one-to-one correspondence between typed kernels and couplings is proved. 
Hence, we formally define the Bayesian inverse as in the following:

\begin{definition}
The Bayesian inverse of a typed kernel $\ka$ from $(X, p)$ to $(Y, q)$, is defined as 
$$(\cdot)^{\dag}: \ka \mapsto \ka^{\dag} \coloneqq \left(D^{Y}_{X} \, \circ \, \ttP(\pi_Y \tm \pi_X) \, \circ \, \alpha^{X}_{Y} \right) (\ka) \,,$$
with $\ttP(\pi_Y \tm \pi_X):  \Gamma((X, p),(Y, q)) \to  \Gamma((Y, q), (X, p))$ being the  permutation map.
\end{definition}

As the Bayesian inverse has been defined as a bijection between $\text{Hom}_\Krn((X,p),(Y,q))$ and $\text{Hom}_\Krn((Y,q),(X,p))$, it is always guaranteed to exist in this setting. 

\begin{proposition}[Bayesian Inversion Theorem]
The Bayesian inverse of a typed kernel $\ka$ from $(X, p)$ to $(Y, q)$ exists and is equivalently one of the following objects: 
\begin{enumerate}
    \item $\ka^{\dag} : (Y, q) \to (X, p)\, \in$ {\normalfont \gls*{krn}} when $\ka$ is seen as element of {\normalfont \gls*{krn}}, such that 
    $ (\ka^{\dag} \circ_{\G} \ka) \circ_{\G} q = \delta_{Y} \circ_{\G} q$ and $ (\ka \circ \ka^{\dag}) \circ_{\G} p = \delta_{X} \circ_{\G} p\;;$ 
    
    \item $\ka^{\dag} : Y \karrow X\, \in \mc{M}(Y,X)$ when $\ka$ is seen as element of $\mc{M}(X,Y)$, such that 
    $ (\ka^{\dag} \circ_{\G} \ka) \circ_{\G} q \equiv_{R}  \delta_{Y} \circ_{\G} q$ and $(\ka \circ_{\G} \ka^{\dag}) \circ_{\G} p \equiv_{R} \delta_{X} \circ_{\G} p \;.$ 
\end{enumerate}
Here, $\delta_{(\cdot)}$ indicates the identical kernel on the set $(\cdot)$, induced by the Dirac delta distribution.
\end{proposition} 
\begin{proof}
    The statement in (1) is a direct consequence of \citet{dahlqvist2016bayesian}.%, as
    %$$(\ka^{\dag} \circ_{\G} \ka) = (\ka^{\dag} \circ_{\G} \ka^{\dag\dag}) = ( D^{Y}_{X} \, \circ \, \ttP(\pi_Y \tm \pi_X) \, \circ \, \alpha^{X}_{Y} ),$$
    %and similarly for $(\ka \circ \ka^{\dag})$.
    As for (2), we are only using \cref{def:equiv-of-kerns}.
\end{proof}

\begin{remark}
We can understand the Bayesian inverse of a corruption kernel $\ka \in \mc M(Z,Z)$ from $(Z, \mc Z, P)$ to $(Z, \mc Z, \t P)$ that distorts $\t P(A) = \int_A \int_Z \ka(z, d\t z)\, P(d z) \enspace \forall A \in \mc Z \,$ 
as a Markov kernel $\ka^\dag \in \mc M(Z,Z)$ satisfying $$\int_B \ka(z, A)\, P(d z) = \int_A \ka^\dag(\t z, B)\, \t P(d \t z) \quad \forall A \in \mc Z, \, B \in \mc Z.$$
This formulation extends the discrete Bayes' rule $P(\t z \mid z)P(z) = P(z \mid \t z)P(\t z) \, {\forall z, \t z \in Z}$.
Hence, in the discrete case, the Bayesian inverse always exists and can be expressed as $$\ka^\dag(z\mid \t z) \coloneqq \frac{P(z)\ka(\t z \mid z)}{\t P(\t z)} \, \text{for} \, z, \t z \in Z \, \text{with} \, \t P(\t z) \neq 0.$$
This formula ensures the uniqueness of $\ka^{\dag}$ within the support of $\t P$, as all components are unique. 
However, outside the support when $\t P$ is zero, the uniqueness may not hold, requiring a non-fixed value for $\t z \in Z$ where $\t P(\t z) = 0$.

In the continuous case, the Bayesian inverse may not exist. To ensure $\ka^\dag \in \mc M(Z,Z)$ is well-defined, it must satisfy the conditions of being a Markov kernel, as defined in \cref{def:markov-kernel}, where the mapping $\t z \to \ka^\dag(\t z,B)$ is $\mc{Z}$-measurable for every set $B \in \mc{Z}$, and the mapping $B \to \ka^\dag(\t z,B)$ is a probability measure on $(Z,\mc{Z})$ for every $\t z \in Z$, \emph{for the standard Borel space $(Z, \mc Z)$}. Under this condition, the Bayesian inverse always exists, and it is uniquely defined within the support of $\t P$, where uniqueness is represented by an equivalence class of kernels that are $\t P$-a.s. equal.
\end{remark}

\subsection{Exhaustiveness of Markovian Corruption}
\label[section]{sec:exhaustive}

As we noticed in \cref{sec:related}, Markov kernels are not the only possibility for modeling corruption, but we proved that given a clean and corrupted space we can always find a Markov kernel that connects the two distributions.
In particular, we define the operations $\alpha^{X}_{Y}$ and $D^{X}_{Y}$ for typed kernels, where one is the inverse of the other by construction (\cref{subsec:category-inversion}). 
They are the operations representing the bijection between the space of Markov kernels typed for $p, q$ and the space of couplings with marginals $p, q$. 
Hence, they are proving that {\itshape for each couple of probability spaces, there exists a Markov kernel sending one into the other corresponding to a possible associated coupling.}

\section{Proofs for Data Processing Equality Results}
\label[section]{sec:proof4} 
Recall that $\pi_Y$ is a prior distribution on $Y$, and the notation $\ka_{\rv X}$ stands for the kernel $\ka$ evaluated on the parameter $\rv X$, \eg, $E_{\rv{Y}}, F_{\rv{X}}$, and  $\ka_{x} \coloneqq \ka_{\rv X=x}$.
The kernel $\delta_Z$ denotes a kernel induced by the Dirac delta measure from $(Z,\mc{Z})$ to $(Z,\mc{Z})$.

In the proofs we will use a continuous notation for measures on $Y$, for the sake of simplicity and homogeneity. 
However, notice that all the $\lambda$ kernels are actually (parameterized) stochastic matrices $\Lambda=[\Lambda_{\t y y}]$, where $\Lambda_{\t y y}=p(\t{\rv Y}=\t y \mid {\rv Y}=y)$  for simple corruptions and $\Lambda_{\t y y}(x)=p(\t{\rv Y}=\t y \mid {\rv Y}=y, {\rv X}=x)$ for dependent corruptions. Note that both $y$ and $\t y$ range in $Y$, and thus they are squared matrices. In \cref{thm:br-s-2dy} and \cref{thm:br-y-corr}, the kernel $\lambda$ acting on the function $\ell \circ \mc{H}$ is actually the transpose of the stochastic matrix $\Lambda$:
\be 
\sum_{\t y \in Y} \lambda_x(y, d\t {y}) \, \ell (h(x), \t y) = \sum_{\t y \in Y} \Lambda_{\t y y}^{\top} ( x ) \,  \ell_{\t y} (h(x)) = \widetilde{(\ell_{y} \circ h)}(x) \,. \notag
\ee

Below, Theorems \ref{thm:br-s-2dx} and \ref{thm:br-s-2dy} are proved based on the Lemmas concerning \gls*{br} changes under dependent $X$ and $Y$ corruptions, respectively.

\begin{lemma}[$X$ corruption]
\label[lemma]{thm: br-x-corr}
Consider the learning problem $(\ell, \mc{H}, P)$, with $\ell$ being a bounded loss, and $E\colon Y \karrow X$ its associated experiment such that $P = \pi_Y \tm E$ for a suitable $\pi_Y$. 
Let $\cemph{ \tau \otm \delta_Y } $ be a corruption acting on this problem, with $\tau \in \mc M (X \tm Y, X)$. Then, we obtain
\be 
 \Big( \ell \circ \mc{H}, (  \pi_Y \tm E ) \circ (\cemph{\tau}\otm\cemph{\delta_Y}) \Big) =
 \Big( \ell \circ \mc{H}, \pi_Y \circ \big( ( E \circ_X \cemph{\tau}) \otm \cemph{\delta_Y} \big) \Big) \brequiv
 \Big( \cemph{\tau}(\ell \circ \mc{H} ) , \pi_Y \tm E \Big) \,. \nonumber
\ee
Moreover, if $\tau \in \mc M (X, X)$, we have
\be 
 \Big( \ell \circ \mc{H}, (  \pi_Y \tm E ) \circ (\cemph{\tau}\otm\cemph{\delta_Y}) \Big) =
 \Big( \ell \circ \mc{H}, \pi_Y \circ \big( ( E \circ \cemph{\tau}) \otm \cemph{\delta_Y} \big) \Big) \brequiv
 \Big( \cemph{\tau}(\ell \circ \mc{H} ) , \pi_Y \tm E \Big) \,. \nonumber
\ee
\end{lemma} 

\begin{proof}
    Let $A \in \mc{X} \tm \mc{Y}$, and $\pi_y$ be the $y$-th entry of the $\pi_Y$ probability vector. By definition of all the objects involved, the action of $\tau \otm \delta_Y$ on $P$ is
    \begin{align*}
        \t P(A) &= \int_{(\t x, \t y) \in A} \int_{(x,y) \in X\tm Y} \tau_y(x, d \t x) \, \delta_y(d\t{y}) \, P(dx,dy)\\ % L: I had deleted this by mistake, @Nan good catch!
        &= \int_{(\t x, \tt y) \in A} \left[ \sum_{y \in Y} \left( \int_{x \in X} \tau_y(x, d\t {x}) \, E_y(dx)  \right) \delta_y(d\t {y}) \, \pi_y \right]\\
        &= \int_{(\t x, \t y) \in A} \left[ \sum_{y \in Y}  (E \circ_X \tau)_y(d\t {x}) \, \delta_y(d\t {y}) \, \pi_y \right] \\
        &=  \left[ \pi_y \circ \big( (E \circ_X \tau) \otm \delta_Y \big) \right] (A) \,.
    \end{align*}
    We can hence rewrite the risk \wrt $\t P \coloneqq \pi_y \circ [ (E \circ_X \tau) \otm \delta_Y ]$ as
    \begin{align}
        \bb E_{ (\t{ \rv X},\t{ \rv Y}) \sim \t P } \left[ \ell(h_{\t{ \rv X}}, \t{ \rv Y}) ) \right]
        &= \sum_{\substack{y \in Y,\\ \t y \in Y}} \left[ \int_{\t x \in X} \ell(h_{\t x}, \t y) \left( \int_{x \in X}  \; \tau_y(x, d\t x)\, E_{y}(dx)  \right) \right] \delta_y(d\t y) \, \pi_y \notag\\
        %&= \sum_{y \in Y} \left[ \int_{x \in X} \left( \int_{\t x \in X} \left(\sum_{\t y \in Y} \delta_y(d\t y) \, \ell(h_{\t x}, \t y)  \right) \tau_y(x, d\t x)  \right)  E_{y}(dx) \right] \pi_y  \notag\\
        &= \sum_{y \in Y} \left[ \int_{x \in X} \left( \int_{\t x \in X} (\delta\ell_y \circ h)(\t x) \, \tau_y(x, d\t x)  \right)  E_{y}(dx) \right] \pi_y  \notag\\
        &= \sum_{y \in Y} \left[ \int_{x \in X} [ \tau (\delta\ell_{y} \circ h) ](x, y) \, E_{y}(dx) \right] \pi_y  \label{eq: y-dependency} \\
        &= \bb E_{ (\rv X,\rv Y) \sim ( \pi_Y \tm E ) } \left[ \big(\tau (\delta\ell_{\rv Y} \circ h)\big) (\rv X, \rv Y) \right] \,. \notag 
    \end{align} 

    %\lilac{Noting that $\inf_{h \in \mc H \subseteq \mc M( X, Y)}$ does not depend on $y$ and the $Y$ space is finite , we can write the equality also in terms of \gls*{cbr}:}
    % \begin{align}
    %     \inf_{h \in \mc H} \bb E_{ (\t{\rv X},\t{\rv Y}) \sim \pi_Y \circ [ (E \circ_X \tau) \otm \delta_Y ] } \left[ \ell(h_{\t{ \rv X}}, \t{ \rv Y}) ) \right] 
    %     &= \bb E_{ \t{\rv Y} \sim \pi_Y } \inf_{h \in \mc H} \bb E_{ \t{\rv X} \sim  \t E_{\t{\rv Y}} } \left[ \ell(h_{\t{ \rv X}}, \t{ \rv Y}) ) \right] \notag \\
    %     &= \bb E_{ \t{\rv Y} \sim \pi_Y } \cbr_{\ell \circ \mc H} [\t E_{\t{\rv Y}}] \,, \notag \\
    %     \inf_{h \in \mc H} \bb E_{ (\rv X,\rv Y) \sim ( \pi_Y \tm E ) } \left[ \big(\tau (\delta\ell_{\rv Y} \circ h)\big) (\rv X, \rv Y)  \right] 
    %     &=  \bb E_{\rv Y \sim \pi_Y} \inf_{h \in \mc H} \bb E_{ \rv X \sim E_{\rv Y} } \left[ \big(\tau (\delta\ell_{\rv Y} \circ h)\big) (\rv X, \rv Y)  \right] \notag \\
    %     &= \bb E_{\rv Y \sim \pi_Y} \cbr_{\tau(\delta\ell \circ \mc H)} [E_{\rv Y}] \,. \label{eq:y-corr-cbr}
    % \end{align}

    Let $\t{E}_{y}(d\t x) \coloneqq (E \circ_X \tau)_y(d\t x)$. We have that the associated \gls*{br} is
    \begin{align*}
        \inf_{h \in \mc H} \bb E_{ (\t{\rv X},\t{\rv Y}) \sim \pi_Y \circ [ (E \circ_X \tau) \otm \delta_Y ] } \left[ \ell(h_{\t{ \rv X}}, \t{ \rv Y}) ) \right] 
        &= \inf_{h \in \mc H} \bb E_{ \t{\rv Y} \sim \pi_Y } \bb E_{ \t{\rv X} \sim  \t E_{\t{\rv Y}} } \left[ \ell(h_{\t{ \rv X}}, \t{ \rv Y}) ) \right] \\ 
        &= \br_{\ell \circ \mc H} [\pi_Y \circ (\t E \otm \delta_Y)  ] \,,
    \end{align*}
    \begin{align}
        \inf_{h \in \mc H} \bb E_{ (\rv X,\rv Y) \sim ( \pi_Y \tm E ) } \left[ \big(\tau (\delta\ell_{\rv Y} \circ h)\big) (\rv X, \rv Y)  \right] 
        &= \inf_{f \in \tau (\ell \circ \mc H)} \bb E_{ (\rv X,\rv Y) \sim ( \pi_Y \tm E ) } \left[ f (\rv X, \rv Y)  \right] \notag \\
        &= \br_{\tau(\ell \circ \mc H)} [\pi_Y \tm E] \,, \label{eq:y-corr-br}
    \end{align}
    which are equal given the previous computations.
    We have defined and used in \cref{eq:y-corr-br} that $f(x,y) \coloneqq [ \tau (\delta\ell_{y} \circ h) ](x, y) , h \in \mc H$. 
    Such functions are the ones populating the minimization set $\tau(\ell \circ \mc H)$, denoting that $\tau$ acts on the composition of the loss and model class while $\delta$ only acts on $\ell$ and leaves it unchanged.
    If $\tau$ is simple, then the equations from \cref{eq: y-dependency} lead to a slightly different model class:
    \begin{align*}
        \br_{\ell \circ \mc{H}}\left[ \pi_Y \circ \big( (E \circ \tau) \otm \delta_Y \big) \right]
        &= \inf_{h \in \mc H} \sum_{y \in Y} \left[ \int_{x \in X} [ \tau (\delta\ell_{y} \circ h) ](x) \, E_{y}(dx) \right] \pi_y \\
        &= \inf_{f \in \tau(\ell \circ \mc{H})} \sum_{y \in Y} \left[ \int_{x \in X} f(x,y) \, E_{y}(dx) \right] \pi_y  =  \br_{\tau(\ell \circ \mc{H})}(\pi_Y \tm E) \,. 
    \end{align*}
\end{proof}

\begin{theorem}[2-dependent $\tau$, simple $\lambda$, \cref{thm:br-s-2dx}]
Consider the learning problem $(\ell, \mc{H}, P)$, with $\ell$ being a bounded loss, and $E\colon Y \karrow X$ its associated experiment such that $P = \pi_Y \tm E$ for a suitable $\pi_Y$. 
Let $\cemph{ (\tau\colon X \tm Y \karrow X) \otm (\lambda\colon Y \karrow Y)} $ be a corruption acting on this problem, then, we obtain

\be 
 \Big( \ell \circ \mc{H}, (  \pi_Y \tm E ) \circ (\cemph{\tau}\otm\cemph{\lambda}) \Big) =
 \Big( \ell \circ \mc{H}, \pi_Y \circ \big( ( E \circ_X \cemph{\tau}) \otm \cemph{\lambda} \big) \Big) \brequiv
 \Big( \cemph{\tau}(\cemph{\lambda\ell} \circ \mc{H} ) , \pi_Y \tm E \Big) \,. \nonumber
\ee
The functions contained in the new minimization set are defined as
$$ \tau (\lambda\ell \circ \mc{H}) \coloneqq \{ (x, y) \mapsto [ \tau (\lambda\ell_{y} \circ h) ](x,y) , h \in \mc H \} \,.$$
\end{theorem}

\begin{proof}
With this corruption formulation, we can replicate the proof of \cref{thm: br-x-corr} up to \cref{eq: y-dependency} by simply plugging in $\lambda$ instead of $\delta_Y$. Therefore, we obtain the thesis.
\end{proof}

We remark that in this case $\t P \neq \pi_Y \tm \t E$ with $\t{E}_{y}(d\t x) \coloneqq (E \circ_X \tau)_y(d\t x)$, \ie, the corrupted experiment is not given by the sole action of $\tau$, but also by the influence of $\lambda$. 
That is clarified further by corruption formula $\t P = \pi_y \circ [ (E \circ_X \tau) \otm \lambda ]$. We conclude that, in this more general case, it does not make sense to distinguish the effect of corruption on $E$ and $\pi$.

\begin{lemma}[$Y$ corruption] 
\label[lemma]{thm:br-y-corr}
Consider the learning problem $(\ell, \mc{H}, P)$, with $\ell$ being a bounded loss, and $F\colon X \karrow Y$ its associated posterior such that $P = \pi_X \tm F$ for a suitable $\pi_X$. 
Let $\cemph{\delta_X \otm \lambda}$ be a corruption acting on this problem, with $\lambda \in \mc M (X \tm Y, Y)$. Then, we obtain
\be 
  \Big( \ell \circ \mc{H}, ( \pi_X \tm F  ) \circ (\cemph{\delta_X}\otm\cemph{\lambda}) \Big) =
  \Big( \ell \circ \mc{H}, \pi_X \circ \big( \cemph{\delta_X} \otm ( F \circ_Y \cemph{\lambda} ) \big) \Big) \brequiv
  \Big( \cemph{\lambda\ell} \circ \mc{H}  , \pi_X \tm F \Big) \,. \nonumber
\ee
Moreover, if $\lambda \in \mc M (Y,Y)$, we have
\be 
  \Big( \ell \circ \mc{H}, ( \pi_X \tm F  ) \circ (\cemph{\delta_X}\otm\cemph{\lambda}) \Big) =
  \Big( \ell \circ \mc{H}, \pi_X \circ \big( \cemph{\delta_X} \otm ( F \circ \cemph{\lambda} ) \big) \Big) \brequiv
  \Big( \cemph{\lambda\ell} \circ \mc{H}  , \pi_X \tm F \Big) \,. \nonumber
\ee
\end{lemma}

\begin{proof}
    Let $A \in \mc{ X} \tm \mc{ Y}$. By definition of all the objects involved, the action of $\tau \otm \delta_Y$ on $P$ is
    \begin{align*}
        \t P(A) &= \int_{(\t x, \t y) \in A} \int_{(x,y) \in X\tm Y} \delta_x(d \t x) \, \lambda_x(y,d\t{y}) \, P(dx,dy) \\
        &= \int_{(\t x, \t y) \in A} \left[ \int_{x \in X} \left( \sum_{y \in Y} \lambda_x(y,d\t {y}) \, F_x(dy)  \right) \delta_x(d\t {x}) \, \pi_X(dx) \right]\\
        &= \int_{(\t x, \t y) \in A} \left[ \int_{x \in X}  (F \circ_Y \lambda)_x(d\t {y}) \, \delta_x(d\t {x}) \, \pi_X(dx) \right] \\
        &=  \left[ \pi_X \circ \big( (F \circ_Y \lambda) \otm \delta_X \big) \right] (A) \,.
    \end{align*}
    We can hence rewrite the risk \wrt $\t P \coloneqq \pi_X \circ [ (F \circ_Y \lambda) \otm \delta_X ]$ as 
    \begin{align}
        \bb E_{ (\t{ \rv X},\t{ \rv Y}) \sim \t P } \left[ \ell(h_{\t{ \rv X}}, \t{ \rv Y}) ) \right]
        &=  \int_{\substack{x \in X,\\ \t x \in X}} \left[ \sum_{\t y \in Y} \ell(h_{\t x},\t y) \left( \sum_{y \in Y}  \; \lambda(x, y, d\t y)\, F_x(dy) \right) \right] \delta_x(d\t x) \, \pi_X(dx) \notag\\
        %&= \int_{x \in X} \left[ \sum_{y \in Y} \left( \sum_{\t y \in Y} \left(\int_{\t x \in X} \delta_x(d\t x) \, (\ell_{\t y} \circ h)(\t x)  \right) \lambda(x, y, d\t y)  \right)  F_x(dy) \right] \pi_X(dx)  \notag\\
        %&= \int_{x \in X} \left[ \sum_{y \in Y} \left( \sum_{\t y \in Y} \delta(\ell_{\t y} \circ h)(x) \, \lambda(x, y, d\t y)  \right)  F_x(dy) \right] \pi_X(dx)  \notag\\
        &= \int_{\substack{x \in X, \\\t x \in X}} \left[ \sum_{y \in Y} \left( \sum_{\t y \in Y} \ell(h_{\t x}, \t y) \, \lambda(x, y, d\t y)  \right)  F_x(dy) \right] \delta_x(d\t x) \, \pi_X(dx) \notag\\
        &= \int_{x \in X} \left[ \sum_{y \in Y} \Big( \int_{\t x \in X} (\lambda\ell)(h_{\t x}, x, y) \delta_x(d\t x) \Big) F_x(dy) \right] \pi_X(dx) \notag \\
        &= \int_{x \in X} \left[ \sum_{y \in Y}  \Big( (\lambda\ell)(h_x, x, y) \Big) F_x(dy) \right] \pi_X(dx) \label{eq: x-dependency} \\
        &= \bb E_{ (\rv X,\rv Y) \sim ( \pi_Y \tm E ) } \left[  (\lambda\ell) (h_{\rv X}, \rv X, \rv Y) \right] \\
        &= \bb E_{ (\rv X,\rv Y) \sim ( \pi_Y \tm E ) } \left[  (\lambda\ell_{(\rv X, \rv Y)} \circ h)(\rv X)  \right] \,.  \notag
    \end{align} 
    Similarly to the proof provided for \cref{thm: br-x-corr}, we can switch to \gls*{br} and obtain
    $$\br_{\ell \circ \mc{H}}[\pi_X \circ \big( (F \circ_Y \lambda ) \otm \delta_X \big)]
    = \br_{\lambda\ell \circ \mc{H}}(\pi_X \tm F)\,,$$
    with functions $\lambda\ell(h_x, x, y) = (\lambda\ell_{(x,y)} \circ h)(x) \in \lambda\ell \circ \mc{H}$.
    If $\lambda$ is simple, then \cref{eq: x-dependency} leads to a simpler model class:
    \begin{align*}
        \br_{\ell \circ \mc{H}}[\pi_X \circ \big((F \circ \lambda) \otm \delta_X \big)]
        &= \inf_{h \in \mc{H}} \int_{x \in X} \left[ \sum_{y \in Y} (\lambda\ell)(h_x, y) F_x(dy) \right] \pi_X(dx) \\
        &= \inf_{f \in \lambda(\ell \circ \mc{H})} \int_{x \in X} \left[ \sum_{y \in Y}  f(x, y)  F_x(dy) \right] \pi_X(dx) \\
        &= \br_{\lambda(\ell \circ \mc{H})}(\pi_X \tm F)\,.
    \end{align*}
\end{proof}
    
% Note that the results can also be expressed in terms of $E$ using the generic corruption formulation: 
% \begin{align}
%     \br_{\ell \circ \mc{H}}((\pi_Y \tm E) \circ \ka) = 
%     &= \inf_{f \in \ell \circ \mc{H}} \int_{\t x \in X} \int_{x \in X}  \sum_{y \in Y, \t y \in Y} f(\t{x},\t y) \, \ka(x,y,d\t{x}d\t{y}) \, P(dx,dy) \notag\\
%     &= \inf_{f \in \ell \circ \mc{H}} \int_{x \in X}  \sum_{y \in Y} \ka f(x, y) \, P(dx,dy) =  \br_{\ka(\ell \circ \mc{H})}(\pi_Y \tm E) \,. \notag 
% \end{align}
% Such result, even if fitting in the comparison of experiments literature \citep{torgersen1991comparison}, does not give us any new insights on how the corruption acts on the sole experiment given the corruption decomposition formula $\ka = \tau \otm \lambda$.

\begin{theorem}[simple $\tau$, 2-dependent $\lambda$, \cref{thm:br-s-2dy}]
Consider the learning problem $(\ell, \mc{H}, P)$, with $\ell$ being a bounded loss, and $F\colon X \karrow Y$ its associated posterior such that $P = \pi_X \tm F$ for a suitable $\pi_X$. 
Let $\cemph{(\tau\colon X  \karrow X) \otm (\lambda\colon X\tm Y \karrow Y)}$ be a corruption acting on this problem, then, we obtain
\be 
  \Big( \ell \circ \mc{H}, ( \pi_X \tm F  ) \circ (\cemph{\tau}\otm\cemph{\lambda}) \Big) =
  \Big( \ell \circ \mc{H}, \pi_X \circ \big( \cemph{\tau} \otm ( F \circ_Y \cemph{\lambda} ) \big) \Big) \brequiv
  \Big( \cemph{\tau}(\cemph{\lambda\ell} \circ \mc{H} ) , \pi_X \tm F \Big) \,. \nonumber
\ee
The functions contained in the new minimization set are defined as
$$ \tau (\lambda\ell \circ \mc{H}) \coloneqq \{ (x, y) \mapsto [ \tau (\lambda\ell_{(x,y)} \circ h ) ](x) , h \in \mc H \} \,.$$ 
\end{theorem}

\begin{proof}
    With this corruption formulation, we can replicate the proof of \cref{thm:br-y-corr} up to \cref{eq: x-dependency} by simply plugging in $ \tau$ instead of $\delta_x$.
    Therefore, we obtain the thesis.
\end{proof}

\begin{theorem}[1-dependent and a 2-dependent, \cref{thm:br-1d-2d}]
Consider the clean learning problem $(\ell, \mc{H}, P)$, with $\ell$ being a bounded loss, $E\colon Y \karrow X$ its associated experiment such that $P = \pi_Y \tm E$ for a suitable $\pi_Y$, and $F\colon X \karrow Y$ its associated posterior such that $P = \pi_X \tm F$ for a suitable $\pi_X$. 

\begin{enumerate}
    \item Let $\cemph{(\tau\colon Y \karrow X) \otm (\lambda\colon X \tm Y \karrow Y)}$ be a corruption acting on the problem, then, we obtain
    \be
    \Big( \ell \circ \mc{H} , ( \pi_Y \tm E ) \circ ( \cemph{\tau}\otm \cemph{\lambda} ) \Big) = 
    \Big( \ell \circ \mc{H} , \pi_Y \circ \big( \cemph{\tau} \otm ( E \circ_X \cemph{\lambda} ) \big) \Big) \brequiv
    \Big( \cemph{\tau}(\cemph{\lambda}\ell \circ \mc{H}), \pi_Y \tm E \Big) \,. \notag 
    \ee
    The functions contained in the new minimization set are defined as
    $$ \tau (\lambda\ell \circ \mc{H}) \{ (x,y) \mapsto \tau [\lambda \ell_{(x,y)} \circ h] (y) , h \in \mc H \} \,.$$
    \item Let $\cemph{(\tau\colon X \tm Y \karrow X) \otm (\lambda\colon X \karrow Y)}$ be a corruption acting on the problem, then, we obtain
    \be
    \Big( \ell \circ \mc{H}, (\pi_X \tm F) \circ (\cemph{\tau} \otm \cemph{\lambda}) \Big) = 
    \Big( \ell \circ \mc{H}, \pi_X \circ \big( ( F \circ_Y \cemph{\tau} ) \otm \cemph{\lambda}  \big) \Big) \brequiv
    \Big( \cemph{\tau}(\cemph{\lambda}\ell \circ \mc{H}), \pi_X \tm F \Big) \,. \nonumber
    \ee 
    The functions contained in the new minimization set are defined as
    $$\tau (\lambda\ell \circ \mc{H}) \{ (x,y) \mapsto  \tau [\lambda \ell_x \circ h] (x, y), h \in \mc H \} \,.$$
\end{enumerate}
\end{theorem}

\begin{proof}
    Consider point 1 and let $A \in \mc X \tm \mc Y$. By definition of all the objects involved, the action of $\tau \otm \lambda$ on $P$ is
    \begin{align*}
        \t P(A) &= \int_{(\t x, \t y) \in A} \left[ \sum_{y \in Y} \left( \int_{x \in X}  \lambda_y(x, d\t {y}) \; E_y(dx) \right) \, \tau(y, d\t {x}) \, \pi_y \right] \\
        &= \int_{(\t x, \t y) \in A} \left[ \sum_{y \in Y} (E \circ_X \lambda)(y, d\t y) \, \tau(y, d\t {x})\, \pi_y \right] \\
        &= \int_{(\t x, \t y) \in A} \left[ \sum_{y \in Y} \big( \tau \otm (E \circ_X \lambda) \big)(y, d\t x, d\t y) \, \pi_y \right] 
        = \left[  \pi_Y \tm \big( \tau \otm (E \circ_X \lambda) \big) \right] (A) \,.
    \end{align*}
    We can then write the associated risk \wrt $\t P \coloneqq \pi_Y \tm \big( \tau \otm (E \circ_X \lambda) \big)$ as 
    \begin{align}
        \bb E_{( \t{ \rv X},\t{ \rv Y} ) \sim  \t P } [\ell(h_{\t{ \rv X}}, \t{ \rv Y}) ]
        &= \sum_{\substack{y \in Y,\\ \t y \in Y}} \left[ \int_{\t x \in X} \ell(h_{\t x}, \t y) \left( \int_{x \in X}  \; \lambda(x, y, d\t y)\, E_{y}(dx) \right) \right] \tau(y, d\t x) \, \pi_y \notag\\
        &= \sum_{y \in Y} \left[ \int_{x \in X} \left( \int_{\t x \in X} \left(\sum_{\t y \in Y} \lambda(x, y, d\t y)\, \ell(h_{\t x}, \t y)  \right) \tau(y, d\t x)  \right)  E_{y}(dx) \right] \pi_y  \notag\\
        &= \sum_{y \in Y} \left[ \int_{x \in X} \left( \int_{\t x \in X} (\lambda\ell_{(x,y)} \circ h)(\t x) \, \tau(y, d\t x)  \right)  E_{y}(dx) \right] \pi_y  \notag\\
        &= \sum_{y \in Y} \left[ \int_{x \in X} [ \tau (\lambda\ell_{(x,y)} \circ h) ](y) \, E_{y}(dx) \right] \pi_y  \notag \\
        &= \bb E_{ (\rv X,\rv Y) \sim ( \pi_Y \tm E ) } \left[ \big(\tau (\lambda\ell_{(\rv X, \rv Y)} \circ h)\big) ( \rv Y) \right] \,, \notag
    \end{align}
    which proves the thesis when minimizing over $h \in \mc H$. For proving point 2, we first rewrite  the action of $\tau \otm \lambda$ on $P$ as
    \begin{align*}
        \t P(A) &= \int_{(\t x, \t y) \in A} \left[ \int_{x \in X} \lambda(x, d\t y) \left( \sum_{y \in Y} \tau(x,y,d \t x) \,  F(x,dy)  \right)  \pi_X(dx) \right] \\
        &= \int_{(\t x, \t y) \in A} \left[ \int_{x \in X} \lambda(x, d\t y) \, (F \circ_Y \tau)(x, d\t {x}) \, \pi_X(dx) \right] 
        = \left[ \pi_X \tm \big( \lambda \otm (F \circ_Y \tau) \big) \right] (A) \,,
    \end{align*}
    and repeat a similar argument but for the $F$ kernel. We find a minimization space of functions $f(x,y) \coloneqq  \tau [\lambda \ell_x \circ h] (x, y) $.
    Thus, we obtain the thesis.
\end{proof}

\begin{corollary}[1-dependent $\tau$ and $\lambda$, \cref{thm:br-1d-1d}]
Consider the clean learning problem $(\ell, \mc{H}, P)$, with $\ell$ being a bounded loss, $E\colon Y \karrow X$ its associated experiment such that $P = \pi_Y \tm E$ for a suitable $\pi_Y$, and $F\colon X \karrow Y$ its associated posterior such that $P = \pi_X \tm F$ for a suitable $\pi_X$. 
Let $\cemph{(\tau\colon Y \karrow X) \otm (\lambda\colon X \karrow Y)}$ be a corruption acting on the problem, then, we obtain
\be
    \Big( \ell \circ \mc{H} , ( \pi_Y \tm E ) \circ ( \cemph{\tau}\otm \cemph{\lambda} ) \Big) = 
    \Big( \ell \circ \mc{H} , \pi_Y \circ \big( \cemph{\tau} \otm ( E \circ \cemph{\lambda} ) \big) \Big) \brequiv
    \Big( \cemph{\tau}(\cemph{\lambda}\ell \circ \mc{H}), \pi_Y \tm E \Big) \nonumber \,. 
\ee
or, equivalently,
\be
    \Big( \ell \circ \mc{H}, (\pi_X \tm F) \circ ( \cemph{\tau} \otm \cemph{\lambda} ) \Big) = 
    \Big( \ell \circ \mc{H}, \pi_X \circ \big( ( F \circ \cemph{\tau} ) \otm \cemph{\lambda} \big) \Big) \brequiv
    \Big( \cemph{\tau}(\cemph{\lambda}\ell \circ \mc{H}), \pi_X \tm F \Big) \,. \notag
\ee 
The functions contained in the new minimization set are defined as
$$ \tau (\lambda\ell \circ \mc{H}) \coloneqq \{ (x, y) \mapsto [ \tau (\lambda\ell_{x} \circ h ) ](y) , h \in \mc H \} \,.$$
\end{corollary} 

\begin{proof}
    We can replicate the proof of \cref{thm:br-1d-2d} by simply substituting $\lambda(x,d\t {y})$ in place of $\lambda(x,y,d\t {y})$ in the first point, and $\tau(y, d\t {x})$ for $\tau(x,y, d\t {x})$ in the second point.
    We then in both cases obtain functions $f(x,y)\coloneqq\tau [( \lambda \ell)_x \circ h] (y)$, \ie, comparing a point $x$ with a kernel on $\mc P(X)$ parameterized by $y$.
    Therefore, we obtain the thesis.
\end{proof}

\begin{theorem}[2-dependent $\ka$ and $\lambda$, \cref{thm:br-2-2}]
Consider the clean learning problem $(\ell, \mc{H}, P)$, with $\ell$ being a bounded loss, and let $\cemph{(\tau\colon X \tm Y \karrow X) \otm (\lambda\colon X \tm Y \karrow Y)}$ be a corruption acting on the problem. 
Then:
\begin{enumerate}
    \item the action of such corruption on the joint probability $P$ is equivalent to the one of the non-decomposed joint corruption;
    \item the action on the minimization set $\ell \circ \mc H$ induces the following {\normalfont \gls*{br}}-equivalence
    $$\big(\ell,\mc H, P \circ (\tau \otm \lambda) \big) \brequiv \big( \tau(\lambda\ell \circ \mc H), P \big)\,;$$
    \item the functions contained in the new minimization set are defined as
    $$ \tau (\lambda\ell \circ \mc{H}) \coloneqq \{ (x, y) \mapsto [ \tau (\lambda\ell_{(x,y)} \circ h ) ](x,y) , h \in \mc H \} \,.$$
\end{enumerate}
\end{theorem}

\begin{proof}
Let $A \in \mc{X} \tm \mc{Y}$. By definition of all the objects involved, the action of $\tau \otm \lambda$ on $P$ is
\begin{align*}
    \t P(A) &= \int_{(\t x, \t y) \in A} \left[ \sum_{y \in Y} \left( \int_{x \in X}  \tau_y(x, d\t {x}) \, \lambda_x(y, d\t {y}) \, E_y(dx) \right) \pi_y \right] \\
    &= \int_{(\t x, \t y) \in A} \left[ \int_{x \in X} \left( \sum_{y \in Y}   \tau_y(x, d\t {x}) \, \lambda_x(y, d\t {y}) \, F_x(dy) \right) \pi_X(dx) \right]\,.
\end{align*}
In both the formulations above, obtained by factorizing the joint probability $P$ in two different ways, we cannot isolate the action of one between $\lambda$ and $\tau$ on $F$ or $E$. That is, because of the dependence of $\lambda$ and $\tau$ on the couple $(x,y)$, and because the action of a kernel on a probability via a combination of \cref{p:comp,p:prod,p:part-chain} requires sequential integration. This concludes point 1.

As for point 2, we now want to consider the action on functions. This uses integration \wrt the corrupted variables $(\t x, \t y)$, and therefore allows sequential integration.
We have that the associate risk is equal to
\begin{align}
        \bb E_{( \t{ \rv X},\t{ \rv Y} ) \sim  \t P} [\ell(h_{\t{ \rv X}}, \t{ \rv Y}) ]
        %&= \int_{\t x \in X, \t y \in Y} \ell(h_{\t x}, \t y)  \left[ \int_{ x \in X,  y \in Y} \tau(x, y, d\t x) \, \lambda(x, y, d\t y) \, P(dx,dy) \right] \\
        &= \int_{\t x \in X, \t y \in Y} \ell(h_{\t x}, \t y)  \left[ \sum_{y \in Y} \left( \int_{ x \in X} \tau(x, y, d\t x) \, \lambda(x, y, d\t y)\, E_{y}(dx)  \right) \pi_y \right] \notag\\
        %&=  \sum_{y \in Y} \left[ \int_{ x \in X} \left( \int_{\t x \in X, \t y \in Y} \ell(h_{\t x}, \t y) \tau(x, y, d\t x) \, \lambda(x, y, d\t y)\, E_{y}(dx)  \right) \right] \pi_y \notag\\
        &= \sum_{y \in Y} \left[ \int_{ x \in X} \left( \int_{\t x \in X} \left( \sum_{\t y \in Y} \ell(h_{\t x}, \t y)\, \lambda(x, y, d\t y) \right) \tau(x, y, d\t x) \, E_{y}(dx)  \right) \right] \pi_y \notag \\
        &= \sum_{y \in Y} \left[ \int_{ x \in X} \left( \int_{\t x \in X} \lambda\ell(h_{\t x}, x, y) \, \tau(x, y, d\t x) \, E_{y}(dx)  \right) \right] \pi_y \notag \\
        &= \sum_{y \in Y} \left[ \int_{ x \in X} \tau [ \lambda\ell_{(x,y)} \circ h ](x,y) \, E_{y}(dx) \right] \pi_y  \\
        &= \bb E_{ (\rv X,\rv Y) \sim ( \pi_Y \tm E ) } \left[ \big(\tau (\lambda\ell_{(\rv X, \rv Y)} \circ h)\big) (\rv X, \rv Y) \right] \,. \notag
    \end{align}
    Following the same reasoning, we can also write
    $$\bb E_{( \t{ \rv X},\t{ \rv Y} ) \sim  \t P} [\ell(h_{\t{ \rv X}}, \t{ \rv Y}) ] = \bb E_{ (\rv X,\rv Y) \sim ( \pi_X \tm F ) } \left[ \big(\tau (\lambda\ell_{(\rv X, \rv Y)} \circ h)\big) (\rv X, \rv Y) \right] \,.$$
    We prove point 2 and 3 minimizing both the obtained risk equalities \wrt $h \in \mc H$.
\end{proof}

%\section{Proofs for {\normalfont \textsc{cl}} and {\normalfont \textsc{gcl}}}
%\label[section]{sec:proof5}
%\input{appendix/proof-correction}

\vskip 0.2in
\bibliography{bib-thesis}

\end{document}